%% file: main.tex
\documentclass[12pt,twoside,a4paper]{report}

\usepackage[utf8]{inputenc}
\usepackage{float}

\usepackage[english]{babel}
\usepackage{csquotes}
\usepackage{biblatex}
\usepackage{fancyhdr}

\usepackage{graphicx}
\usepackage{subfigure}

\graphicspath{ {./figures/} }

\usepackage[table]{xcolor}
\usepackage{array}
\usepackage{multirow}

\usepackage{amsmath}
\newcommand{\reporttitle}{AI in Pursuit of Happiness, Finding Only Sadness:
Multi-Modal Facial Emotion Recognition Challenge}
\newcommand{\reportauthor}{Carl Norman}
\newcommand{\supervisor}{Dimitrios Kollias}
\newcommand{\degreetype}{Computing (Machine Learning)}

\input{includes}


\date{September 2019}

\begin{document}

% load title page
\input{titlepage}

% page numbering etc.
\pagenumbering{roman}
\clearpage{\pagestyle{empty}\cleardoublepage}
\setcounter{page}{1}
\pagestyle{fancy}
\fancyhf{}

\fancyfoot[CE,CO]{\thepage}

%%%%%%%%%%%%%%%%%%%%%%%%%%%%%%%%%%%%
\begin{abstract}

    The importance of automated Facial Emotion Recognition (FER) grows the more common human-machine interactions become, which will only continue to increase dramatically with time. A common method to describe human sentiment or feeling is the categorical model the `7 basic emotions', consisting of `Angry', `Disgust', `Fear', `Happiness', `Sadness', `Surprise' and `Neutral'. The `Emotion Recognition in the Wild' (EmotiW) \cite{emotiw} competition is now in its 7th year and has become the standard benchmark for measuring FER performance. The focus of this paper is the EmotiW sub-challenge of classifying videos in the `Acted Facial Expression in the Wild' (AFEW) dataset, consisting of both visual and audio modalities, into one of the above classes. 
    
    Machine learning has exploded as a research topic in recent years, with advancements in `Deep Learning' a key part of this. Although Deep Learning techniques have been widely applied to the FER task by entrants in previous years, this paper has two main contributions: (i) to apply the latest `state-of-the-art' visual and temporal networks and (ii) exploring various methods of fusing features extracted from the visual and audio elements to enrich the information available to the final model making the prediction. 
    
    There are a number of complex issues that arise when trying to classify emotions for `in-the-wild' video sequences, which the above two approaches attempt to directly address. There are some positive findings when comparing the results of this paper to past submissions, indicating that further research into the proposed methods and fine-tuning of the models deployed, could result in another step forwards in the field of automated FER.

\end{abstract}

%%%%%%%%%%%%%%%%%%%%%%%%%%%%%%%%%%%%
\section*{Acknowledgments}

Firstly I would like to thank my supervisor Dimitrios for letting me explore this topic and providing crucial guidance throughout. 

\medskip

The staff, lecturers and course supervisors of the Imperial College London Department of Computing for expertly organising and delivering the MSc in Computing (Machine Learning) course, providing key instruction during the first half of the academic year that laid the foundations for this paper. Also, a special mention for the CSG team for patiently helping to fix bugs throughout this project.

\medskip

Finally, I am eternally grateful to friends and family for constantly supporting me during this master's and putting up with the trials and tribulations it has come with.

\clearpage{\pagestyle{empty}\cleardoublepage}

%%%%%%%%%%%%%%%%%%%%%%%%%%%%%%%%%%%%
\section*{}

\begin{center}

    Dedicated to Harnam Aulakh, a wise and querky man, but a better uncle you will not find...

\end{center}

\clearpage{\pagestyle{empty}}

%%%%%%%%%%%%%%%%%%%%%%%%%%%%%%%%%%%%
% \begin{figure}[tb]
% \centering
% \includegraphics[width = 0.4\hsize]{./figures/imperial}
% \label{fig:logo}
% \end{figure}

% \clearpage
%%%%%%%%%%%%%%%%%%%%%%%%%%%%%%%%%%%%

%%%%%%%%%%%%%%%%%%%%%%%%%%%%%%%%%%%%
%--- table of contents
% \fancyhead[RE,LO]{\sffamily {Table of Contents}}
\tableofcontents 

\clearpage{\pagestyle{empty}\clearpage}
\pagenumbering{arabic}
\setcounter{page}{1}
% \fancyhead[LE,RO]{\slshape \rightmark}
% \fancyhead[LO,RE]{\slshape \leftmark}
\pagestyle{fancy}
\fancyhf{}
\fancyhead[LE,LO]{\leftmark}
\fancyhead[RE,RO]{\rightmark}

% \pagenumbering{gobble}
\fancyfoot[CE,CO]{\thepage}

%%%%%%%%%%%%%%%%%%%%%%%%%%%%%%%%%%%%
\chapter{Introduction}

The task of Facial Emotion Recognition (FER) remains a difficult one to this day. The applications are endless across robotics, surveillance and many other human-computer interactions. The ability for an automated system to respond intelligently to emotions could spawn robots able to seamlessly adapt to their social surroundings or customer service programs capable of tailoring responses to the mood of an individual, potentially improving the relationship between man and machine significantly. Also, automated emotion recognition research can assist in identifying and flagging complex behavioural patterns such as depression \cite{acharya2018automated} \cite{zuroff1986emotion}, autism \cite{nasser2019artificial} \cite{ciaramidaro2018transdiagnostic}, spectrum disorders \cite{tagaris1} \cite{tagaris2} \cite{healthcare} and schizophrenia \cite{kim2016deep} \cite{ciaramidaro2018transdiagnostic}.

For a long time the problem was tackled using ``hand-crafted or shallow learning'' \cite{deep_fer_survey} feature extractors and then simple models applied to the output. However, with the advancement of `Deep Learning', performance on the task has accelerated in recent years \cite{training}. Modern networks achieving greater performance by learning complex mappings beyond our own comprehension \cite{dnn}. 

In the field of machine learning and computer vision there are 3 main emotion descriptor models, the most popular one being the categorical `7 basic emotions' based on the early twentieth century ``cross-culture study" \cite{deep_fer_survey} by Ekman and Friesen. However, this model is perceived to be  ``limited in the ability to represent the complexity and subtlety of our daily affective displays" \cite{deep_fer_survey}. Two alternative proposals, `Facial Action Coding System' (FACS) \cite{ekman1978facial} and the continuous `2-D valence and arousal space', are ``considered to represent a wider range of emotions" \cite{deep_fer_survey}.

The `Emotion Recognition in the Wild' (EmotiW) \cite{emotiw} challenge was first launched in 2013 and has become the standard benchmark for FER models. This paper will be focusing on the ``Audio-video based emotion recognition sub-challenge" (AV) \cite{emotiw}, building on models from previous years by applying recent deep learning techniques. The task is to classify videos with accompanying audio according to the `7 basic emotion' descriptor model mentioned above (the classes are as follows: `Angry', `Disgust', `Fear', `Happiness', `Sadness', `Surprise' and `Neutral').

This paper plans to explore multiple ways of performing the following 2-stage process: (i) extract distinct informative features from the raw data and (ii) use these features to perform accurate classification. This approach is applicable for both the visual and audio modalities. Ultimately fusing the results of both streams to produce the final predictions.

There are two ways this paper hopes to contribute to the wider research on FER. Firstly, apply recent `state-of-the-art' models (visual and temporal) to the problem and benchmark the results to previous deep learning frameworks used. Secondly, a number of other approaches are only concerned with combining model outputs in the final stage following a classic ensemble approach, where as this paper will try a few alternative ways of fusing feature maps of both modalities earlier in the network and then inputting the combined descriptor to a separate classifier.

The rest of this paper is arranged as follows. Chapter 2 summarises the background research and related works used and built upon in this paper for the FER task. Chapter 3 discusses the legal and ethical implications of this paper and future works. Chapter 4 outlines the approach taken in this paper for each stage of the pipeline and provides an explanation for certain key choices made. Chapter 5 outlines how this approach was actually carried out, including pre-processing of the data, structure of the general workflow and modelling processes involved. Chapter 6 follows the different training and evaluation stages of the project, laying out the parameter selections, reporting performance metrics and inferences to be made. Chapter 7 compares the results of this project to other entrants to the EmotiW FER AV 2018 challenge. Chapter 8 includes the final conclusions to be made based on accuracy levels reported in the previous two chapters, with final thoughts on the project as a whole. Chapter 9 lists the problems encountered during this dissertation and how they were overcome. Chapter 10 discusses future improvements to this project or completely new ideas that could possibly boost performance.

%%%%%%%%%%%%%%%%%%%%%%%%%%%%%%%%%%%%
\chapter{Background and Related Work}

This chapter begins by providing a brief overview of the different descriptor models used to characterise the expression of emotion, followed by a summary of the datasets involved in this project and how they are used for training / evaluation purposes. The data available largely informs the general approach to the task, with possible concerns and downsides of automated FER discussed. The method followed in this paper as outlined in the introduction will apply a range of classic `Deep Learning' techniques as well as a number of state-of-the-art networks, which have been detailed in order to make the explanation of the findings in this project easier to comprehend. Finally, there is a review of the proposed methods by the entrants to the 2018 EmotiW FER AV competition, which has helped to better understand what has and has not worked in the past.

\section{Human Emotion Descriptor Models} \label{emotion_tasks}

There are three main approaches used to describe the display of human emotion, (i) expression recognition, (ii) action unit detection and (ii) valence-arousal estimation. Each of these forms a sub-task that be can tackled computationally.

Recognition of a basic expression refers to the act of classifying a sentiment or feeling exhibited as one of the so-called six universal emotions (i.e. Anger, Disgust, Fear, Happy, Sad, Surprise) or the Neutral state (according to the seminal work of Ekman \cite{ekman2003darwin}). Besides typical facial expressions displayed in social communications, emotions can also manifest themselves as `Micro-Expressions' (ME). A ME is defined as a very brief and involuntary facial movement occurring in accordance with an experienced emotional state. Especially in high-stake situations, humans are likely to display MEs, despite trying to conceal or mask their true feelings (e.g. to gain an advantage or avoid some loss \cite{ekman2003darwin}). In comparison to ordinary facial expressions, a ME is very short, lasting approximately $\frac{1}{25}$ to $\frac{1}{3}$ of a second (the precise length varying in literature). Furthermore, the intensities of related muscle movements can be extremely subtle. The detection and interpretation of micro-expressions has been another area of active research. 

Detection of facial `Action Units' (AU) has also received much attention. The FACS \cite{ekman1978facial} provides a standardised taxonomy of facial muscle movements and has been widely adopted as a common approach for systematically categorising the physical manifestation of complex facial expressions. A related problem of particular interest is estimating the intensity of a particular activated AU.

Finally, valence and arousal form the axis of a 2-dimensional latent continuous space; valence indicates how positive or negative an emotional state is, whilst arousal measures the power of the emotional activation.

\section{Data} \label{Data}

\subsection{Databases}

\subsubsection{Controlled Images} \label{Controlled_Images}

The paper \cite{deep_fer_survey} lists a range of laboratory-controlled FER databases (e.g. CK+, MMI, Multi-PIE, etc.), some of these are sequences and others are groups of standalone static images. Although they may have some use in the pre-training of our visual networks, given our specific task relates to FER ``in-the-wild" and there are other datasets that meet this criteria (see the section below), these instead will form the basis of our first training stage.

\subsubsection{In-the-Wild Static Images} \label{Static_Images}

The following databases were used for training the CNN networks for feature extraction. The EmotiW challenge is an in-the-wild task to classify the 7 basic human emotion categories, so databases with real-world images have been chosen and amended to match the labelling of the EmotiW database. 

\begin{enumerate}
  
  \item \textbf{FER2013} \cite{Goodfellow2015ChallengesIR}: Database contains 48*48 pixel grayscale images of faces. Contains 35,887 images, with the following categorical breakdown: 4953 “Anger”, 547 “Disgust”, 5121 “Fear”, 8989 “Happiness”, 6077 “Sadness”, 4002 “Surprise”, and 6198 “Neutral”. Split 80\%, 10\% and 10\% across training, validation and test sets. The data is collected from a Google search for images with certain keywords
  
  \item \textbf{RAF-DB} \cite{rafdb}: The `Real-world Affective Faces Database' (RAF-DB) is a large very diverse (e.g. age, gender, ethnicity, lighting, occlusions, etc.) database of 29,672 images. As with FER2013, there are 7 emotional categories for supervised learning
  
  \item \textbf{AffectNet} \cite{Mollahosseini2019AffectNetAD}: The largest database (c.440k) available for in-the-wild facial expression images. Eleven discrete categories are defined in AffectNet as: ``Neutral, Happy, Sad, Surprise, Fear, Anger, Disgust, Contempt, None, Uncertain, and Non-face". For the purposes of this paper, only the first 7 categories were used (i.e. images for other categories were filtered out)

\end{enumerate}

Once the CNN models have been trained on the above 3 combined databases, the models are then fine-tuned on the EmotiW database discussed in the section below. 

The `Static Facial Expressions in the Wild' (SFEW) dataset could also have been used for pre-training the CNNs, but since it is a subset of the EmotiW database (700 images and only 6 emotion categories) I have opted not to use it.

\subsection{AFEW Dataset} \label{AFEW}

\begin{figure}[htbp]
    \centering
    \includegraphics[width=0.8\textwidth]{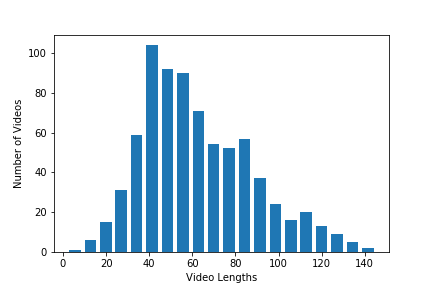}
    \caption{Breakdown of AFEW training dataset sequence lengths}
    \label{fig:train_lens}
\end{figure}

The series of EmotiW challenges have used the `Acted Facial Expression In The Wild' (AFEW) dataset \cite{Dhall2017FromIT} since it first launched in 2013. The dataset has ``vastly different environmental conditions in both audio and visual" \cite{deep_fer_survey} consisting of ``real world scenes taken from movies/television sources". In total it contains 1,809 videos ``split into three sets: training set (773 video clips), validation set (383 video clips) and test set (653 video clips)" \cite{affwild11} and ``is divided into three data partitions in an independent manner... which ensures data in the three sets belong to mutually exclusive movies and actors". The breakdown of videos by emotional classification and sequence length can be found in figures \ref{fig:train_lens}, \ref{fig:valid_lens}, \ref{fig:test_lens}, \ref{fig:train_classes} and \ref{fig:valid_classes} below. Note that the sequence lengths displayed in the graphs are based on the number of frames captured by the face detection and alignment software (see section \ref{data_process}) rather than the actual video sequence length. There is a slight difference in the length distributions between training and the validation / test datasets (more negatively skewed), the impact of this will be discussed in Chapter \ref{issues_encountered}.

\begin{figure}[htbp]
    \centering
    \includegraphics[width=0.8\textwidth]{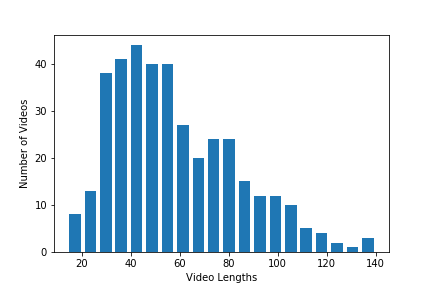}
    \caption{Breakdown of AFEW validation dataset sequence lengths}
    \label{fig:valid_lens}
\end{figure}

\begin{figure}[htbp]
    \centering
    \includegraphics[width=0.8\textwidth]{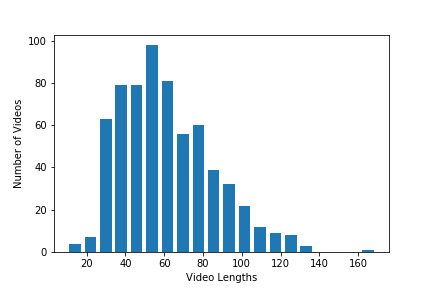}
    \caption{Breakdown of AFEW test dataset sequence lengths}
    \label{fig:test_lens}
\end{figure}

\begin{figure}[htbp]
    \centering
    \includegraphics[width=0.8\textwidth]{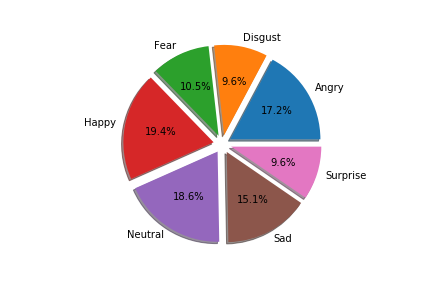}
    \caption{Breakdown of AFEW training dataset emotional classes}
    \label{fig:train_classes}
\end{figure}

\begin{figure}[htbp]
    \centering
    \includegraphics[width=0.8\textwidth]{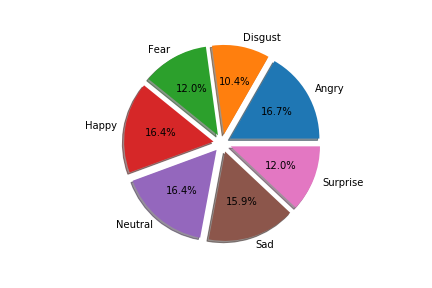}
    \caption{Breakdown of AFEW validation dataset emotional classes}
    \label{fig:valid_classes}
\end{figure}

\subsubsection{Dynamic Images} \label{Dynamic_Images}

 The AFEW videos have a `Frame Per Second' rate (FPS) of 25, which means the video is made up of static frames every 0.04 seconds. Each video in `.avi' format has been split into separate `.jpg' images according to this FPS rate, making up a dynamic sequence of images for each video. 

\subsubsection{Audio} \label{Audio_data}

The main dataset for the audio model is the raw-audio extracted from the AFEW videos mentioned above. This process involved converting each `.avi' video file into a `.wav' file (taking just the audio). 

The tool openSMILE \cite{openSMILE} (see section \ref{audio_models} for further information) calculates `Low-Level Descriptors' (LLD) features for an audio clip, the configuration applied has been built-up over time based on other audio competitions that the openSMILE software has been successful in.

\subsection{Data Augmentation}

\subsubsection{Images}

Once the base models have allowed to run on the datasets discussed above, data augmentation can be applied to images and video sequences to increase the size of the training dataset to improve model performance. Both on-the-fly and offline augmentations (e.g. perturbations, transformations, cropping, flipping, etc.) may be considered.

\subsubsection{Audio}

The common augmentation applied is varying the audio characteristics (e.g. FPS and sample rates) when extracting the raw-audio. Given the one method being explored in this paper is aligning the image frames with audio clips, augmentation options are slightly limited.

\section{Discussion} \label{discussion}

As discussed in \cite{deep_fer_survey} there are four main issues with applying deep learning models to the FER task currently. The problems are summarised below, with some solutions put forward that are further explored in the latter part of this paper:

\begin{enumerate}
  
  \item \textbf{Overfitting}: Modern deep learning models require large amounts of high quality data to accurately solve complex tasks, such as FER, on unseen in-the-wild data. The AFEW database with only 773 training videos is relatively small, increasing the importance of data augmentation, limiting the complexity of models and the pre-training of models on other comparable data sources
  
  \item \textbf{Subject Variability}: Faces of human beings vary significantly based on a number of ``personal attributes, such as age, gender, ethnic backgrounds and level of expressiveness" \cite{deep_fer_survey}, which makes it hard for a model to achieve high accuracy levels on test data. Increasing the size and variability of the datasets is important, but sometimes difficult and expensive to do. Transfer learning (e.g. training on large / varied datasets like RAF-DB and AffectNet) and multi-task learning are efficient alternate methods to ensure the model generalises well for new face types
  
  \item \textbf{Environmental Variability}: Different ``poses, illumination and occlusions" \cite{deep_fer_survey} hinders model performance because the variability is not useful information to the classification task. Pre-processing is key to improving model behaviour by standardising the visual and audio data. This allows the model to focus on only the important features. Also, increasing the quantity and quality of the datasets as mentioned above is a key consideration
  
  \item \textbf{Imbalance}: It is hard for a model to learn an expression well if it is infrequently encountered. The model will skew predictions towards more common categories as this will improve the metrics used to train the models. However, this can become an issue when applying the model to unseen data with a different class distribution. To address this problem the loss function can be amended to penalise incorrect predictions on smaller classes and data augmentation can be used to increase the amount of data for these lesser categories. Given the video sequences in the AFEW dataset, this problem is further exacerbated by a number of clips being unrelated to the actual classification (i.e. face is impassive, equivalent to neutral, for the majority of the video, with only a select few frames showing the labelled emotion).

\end{enumerate}

\section{Models and Training}

The `EmotiW: Audio-video based emotion recognition sub-challenge' is a supervised multi-class classification task, which means we are trying to learn a mapping from the input data (in this case the data is multi-modal and temporal) to a known target label. As discussed in \cite{deep_fer_survey}, deep learning has become the chosen approach in recent years due it's ability to handle the large amounts of complex data better than previous handcrafted methods. 

Machine learning problems can largely be reduced to three main decision areas and therefore the sources of error:

\begin{enumerate}
  
  \item \textbf{Functional Approximation}: Choice of model to be applied. If too simple patterns in the data will not be captured (i.e. `Underfitting'), too complex will result in `Overfitting' and if it is just not a suitable type poor accuracy will be achieved
  
  \item \textbf{Statistical Estimation}: Handling of the data to best represent the true data population and thus minimise the generalisation gap
  
  \item \textbf{Optimisation Theory}: Method applied to find the optimal parameters for the model

\end{enumerate}

\begin{figure}[htbp]
    \centering
    \includegraphics[width=0.8\textwidth]{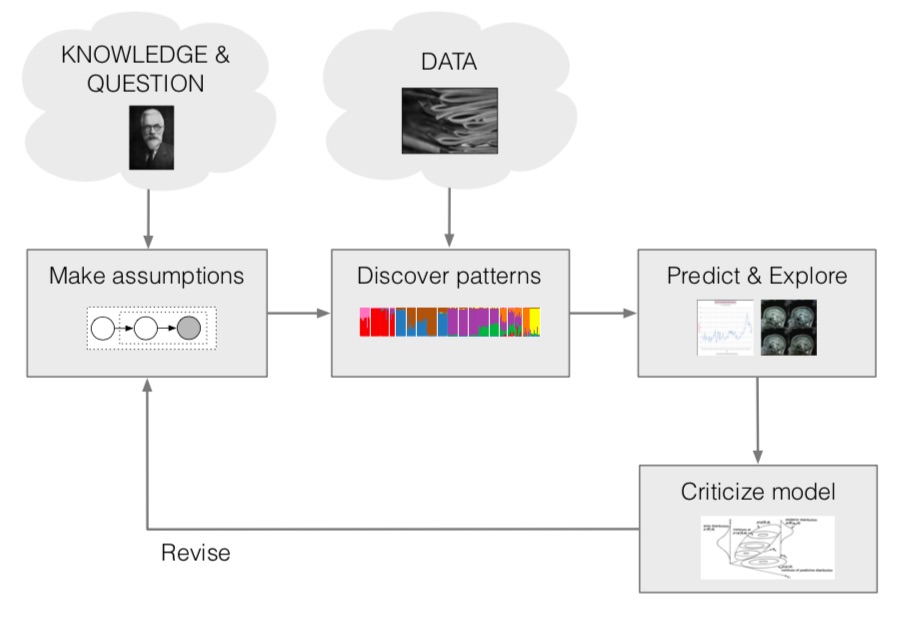}
    \caption{Machine learning model feedback loop \cite{feedback_model}}
    \label{fig:model_fb}
\end{figure}

Various choices are made for the above three areas, the model is run and feedback is provided by analysing results. Changes are then made and the process repeated until we have achieved a satisfactory outcome. This loop has been applied throughout this project and can be seen clearly in figure \ref{fig:model_fb}.

The section below sets out the basic deep learning theory, before outlining the audio and visual models being considered (i.e. functional approximation). Next we discuss the training procedure followed, including optimisation approach, loss evaluation, regularisation and hyperparameter choices (e.g. statistical estimation and optimisation theory).

\subsection{Deep Learning} \label{deep_learning_basics}

\subsubsection{Deep Feed-Forward Network}

Deep Feed-Forward Networks (FFN) are the basis of all deep learning. From this point, modifications are made to improve performance and become the modern algorithms discussed throughout the rest of this paper. The original idea is based on the structure of the human brain, with the computational version made up of perceptrons trying to recreate neurons firing.

"The goal of a feed-forward network is to approximate some 
function $f^{*}$. For example, for a classifier, $y=f^{*}(x)$ 
maps an input $x$ to a category $y$. A feed-forward network 
defines a mapping $y=f(x;\theta)$ and learns the value of the 
parameters $\theta$ that results in the best function 
approximation" \cite{goodfellow}. The aim of the training 
process is to find the optimal set of parameters $\theta^{*}$ 
that minimises the loss function, which is then our best estimate of the true function.

The early neural networks only had a single layer. Although the `Universal Approximation' theory states that `a perceptron with one hidden layer of finite width can arbitrarily accurately approximate any continuous function' \cite{deep_learning_course}, realistically this is not practical. It is difficult to know how wide the layer needs to be and the width grows exponentially with dimensions. Also, for a classification problem only linearly separable problems can be solved.

\begin{figure}[htbp]
    \centering
    \includegraphics[width=0.6\textwidth]{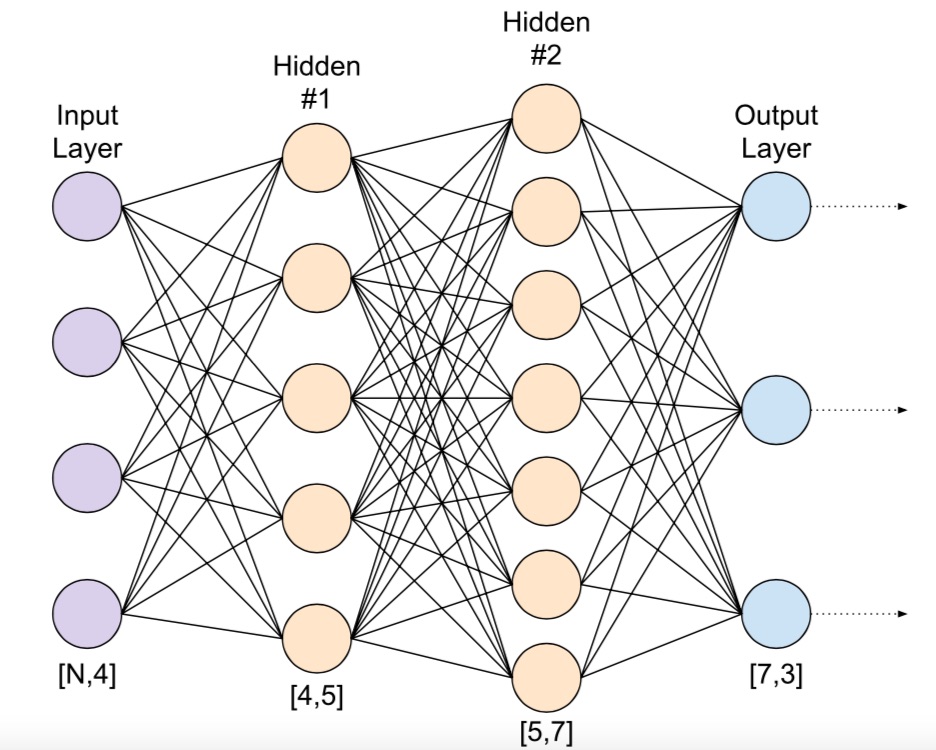}
    \caption{Deep Feed-Forward Network \cite{ann_page}}
    \label{fig:ann}
\end{figure}

The benefits of going deeper, such as in figure \ref{fig:ann}, are:

\begin{itemize}
  \item Using non-linear activation functions across many layers helps increase non-linearity of the model and therefore allows more complex functions to be approximated
  \item Typically fewer parameters are required for deeper-narrow networks than shallow-wide networks, making them easier to train
\end{itemize}

This idea of using deeper networks which can still be easily trained is at the heart of the models explored in the next section.

\subsubsection{Backpropogation}

The aim of the training process is to minimise some defined loss function for the data. To do this we apply the backpropogation algorithm. ``The input $x$ provides the initial information that then propagates up through the hidden units at each layer and finally produces output $\hat{y}$" \cite{goodfellow}, which in turn can be used to calculate our loss. 

To find $\theta^{*}$, we would like to with each iteration through the data to take a step towards our global minimum. ``The backpropagation algorithm allows the information from the loss to then flow backward through the network in order to compute the gradient" \cite{goodfellow} with respect to any parameter by the use of the chain rule. The negative gradient is the direction of steepest descent, hence a move in this direction will result in a decrease in the loss function. When the backpropogation algorithm is applied to a batch of data with the below update rule, it is called `Stochastic Gradient Descent':

\begin{equation} \label{eq:learning}
\theta^{(t+1)} = \theta^{(t)} - \alpha^{(t)}\nabla{}_{\theta}l(f(x_{t};\theta),y_{t})
\end{equation}

In the above equation \ref{eq:learning}, $\alpha$ represents the learning rate, which controls the step-size taken by the algorithm in the direction of steepest descent. The algorithm will update parameters recursively after every iteration (i.e. batch) until convergence or the set maximum number of iterations is reached.

The batch-size influences the training process as well:

\begin{itemize}
    \item \textbf{Small Batches}: Noisy gradient because fewer samples, so less likely to reflect the true direction of steepest descent, but the additional noise can help escape poor local minima
    \item \textbf{Large Batches}: More memory intensive to implement, but gradient likely to be a better representation of the true direction of steepest descent
\end{itemize}

\subsubsection{Activation Functions}

\begin{figure}[htbp]
    \centering
    \includegraphics[width=1.0\textwidth]{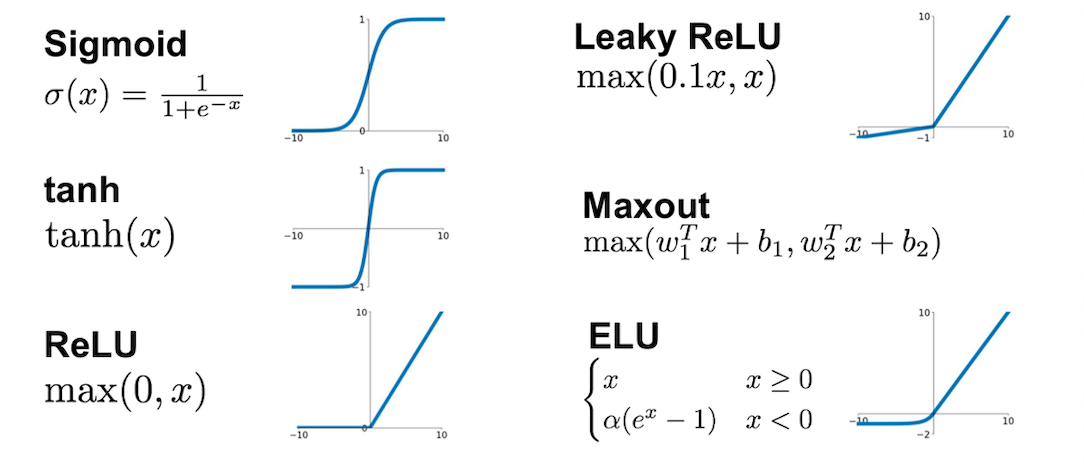}
    \caption{Summary of most common activation functions \cite{activation_image}}
    \label{fig:activation}
\end{figure}

There are multiple activation functions used throughout the machine learning field, graphs and equations for the most common activation functions can be seen in figure \ref{fig:activation}. Each example is non-linear which helps the model capture complex relationships in the data. The two I will focus on are the sigmoid function and Rectified Linear Unit (ReLU), both commonly used in the area of computer vision and classification problems:

\begin{itemize}
    \item The sigmoid function is often used in binary classification problems (can be interpreted as a probability), but also as a gating function (see Squeeze-and-Excitation, LSTM and GRU models). It is expensive to compute (due to being an exponential function), is non-centred and can often lead to vanishing gradients during training, hence it is not used in hidden layers
    \item The ReLU function is cheaper to compute and better at letting gradients flow backwards. Other variants such as Leaky ReLU and ELU address the issue of ReLU being non-centred and having 0 gradient for negative values, but most cutting-edge architectures (like those explored in this paper) still prefer to use the ReLU function
    
\end{itemize}

\medskip

The EmotiW challenge is a multi-class classification problem, with 7 different emotional categories. The ``softmax" function converts the logit output for each class in the final layer into an interpretable probabilistic output (similar to the sigmoid function for the binary problem) and so is useful when applied in the final layer.

\begin{equation} \label{softmax}
    softmax(a_{i})=\frac{\exp{a_{i}}}{\sum_{j=0}^{k}\exp(a_j)}
\end{equation}

\subsection{Visual Models} \label{visual_models}

Applying machine learning to images remained a difficult problem for many years. Compared to data in tabular form (i.e. a spreadsheet with m features), even a small colour image of 96x96x3 pixels has 27,648 inputs per sample. If the network is fully connected, the number of neurons becomes unmanageable memory-wise and extremely difficult to train deep networks, this is an example of the ``curse of dimensionality" \cite{deep_learning_course}. As explained below, Convolutional Neural Networks (CNNs) and it's variants help to address these problems and can be applied end-to-end. A video is just a sequence of individual images and hence the below models can be used to extract useful features and linked to capture the temporal dimension as seen in the next section.

\subsubsection{Convolutional Neural Networks} \label{CNNs}

\begin{figure}[htbp]
    \centering
    \includegraphics[width=0.8\textwidth]{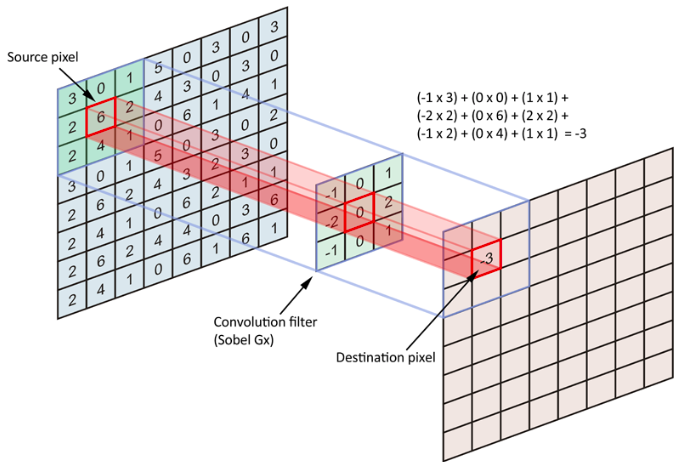}
    \caption{2-D Convolution operation \cite{convolution_page}}
    \label{fig:convolution}
\end{figure}

A key advantage of CNNs is there ability to reduce the number of parameters, this is mainly done through weight sharing and sparse connectivity to create feature maps. The mechanism used, taken from signal processing, is a convolution as seen in figure \ref{fig:convolution}. 
The convolution is essentially a filter on the input signal, for example a 3x3 convolutional kernel is applied (multiplied element-wise and then summed) to the 3x3 pixel input window. The kernel is passed over the whole image input to give a single feature map output.

Mathematically a discrete 2-D convolution is represented in equation \ref{conv}, where I is the 2-D image (current source pixel is $(i,j)$) and K is our 2-D kernel of height m and width n.  

\begin{equation} \label{conv}
    S(i,j) = (I*K)(i,j) = \sum_{m}\sum_{n}I(m,n)K(i-m,j-n)
\end{equation}

Depending on the weights in the kernel, different features will be learned, in figure \ref{fig:convolution} the horizontal Sobel filter is being applied. In this case, changes in horizontal pixel intensity will result in a large output value, indicating a vertical edge at this point of the input. 

\begin{figure}[htbp]
    \centering
    \includegraphics[width=0.6\textwidth]{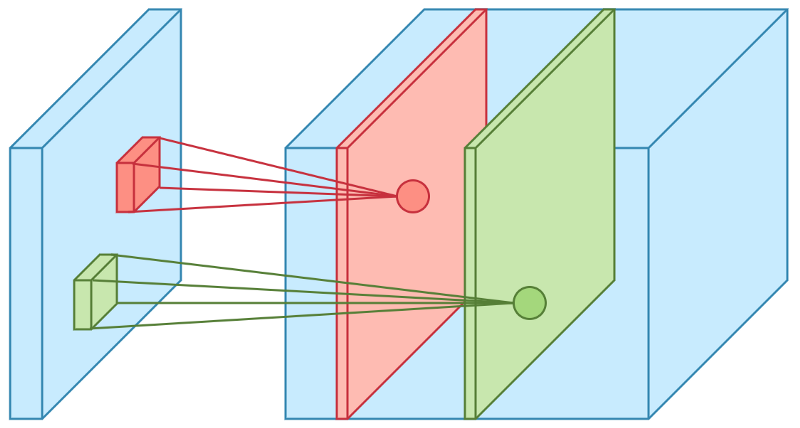}
    \caption{Multiple convolutions applied to create different feature maps \cite{cnn_page}}
    \label{fig:feature_maps}
\end{figure}

Multiple convolutions are applied to the same input image, with the resultant feature maps stacked to create a convolutional layer, as seen in figure \ref{fig:feature_maps}. The CNN network learns these weights through the backpropogation algorithm, picking up specific features according to the data structure. 

Since the weights in each convolutional kernel remain the same (i.e. weight sharing) and the kernel size is only applied to a section of the image (i.e. sparse connectivity) the number of parameters is vastly reduced. For example, if 64 3x3x3 convolutional kernels are applied to our 96x96x3 image input (including bias terms), there would be 1,792 parameters to learn (64*(3*3*3+1)) (i.e. $O(k \times n)$), for a fully connected layer of 64 neurons there would be 1,769,536 parameters to learn ((96*96*3 + 1) * 64) (i.e. $O(m \times n)$) and the output would be less useful given its lack of spatial information. 

An activation function is applied to each neuron in the feature map output to help increase non-linearity to allow the model to learn more complex functions.

Currently a neuron in the first convolutional layer can only see a small section of the input image, to allow hierarchical learning we want to increase the receptive field (size of the image a neuron is exposed to) throughout the network. This allows the model to learn low-level features in the early layers (i.e. edges, circles, textures, etc.) and high-level features in the latter layers (i.e. faces, animals, buildings, etc.). The most common technique used to do this is `pooling', which ``replaces the output of the net at a certain location with a summary statistic of the nearby outputs" \cite{goodfellow}. Two types of pooling are often applied, `max pooling' and `average pooling'. There are no parameters for the model to learn in a pooling layer, so no complexity is added.

\begin{figure}[htbp]
    \centering
    \includegraphics[width=1.0\textwidth]{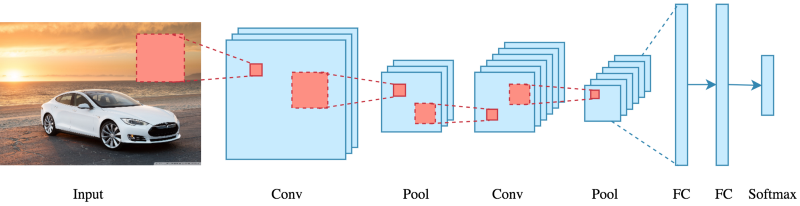}
    \caption{A typical CNN structure \cite{cnn_page}}
    \label{fig:cnn}
\end{figure}

A classical CNN network comprises of blocks containing a convolutional layer followed by a pooling layer as seen in figure \ref{fig:cnn}. For classification tasks, it is common to have 1 or more fully connected layers at the end once the number of neurons has been sufficiently reduced. This part of the network has been proven to be efficient at learning the complex mapping from the flattened final feature maps to the target output. 

Two further benefits of the CNN architecture are they help to solve the (i) shift invariance ($f(\mathbf{x}) = f(Sx)$) and (ii) shift equivariance ($Sf(\mathbf{x}) = f(S\mathbf{x})$) problems (for some shift S). This is a major issue for feed-forward networks, where shifting the image only slightly (key for video sequences) would severely impact the output. It is an important result of CNNs that after shifting a person slightly in the frame, the network can still recognise that there is still a person present (and even recognise that it is the same human being). 

The models discussed in the following sections have the same basic building blocks as explained above, but due to certain amendments are able to go deeper more efficiently and therefore achieve improved performance.

\subsubsection{VGG \cite{VGG}} 

\begin{figure}[htbp]
    \centering
    \includegraphics[width=0.6\textwidth]{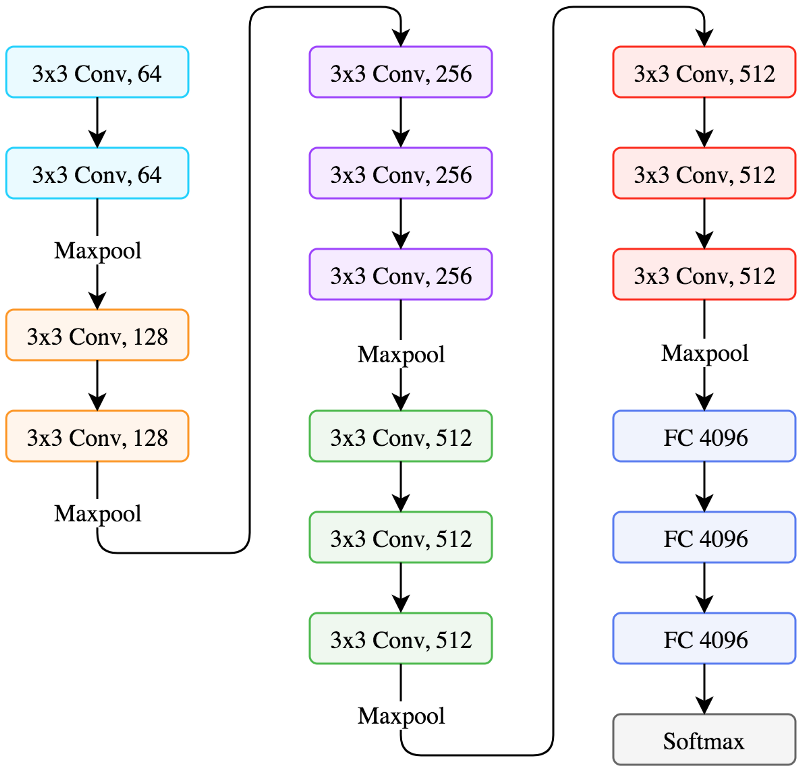}
    \caption{The VGG architecture \cite{cnn_page}}
    \label{fig:VGG}
\end{figure}

The VGG-Face model is a 16 layer (excluding pooling and softmax layers) architecture published in 2014 (see figure \ref{fig:VGG} for overview) that achieves impressive results despite it's relative simplicity compared to current state-of-the-art models. The network only uses 3x3 convolutional kernels (unlike other models such as LeNet \cite{lecun} and AlexNet \cite{alex_net}), which `reduces the number of parameters and showed that using consistent filter sizes improves performance' \cite{deep_learning_course}. 

The model was originally trained for ``face identification and verification" \cite{VGG}, using ``triplet-loss" \cite{VGG} for this purpose. The research team created their own dataset of 2.6m facial images of 224x224 pixels, with further data augmentation applied. In this paper, we have used the pre-trained VGG-Face model, but fine-tuned on the datasets mentioned in section \ref{Data}.

The network by stacking convolutional layers (in twos and threes) allows ``the use of two ReLu operations, and more non-linearity gives more power to the model" \cite{cnn_page}. Dropout is applied to the first two FC layers for regularisation purposes.

\subsubsection{ResNet \cite{resnet_paper}}

The ResNet model first introduced in 2015 and allowed CNN models to go deeper than ever before. There are two problems with training incredibly deep networks, (i) getting a meaningful loss result on the feed-forward loop as ``accuracy gets saturated and then degrades rapidly" \cite{resnet_paper} with depth and (ii) propagating that loss back through the network to then learn efficiently. 

Early in the training process, the weights are near their initialised values and thus small in magnitude. In the forward pass the response to the input decreases through the network until it vanishes and in the backward pass the gradient eventually tends to 0 (i.e. vanishing gradient problem), making it extremely difficult for the network to learn. In fact, by adding layers to most standard deep CNN models performance will actually decrease.

\begin{figure}[htbp]
    \centering
    \includegraphics[width=0.8\textwidth]{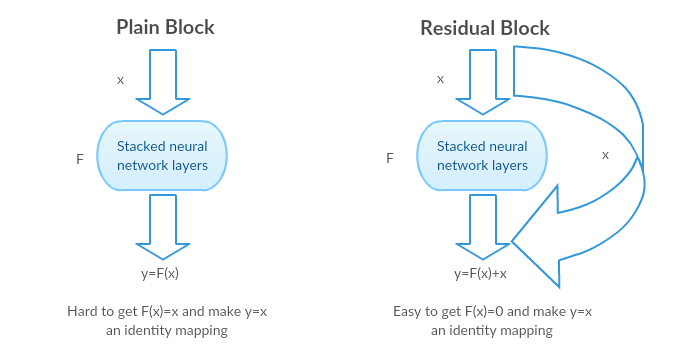}
    \caption{Example of skip connection \cite{resnet_image}}
    \label{fig:skip}
\end{figure}

\begin{figure}[htbp]
    \centering
    \includegraphics[width=0.8\textwidth]{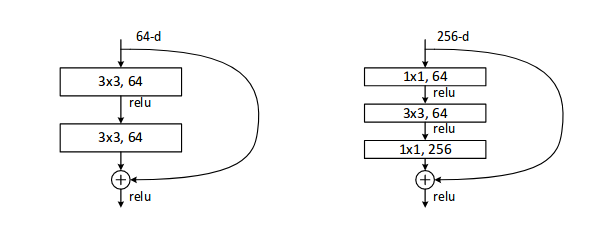}
    \caption{Example of ResNet block \cite{resnet_image}. Note the Bottleneck class implements a 3 layer block and Basicblock class implements a 2 layer block, the former is used in this paper}
    \label{fig:resnet_block}
\end{figure}

The solution proposed was to include skip connections as seen figure \ref{fig:skip}. If the input signal is very weak, the input is still carried forward through the identity mapping and added to the output of the stacked convolutional layers (e.g. two or three convolution operations plus ReLu activation function, as seen in figure \ref{fig:resnet_block}). For the mathematical representation see equation \ref{res_eq}. This means that the input or the loss can be propagated much further forward or backwards respectively through the network, making training significantly easier.

\begin{equation} \label{res_eq}
    \mathbf{y} = F(\mathbf{x}, W_{i}) + \mathbf{x}
\end{equation}

Pooling is still applied within the model, at certain points a modified skip connection is utilised to ensure the dimensionality of the input and output of the stacked convolutional layers match.

\begin{figure}[htbp]
    \centering
    \includegraphics[width=1.0\textwidth]{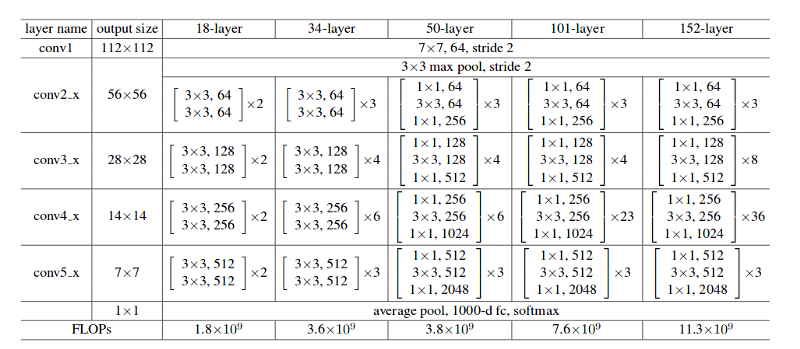}
    \caption{Example of ResNet architectures \cite{resnet_image}}
    \label{fig:resnet_structures}
\end{figure}

As previously mentioned, the VGG model had 16 layers, but with the introduction of skip connections ResNet models were able to achieve unrivalled performance for 50+ layers as detailed in figure \ref{fig:resnet_structures}. In this project, due to the smaller dataset I am using the ResNet 50-layer model, which has been pre-trained on the ImageNet dataset (achieving top-5 error rates of 5.25\% vs. 8.43\% for VGG \cite{resnet_paper}).

\subsubsection{ResNeXt \cite{resnext_paper}}

\begin{figure}[htbp]
    \centering
    \includegraphics[width=0.8\textwidth]{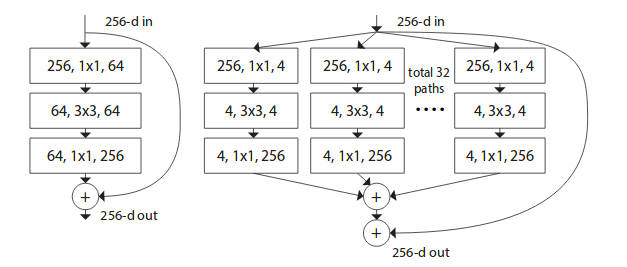}
    \caption{Comparison between ResNet and ResNext blocks \cite{resnext_paper}}
    \label{fig:resnext_block}
\end{figure}

The ResNeXt model follows a similar set-up to the ResNet model, with the main difference being within the convolution blocks used (see figure \ref{fig:resnext_block}) that exploit a ``split-transform-merge strategy" \cite{resnext_paper}. The ResNeXt paper introduces the idea of `cardinality' (``the size of the set of transformations"), where each block has a ``multi-branch architecture" \cite{resnext_paper}. The more branches within the block the higher the cardinality, which the paper argues is ``more effective than going deeper or wider" and hence limits complexity.

The transformation as outlined on the RHS of figure \ref{fig:resnext_block} has cardinality of 32 (and width 4), which has roughly the same number of parameters as the ResNet block (cardinality 1 but width 64). Equation \ref{resnext_eq} shows the aggregation of the new transformations $T_{i}$. 

\begin{equation} \label{resnext_eq}
    \mathbf{y} = \sum_{i=1}^{C} T_{i}(\mathbf{x}) + \mathbf{x}
\end{equation}

\begin{figure}[htbp]
    \centering
    \includegraphics[width=0.7\textwidth]{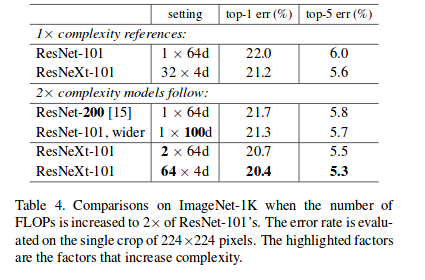}
    \caption{Comparison between ResNet and ResNext performance \cite{resnext_paper}}
    \label{fig:resnext_results}
\end{figure}

As with the ResNet model above, a pre-trained (on ImageNet dataset) version of the ResNext 50-layer is used in this project, with the added mechanism of squeeze-and-excitation (see relevant subsection below). A comparison between ResNet and ResNext performance can be seen in figure \ref{fig:resnext_results}.

\subsubsection{DenseNet \cite{densenet_paper}}

The `Densely Connected Convolutional Network' (DenseNet) model was published in 2016. It uses multiple connections (``connects each layer to every other layer", i.e. $\frac{L(L+1)}{2}$ direct connections enabling feature reuse) to allow further deepening of the convolutional networks by ``improving information flow forwards and gradient flow backwards" (direct access to the input and loss function). The paper argues that this dense connectivity actually results in fewer parameters ``as there is no need to re-learn redundant feature maps" (i.e. some ``ResNet layers do not really contribute" and so could be removed) and has a regularising effect. Also, the narrowness of DenseNet (i.e. number of filters in each block) helps reduce complexity. 

\begin{figure}[htbp]
    \centering
    \includegraphics[width=0.8\textwidth]{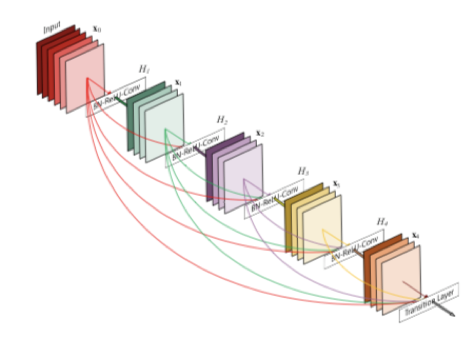}
    \caption{DenseNet of 5 layers \cite{densenet_paper}}
    \label{fig:desnenet}
\end{figure}

As seen in figure \ref{fig:desnenet}, each layer (`dense block') has access to all previous layers (on the forward pass) by concatenating them and then applying the composite function $H_{l}(.)$. The concatenation results in a `growth rate' (k) in the number of feature maps per dense block, i.e. $k = 12$, which is still fewer channels than other network architectures. This narrowness can be increased further by compressing the feature maps (based on compression hyperparameter $0<\theta<1)$.

\begin{equation}
    \mathbf{y}_{l} = H_{l}([\mathbf{x}_{0},...,\mathbf{x}_{l-1}]) 
\end{equation}

Pooling is once again utilised within the network. To ensure that the dimensions match, down-sampling between the dense blocks is required to change the size of the previous layers. These transformations are called `transition layers', which consist of a convolution and pooling. This allows the model to learn which feature maps are most important.

The benefits of the DenseNet model are the improved flow of information and through the dense connections access for the model to ``collective information" (i.e. all feature maps). Performance on the CIFAR10 dataset (with data augmentation) shows a loss rate of 4.51\% vs. 6.61\% for the ResNet model.

\subsubsection{Squeeze-and-Excitation \cite{squeeze_excitation}}

Squeeze-and-Excitation (SE) is not a network, but a block to be included in other CNN networks to improve the ``quality of spatial encodings throughout its feature hierarchy" \cite{squeeze_excitation} that improves channel inter-dependencies. It provides a significant boost to existing CNN architectures with minimal additional complexity.

The convolutional operation within a CNN has the effect of ``fusing the spatial and channel information" of an image, essentially applying a 2D-filter to each channel in the previous layer and weighting the outputs equally (i.e. summed together). Instead SE integrates learning mechanisms within this operation to reflect the relative importance of the different feature maps (i.e. ``selectively emphasise informative features and suppress less useful ones" \cite{squeeze_excitation}). 

\begin{figure}[htbp]
    \centering
    \includegraphics[width=1.0\textwidth]{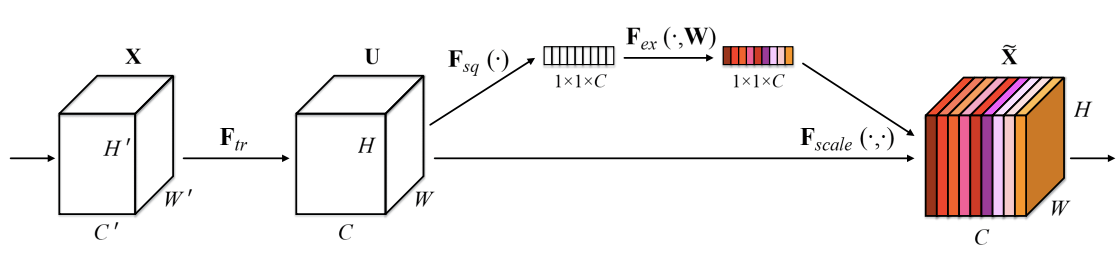}
    \caption{SE block process \cite{squeeze_excitation}}
    \label{fig:SE_image}
\end{figure}

\begin{figure}[htbp]
    \centering
    \includegraphics[width=0.6\textwidth]{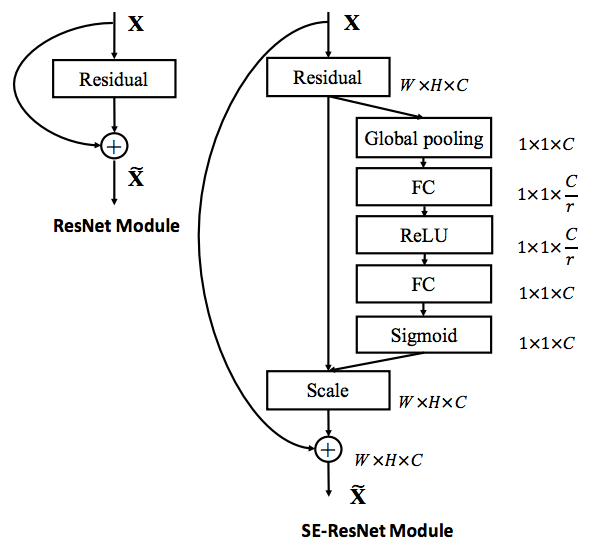}
    \caption{Squeeze operation in greater detail \cite{squeeze_excitation}}
    \label{fig:SE_op}
\end{figure}

The SE block is made up of the `squeeze' ($F_{sq}$ in figure \ref{fig:SE_image}) operation, which takes the input $(H \times W \times C)$ and transforms them into a channel-wise embedding $(1 \times 1 \times C)$ using ``global average pooling to generate channel-wise statistics". This is followed by the `excitation' operation ($F_{ex}$ in figure \ref{fig:SE_image}) that ``takes the embedding as an input and produces a collection of per-channel modulation weights" through a simple gating mechanism (seen in figure \ref{fig:SE_op}). These weights represent the relative importance (similar to an attention mechanism) and when applied to the feature maps generates the SE block output.

The SE block can be applied at different depths, in earlier layers it ``excites informative features in a class-agnostic manner" and in later layers ``the SE blocks become increasingly specialised, and responds to different inputs in a highly class-specific manner". The squeeze and excitation operations are ``very computationally lightweight".

The SE block can be integrated into many standard CNNs, but for the purposes of this paper the mechanism has been applied to the ResNet and ResNext models. In both instances, SE-ResNet 50-layers and SE-ResNeXt 50-layers, were first trained on the ImageNet dataset and then fine-tuned for FER. The benefit of including the SE block on the top-5 error was ~0.9\% and ~0.4\% respectively \cite{squeeze_excitation}.

\subsubsection{NASNet \cite{nasnet_paper}}

The Neural Architecture Search (NAS) is an automated machine learning algorithm that searches (i.e. trial and error) for the best neural net architecture based on the data. This ``reduces the need for architectural engineering" \cite{nasnet_paper} and the one-size fits all approach. However, the search itself is computationally expensive on large datasets but the models created are often computationally efficient relative to the performance (i.e. an eager process).

\begin{figure}[htbp]
    \centering
    \includegraphics[width=0.8\textwidth]{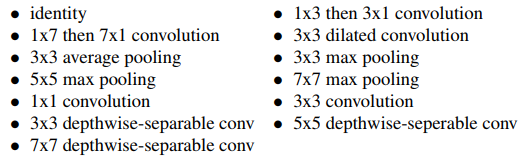}
    \caption{NASNet building blocks  \cite{nasnet_paper}}
    \label{nasnet_blocks}
\end{figure}

The paper outlines the NASNet search space, which takes pre-trained (on the CIFAR-10 database) building blocks (see figure \ref{nasnet_blocks}) and uses these to design the optimal convolutional networks. This limits the search to finding the best cell combinations (i.e. those of different weights) rather than the best cell structures, which increases generalisability and improves search times. 

There are two types of convolutional cells included in the search space: (i) ``convolutional cells that return a feature map of the same dimension" and (ii) ``convolutional cells that return a feature map where the feature map height and width is reduced by a factor of two" (i.e. uses a stride of 2).  

\begin{figure}[htbp]
    \centering
    \includegraphics[width=0.8\textwidth]{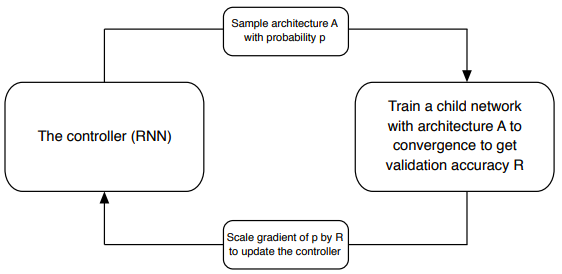}
    \caption{Overview of Neural Architecture Search process \cite{nasnet_paper}}
    \label{nasnet_loop}
\end{figure}

The controller of the search is a Recurrent Neural Network (RNN) architecture. It trains a child network to convergence to get a validation accuracy, this performance is fed back to the controller to find ``better architectures over time". The process loop can be seen in figure \ref{nasnet_loop}. To help inform the decisions of the controller, reinforcement learning can be used, as opposed to just random search.

The resultant architectures ``resemble current state-of-the-art networks", but the NAS algorithm is able to ``find interesting connections" specific to the task at hand. This is helpful for this paper as a few of the cutting edge models described above have not been applied to the FER task directly. The problem remains having the time and computational power to conduct the full search to find the optimal network, especially on large datasets. 

\begin{figure}[htbp]
    \centering
    \includegraphics[width=1.0\textwidth]{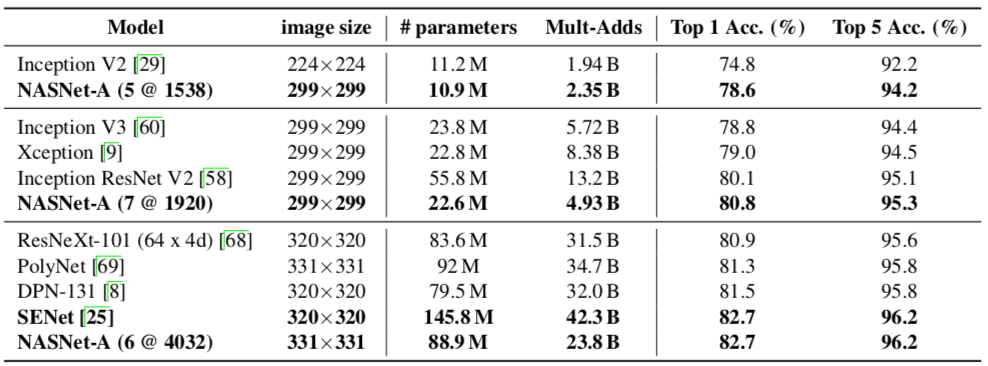}
    \caption{Overview of Neural Architecture Search results \cite{nasnet_paper}}
    \label{nasnet_results}
\end{figure}

In figure \ref{nasnet_results} we can see the performance versus a number of the models mentioned above. On the ImageNet dataset it beats the ResNeXt model and draws with the SENet model, but has far fewer parameters than the latter.

\subsection{Sequence Models} \label{sequence_models}

Convolutional networks have been proven to be very effective when applied to static images, but not as good at dealing with sequences (e.g. a video) where the current frame is dependent on the frames surrounding it. 3D-CNNs and linking CNNs sequentially are possibilities, but both options are computationally expensive and still not very good at modelling these temporal dependencies. 

The solution is to use `recurrent cells' to form a Recurrent Neural Network (RNN). RNNs are able to store information (e.g. have a memory), which can be passed through time and used to inform decisions.

\subsubsection{RNN}

RNNs are very flexible networks, able to deal with different sequential input to output scenarios. In this paper, we will focus on the input being a sequence of frames, with an output for each recurrent cell being a feature map (i.e. a many-to-many mapping). The feature maps are then passed through a fully connected layer to get a single classification of the 7 basic emotions per frame (see section \ref{seq_model_imp}). 

\begin{figure}[htbp]
    \centering
    \includegraphics[width=0.8\textwidth]{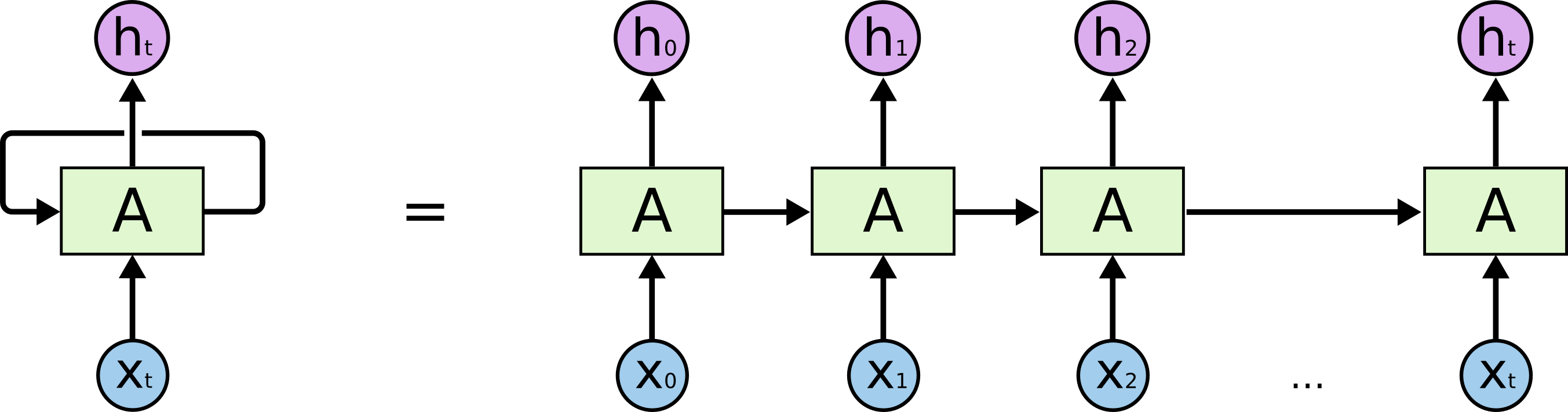}
    \caption{Unrolled recurrent cell \cite{RNN_images}}
    \label{RNN_unrolled}
\end{figure}

On the LHS of figure \ref{RNN_unrolled} is the `direct feedback recurrent cell', which when unrolled shows the temporal element clearly (e.g. input and output at each time step). There are two inputs to each cell (i) the new input $x_{t}$ and (ii) the output of the previous recurrent cell $h_{t-1}$. The use of (ii) allows the model to have a memory, this means that the final cell in the sequence will have some recollection of the first input (i.e. repeat application of equation \ref{rnn_eq_mem}). 

The typical form for an RNN cell output can be seen in equation \ref{rnn_eq_form}, with the same weights and function used at every time step. The output of the cell can then be used to give the output of the network for that frame as seen in equation \ref{rnn_eq_out}.

The network has relatively few weights to learn (compared to a CNN), with $W_{hh}$ and $W_{xh}$ controlling what information is used from the memory and the input respectively. 

\begin{equation} \label{rnn_eq_mem}
    h_{t} = f_{w}(h_{t-1}, x_{t})
\end{equation}{}

\begin{equation} \label{rnn_eq_form}
    h_{t} = tanh(W_{hh}h_{t-1} + W_{xh}x_{t})
\end{equation}{}

\begin{equation} \label{rnn_eq_out}
    y_{t} = W_{hy}h_{t}
\end{equation}{}

The backpropogation algorithm is once again used for training. However, vanishing / exploding gradients are a major problem for RNNs. Taking the gradient of the loss with respect to the parameters is done for the whole sequence, which means repeatedly taking the derivative of equation \ref{rnn_eq_form}. Applying the chain rule results in matrix multiplication of the weights T times, where T is the length of the whole sequence. Hence if $W_{hh} > 1$ we have exploding gradients and $W_{hh} < 1$ we have vanishing gradients for very long sequences.

Also, RNNs have a fairly simple memory mechanism, which makes modelling long-term complex dependencies difficult. This prompted work on the LSTM and GRU networks discussed below, which with improved memory control can help solve this problem and allow for better information flow to address the vanishing / exploding gradient issue.

\begin{figure}[htbp]
    \centering
    \includegraphics[width=0.8\textwidth]{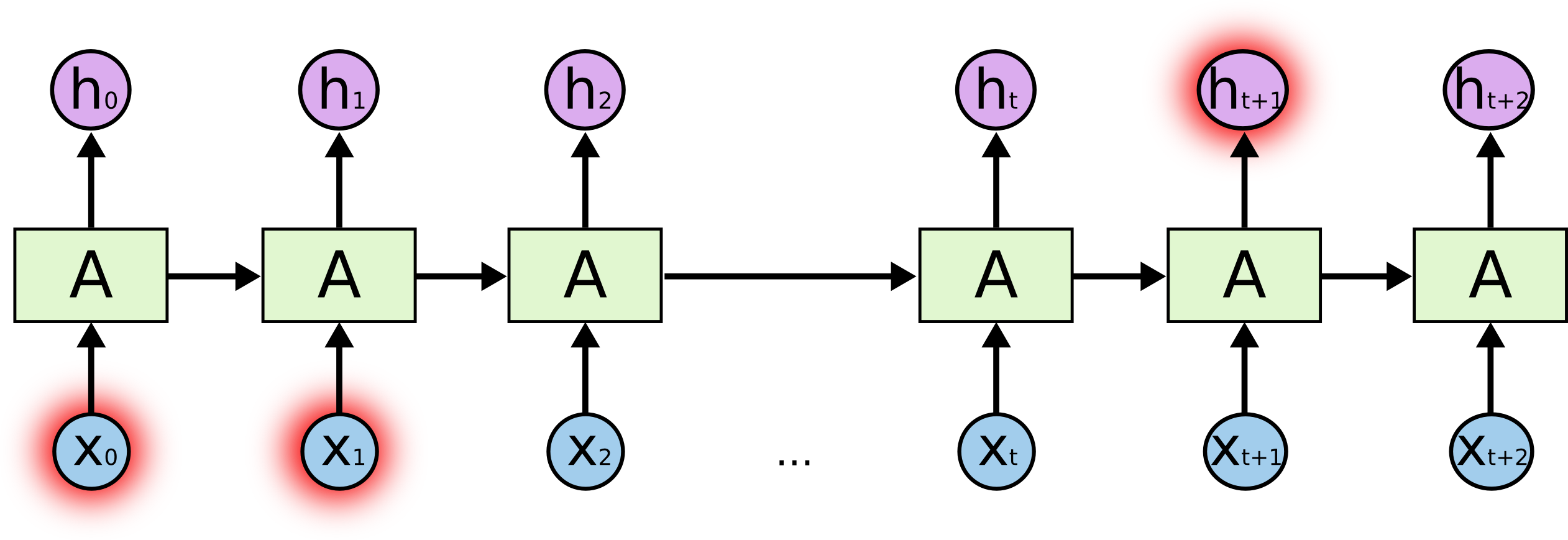}
    \caption{Long term dependencies in an RNN \cite{RNN_images}}
    \label{fig:RNN_longterm}
\end{figure}

Feeding the video frames directly into an RNN would achieve poor performance, because the simple network would not be able to handle all the information provided. Instead we use the CNNs discussed in the above section as feature extractors, with the final layer (essentially an embedding for the image) used as an input to the RNN. This helps combine the power of CNNs at handling images with the ability of RNNs to model temporal relationships well. 

\begin{figure}[htbp]
    \centering
    \includegraphics[width=0.6\textwidth]{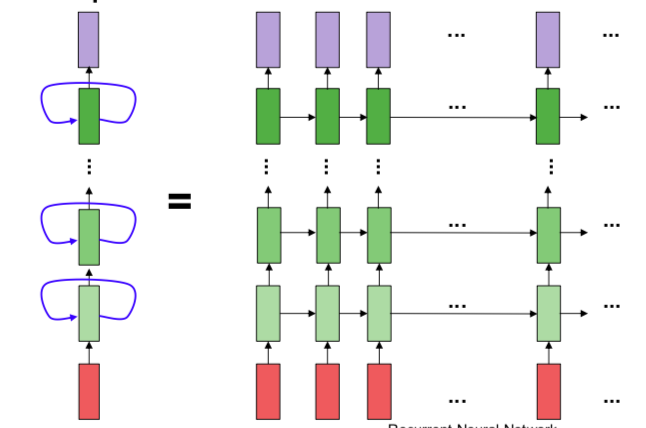}
    \caption{Overview of deep RNN \cite{RNN_images}}
    \label{fig:RNN_deep}
\end{figure}

Deeper networks are able to capture more complex mappings. This approach can also be applied to RNNs by stacking layers on top of each other, with the output from one layer feeding into the cell in the layer above as well as the cell for the next time step (see figure \ref{fig:RNN_deep}).  

\subsubsection{Bi-Directional RNNs}

\begin{figure}[htbp]
    \centering
    \includegraphics[width=1.0\textwidth]{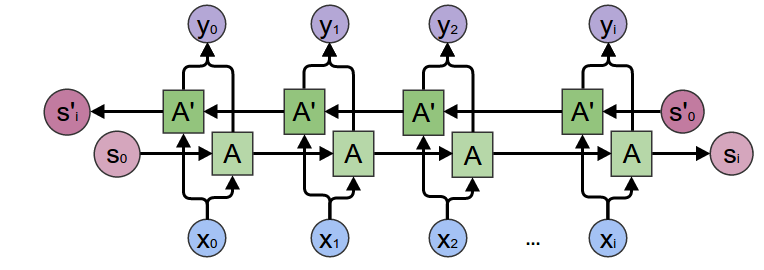}
    \caption{A Bi-Directional RNN structure \cite{RNN_images}}
    \label{fig:bi_direct}
\end{figure}

We have already mentioned that RNNs are better than other neural networks at handling forward temporal context, but this can be taken a step further. Bi-directional RNNs have two hidden layers (forward and backward) to learn both past and future context. The structure can be seen in figure \ref{fig:bi_direct}, where any output $y_t$ has access to information at $t-s$ and $t+s$ (for any $s$). 

Although this technique cannot be applied in real-time (only historical data accessible), where the all information is known beforehand (i.e. whole video available) the additional context can help boost performance. 

\subsubsection{LSTM \cite{lstm}}

The Long Short-Term Memory (LSTM) network has control gates to help manage the flow of new and historical information. There are three gates to consider \cite{deep_learning_course}:

\begin{itemize}
    \item \textbf{input gate}: what and how much new information to write to memory
    \item \textbf{forget gate}: what and how much of the memory to erase / keep
    \item \textbf{output gate}: what and how much information to filter from the output of the cell
\end{itemize}

\begin{figure}[htbp]
    \centering
    \includegraphics[width=1.0\textwidth]{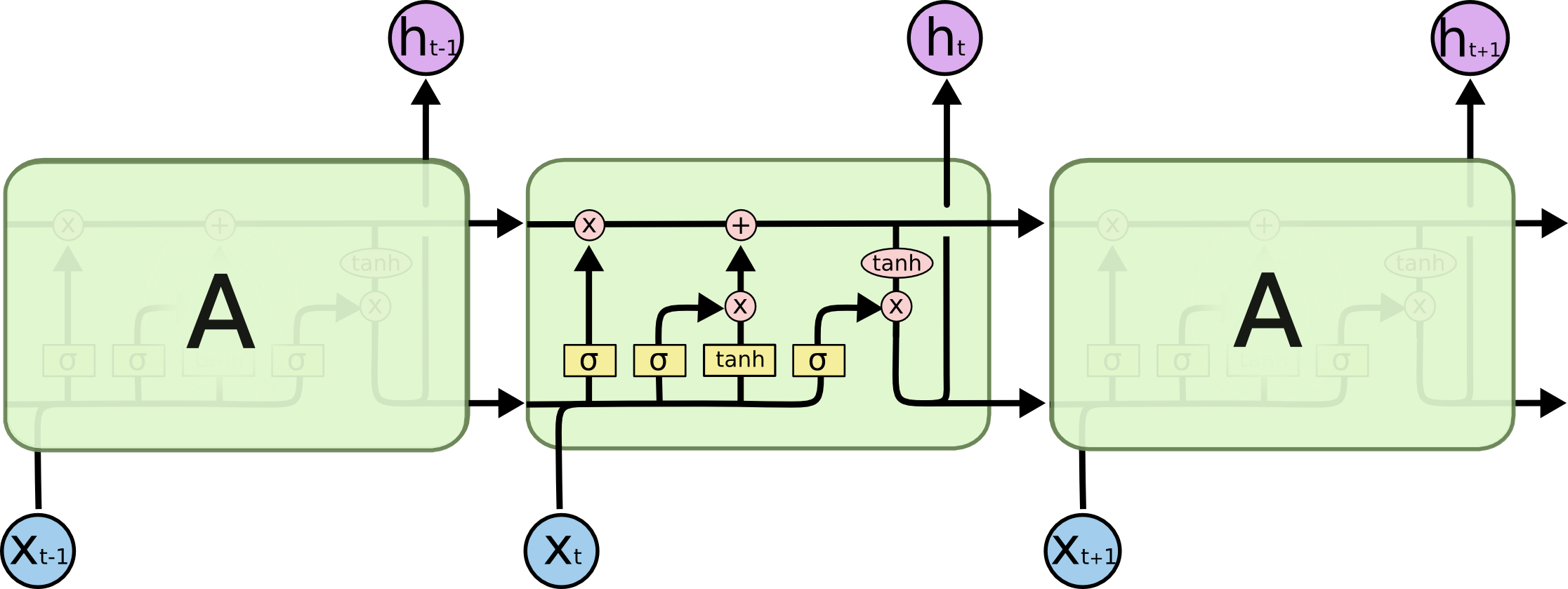}
    \caption{Overview of the LSTM structure and LSTM cell \cite{RNN_images}}
    \label{fig:LSTM}
\end{figure}

Each gate has it's own weight that is learned by the model and the amount of information allowed to pass / erased is a function of the input and previous cell output. The cell state $c_t$ is the memory of the network at any given time step. See figure \ref{fig:LSTM} for a clearer representation of how the gates, inputs and outputs interact.

\subsubsection{GRU}

\begin{figure}[htbp]
    \centering
    \includegraphics[width=1.0\textwidth]{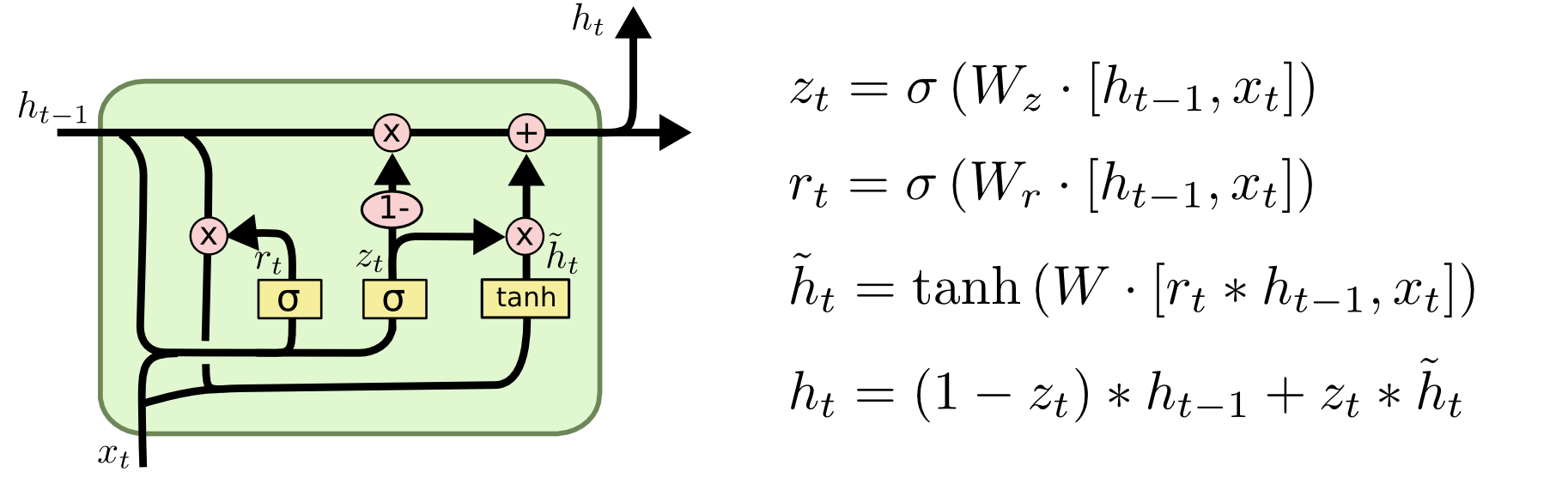}
    \caption{Overview of the GRU structure and calculations involved \cite{RNN_images}}
    \label{fig:GRU}
\end{figure}

The Gated Recurrent Unit (GRU) network follows on from the LSTM model, using the idea of control gates to influence the memory of the network. As can be seen in \ref{fig:GRU}, there are fewer control gates (only 2) and therefore fewer parameters to learn. This is particularly helpful when dealing with smaller datasets such as that provided in the EmotiW 2019 challenge.

Performance between ``GRUs and LSTM is comparable" \cite{gru_comp} and problem dependent, but because the former has fewer parameters it is typically easier to train.

\subsection{Audio Models} \label{audio_models}

Three different approaches were considered, summaries of these are included below:

\begin{enumerate}

    \item \textbf{Raw Form}: Apply a CNN + RNN architecture or pre-trained state-of-the art model SoundNet \cite{soundnet}
    
    \item \textbf{Spectrograms}: A transformation of the audio signal into the dimensions: amplitude, frequency and time (essentially a feature map). A sequential neural network, such as a LSTM or GRU, can then be applied to the spectrogram to give a classification for the audio clip 
    
    \item \textbf{openSMILE \cite{openSMILE}}: An open-source ``toolkit for flexible feature extraction for signal processing and machine learning applications". Depending on the approach taken, a FFN or RNN model will use these features to give the final output
    
\end{enumerate}

The focus of this section is on openSMILE for the reasons outlined in Chapter \ref{design}. The tool has a range of processing and feature output options for audio, with proven performance in a range of audio challenges. The user can either design their own configuration files from scratch according to their specific task, use an existing setup previously applied in a competition by the openSMILE team or amend one of these existing configurations. An example configuration file is given in the accompanying code files.

The standard feature output of openSMILE is calculated as follows:

\begin{enumerate}
    \item Evaluate Low-Level Descriptors (LLDs) for the raw-audio (see Appendix A for list of common LLDs) for windows starting at certain step sizes 
    \item Differential of the LLDs across the audio file
    \item Statistical analysis (e.g. mean, moments, regressions, transformations, etc.) of the LLDs and their differentials to give the `functionals' for the audio signal
\end{enumerate}

\subsection{Attention Mechanism} \label{attention}

\begin{figure}[htbp]
    \centering
    \includegraphics[width=0.8\textwidth]{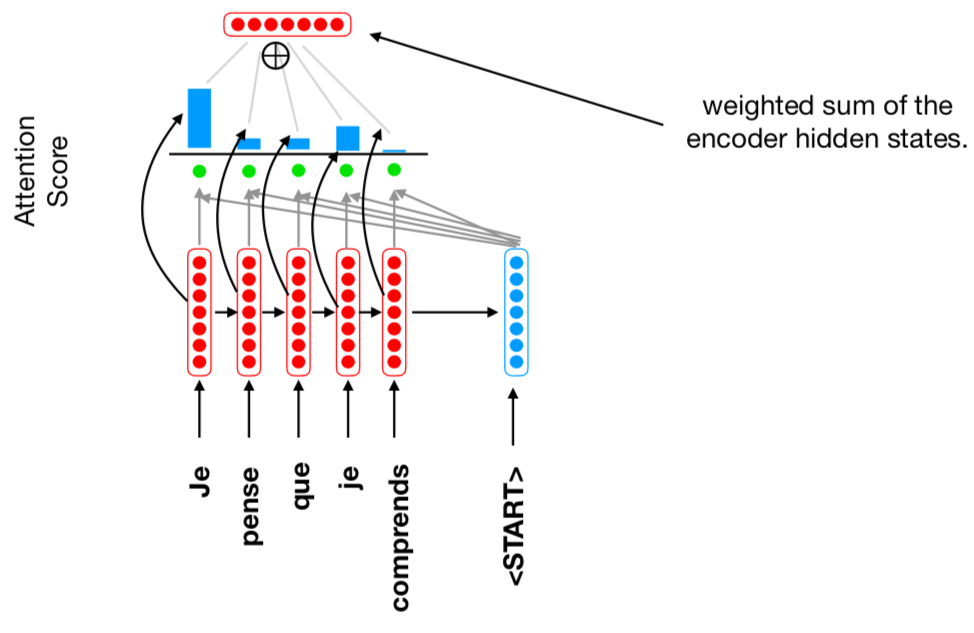}
    \caption{Overview of the Attention Mechanism \cite{nlp_course}}
    \label{fig:attention_fig}
\end{figure}

Although figure \ref{fig:attention_fig} shows the attention mechanism being applied to an NLP problem, the actual process is similar to that followed within a neural network. The overall idea is to calculate scores per feature map of the RNN to boost / suppress certain characteristics of the network.

The steps of implementing attention are:

\begin{itemize}

    \item Pass input $ \mathbf{x} \in \mathcal{R}^d $ into the main neural network $f_{\theta}(\mathbf{x})$ (see equation \ref{rnn_output}), giving an output $\mathbf{z} \in \mathcal{R}^k $
    
    \item Pass input $ \mathbf{x} \in \mathcal{R}^d $ into a simpler neural network $f_{\phi}(\mathbf{x})$ (see equation \ref{attention_output}), which gives a score $\mathbf{a} \in \mathcal{R}^k $ that lies in a certain range (e.g. $a_{i} \in [0,1] $). This can be thought of as learning the importance of different feature maps
    
    \item The two outputs can then be combined element-wise (see equation \ref{attention_mix}), which helps to control the output levels of the main neural network
    
    \item During the training process, parameters $\theta$ and $\phi$ jointly learn to minimise the chosen loss function
    
\end{itemize}

Equations for the attention mechanism are \cite{attention_eq}:

\begin{equation} \label{rnn_output}
    \mathbf{z} = f_{\theta}(\mathbf{x})
\end{equation}

\begin{equation} \label{attention_output}
    \mathbf{a} = f_{\phi}(\mathbf{x})
\end{equation}

\begin{equation} \label{attention_mix}
    \mathbf{g} = \mathbf{a} \odot \mathbf{z}
\end{equation}

\subsection{Optimisation Theory}

Although Stochastic Gradient Descent works well for simple problems, for larger models the method has proven to be inefficient and often gets stuck in local minima rather than finding the true global minimum (or at least a local minima near this level). There are a number of optimisation algorithms available in TensorFlow, in this paper we shall consider the Adam \cite{adam} algorithm (Adaptive Moment Estimation). 

\begin{figure}[htbp]
    \centering
    \includegraphics[width=0.6\textwidth]{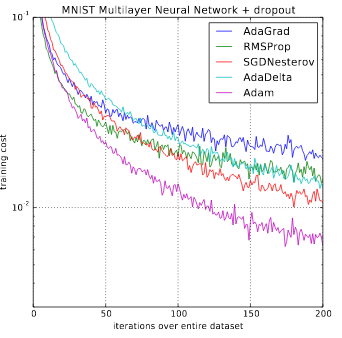}
    \caption{Adam performance overview \cite{adam}}
    \label{fig:adam_im}
\end{figure}

Adam is an efficient and consistent approach (see figure \ref{fig:adam_im}), which utilises the 1st moment to counter momentum ($m_{t}$) and 2nd moment to scale parameters appropriately ($v_{t}$) (taking characteristics of both 'AdaGrad' and 'RMSProp' algorithms). This is particularly useful when problems are poorly defined and therefore the direction of steepest descent doesn't necessarily point towards the global minimum (i.e. takes a non-direct inefficient route). The parameter update form of the Adam algorithm can be seen in equations \ref{adam_eq} - \ref{adam_eq_last}.

\begin{equation} \label{adam_eq}
    \mathbf{m}^{(t)} = \beta_{1}\mathbf{m}^{(t-1)} + (1-\beta_{1})\nabla\hat{L}(\theta^{(t)})
\end{equation}

\begin{equation} 
    \mathbf{v}^{(t)} = \beta_{2}\mathbf{v}^{(t-1)} + (1-\beta_{2})\nabla\hat{L}^{2}(\theta^{(t)})
\end{equation}

\begin{equation} 
    \hat{\mathbf{m}}^{(t)} = \frac{\mathbf{m}^{(t)}}{(1-\beta_{1}^{t})}, 
    \hat{\mathbf{v}}^{(t)} = \frac{\mathbf{v}^{(t)}}{(1-\beta_{2}^{t})} 
\end{equation}

\begin{equation} \label{adam_eq_last}
    \theta^{(t+1)} = \theta^{(t)} - \alpha \frac{\hat{\mathbf{m}}^{(t)}}{\sqrt{\hat{\mathbf{v}}^{(t)}}+\epsilon} 
\end{equation}

\subsection{Loss Function} \label{loss_func}

For multi-class classification problems it is common to use the categorical cross-entropy loss function. However, as argued in \cite{deep_fer_survey}, this has the effect of ``forcing features of different classes to remain apart, but FER in real-world scenarios suffers not only high inter-class similarity but also high intra-class variation". Instead loss functions can penalise the ``distance between deep features and their corresponding class centres". 

The idea being similar classes will actually have similar high and low level features, just minor differences should exist between the class embeddings. This is particularly important for FER, where some of the emotions are very near to each other. 

The survey \cite{deep_fer_survey} proposes a two alternatives that encompass the above thinking and could be exploited in this paper:

\begin{itemize}
    \item Island loss 
    \item Locality-preserving loss 
\end{itemize}

\subsection{Regularisation} \label{regularization}

Given the small size of the AFEW dataset, regularisation is an important factor in training models. The methods used in the CNN models are:

\begin{itemize}
    \item \textbf{Batch Normalisation}: Normalise each / certain layers to better scale the layer and hence improve learning
    
    \item \textbf{Dropout}: Randomly removing neurons from each / certain layer to explore model connectivity
\end{itemize}

Other methods to be considered for the final models are:

\begin{itemize}
    \item Add regularisation term
    
    \item Data augmentation
    
    \item Early stopping
\end{itemize}

\section{Result Aggregation}

To achieve the final classification, the outputs from the multiple models need to be combined (i.e. ensemble method). There are many approaches in this field, with a number of different ones mentioned in the next section. 

Simple approaches such as a weighted average have shown good performance, whilst the winning models from the EmotiW 2018 VA challenge used slightly more complicated methods (summarised in the section below). Once I have obtained final results for the different modalities, I will try a range of options to see what works best for this paper.

\section{Previous EmotiW Submissions} \label{other_papers}

There were 4 papers published from the EmotiW 2018 FER AV challenge, these achieved good performance, but also were commended for their interesting approaches to the problem. I have summarised the key findings below and what can be learnt from these models for my own project:

\begin{itemize}
    \item \cite{occam}: Ranked joint 4th in the EmotiW challenge, this paper is highlighted for being lightweight (i.e. following Occam's razor). Key concepts are:
    \begin{itemize}
        \item \textbf{Visual}:
        \begin{itemize}
            \item Applying just one model (ResNet-18) but pre-training extensively on other facial databases. Trained on AffectNet dataset, but also the `Valence and Arousal' task
            \item Use temporal pooling which they argue performs better than their LSTM model
            \item Also split each sequence into 16 subsets, taking the feature map with the highest score, then applying a classifier to this. The aim being to reduce the impact of frames that the model is not certain about
        \end{itemize}
        \item \textbf{Audio}: Extracting features using openSMILE to a 1,582 dimensional feature map, then applied a (i) random forest classifier and (ii) forward neural net with 64 hidden units, with the latter performing better
        \item \textbf{Other Points}:
        \begin{itemize}
            \item Simple fusion approach of averaging the score limits the number of parameters and hence outperforms complicated methods
        \end{itemize}
    
    \end{itemize}
    
    \item \cite{lu}: Ranked joint 4th in the EmotiW challenge, this paper is noted for the `Multiple Spatio-Temporal Feature Fusion' (MSFF) approaches applied. Key concepts are:
    \begin{itemize}
        \item \textbf{Visual}:
        \begin{itemize}
            \item VGG-Face and ResNet-50 models used for feature extraction, then inputted into Bi-directional LSTM for dynamic element. Fully connected layers fine-tuned first, before training whole network
            \item Sequences of 8 frames inputted and images are of size 224x224
            \item 3D CNNs are used to also capture the spatio-temporal relationship of adjacent frames
        \end{itemize}
        \item \textbf{Audio}: 
        \begin{itemize}
            \item First ``filtered futile part and removed the background noise" \cite{lu} from the raw-data, then transformed to spectrograms with significant data augmentation applied
            \item A VGG-BLSTM framework is used for the final classification, using the VGG model for feature extraction trained on other speech emotion databases
        \end{itemize}
        \item \textbf{Other Points}:
        \begin{itemize}
            \item Significant data augmentation applied to each frame, `such as flipping, mirroring, panning and random cropping, etc.' to improve the ` robustness of the model'
            \item Novel fusion strategy proposed helps boost results, which uses `score matrices' for the different networks to weight their contribution 
        \end{itemize}
    \end{itemize}
    
    \item \cite{fan}: Ranked 3rd in the EmotiW challenge, this paper uses a `Deeply Supervised CNN' (DSN) to improve performance throughout the network architecture:
    \begin{itemize}
         \item \textbf{Visual}:
         \begin{itemize}
             \item DSN works by ``taking the multi-level and multi-scale features extracted from different convolutional layers to provide a more advanced representation of emotion recognition"
             \item The outputs of the shallow and deep layers (each individually optimised) are linked through skip connections and fused to produce the final output (``together achieve complementary effect")
             \item This mechanism is applied to VGG-Face, ResNet-50 and DenseNet models, not other cutting-edge CNN models
         \end{itemize}
         
         \item \textbf{Audio}: Approach not mentioned in the paper
         
         \item \textbf{Other Points}:
        \begin{itemize}
            \item Significant data augmentation applied to each frame not only in training (10+ transformations per image), but also during testing phase by randomly cropping the image and averaging the prediction
            \item CNN models initially trained on facial recognition task (c. 6 million faces in database) then fine-tuned
            \item FACS analysis shows ``Happy and Surprise facial expressions might consist of more distinguishable action units"
            \item ``Class-wise ensemble method" used, which uses weights of classes to average the prediction
        \end{itemize}
         
    \end{itemize}

    \item \cite{liu}: Ranked 1st in the EmotiW challenge with 61.87\% accuracy, their approach was to use multiple model types and then fuse the results in a intelligent way. Key concepts are:
    
    \begin{itemize}
        \item \textbf{Visual}: Three methods utilised:
        \begin{itemize}
            \item CNN model: DenseNet (3 types) and Inception Net (1 type) networks are used for feature extraction. Feature maps were normalised and then SVM used for classification of frames. `5-fold cross-validation' \cite{liu} used to fine-tune the parameters
            \item Landmark model: 3D facial landmarks and euclidean distances give a 34 dimensional feature map, with SVM used to generate predictions
            \item Temporal model: Found VGG to LSTM to be ``most stable and highest performing method''. Applied to sequences of 16 frames (overlap between clips) with images of 224x224. Significant data augmentation used for training, with 128 hidden units giving the highest accuracy
        \end{itemize}
        \item \textbf{Audio}: Two methods utilised:
        \begin{itemize}
            \item openSMILE to extract 1,582 dimensional feature map, then PCA applied with SVM for the final classification. Gave c. 31\% accuracy
            \item SoundNet framework followed by 4 fully connected layers, trained for 100 epochs. Gave c. 33\% accuracy
        \end{itemize}
        \item \textbf{Other Points}:
        \begin{itemize}
            \item Collected own large dataset of emotion recognition video clips called STED to help with training, overcomes one of the major obstacles for the EmotiW FER AV challenge and makes performance comparison difficult
            \item Fusion weights computed based on performance on the AFEW dataset, but added class weights to give c. 4\% boost
            \item Batch size of 8, with decaying learning rate for training starting at 0.01 with `decay coefficient' of 0.95
            \item `3-fold cross validation to tune models' 
            \item `Surprise and Disgust emotions were hard to discriminate'
        \end{itemize}

    \end{itemize}
    
\end{itemize}

Common themes and conclusions to draw from the above papers are:

\begin{itemize}
    \item \textbf{Data}: The lack of training data in the challenge means performance is becoming saturated, with data augmentation the most common approach to addressing this issue
    \item \textbf{Features}: CNN models are used for image feature extraction, which are then used for classification directly or fed into an RNN 
    \item \textbf{Ensemble methods}: Different models with multiple initialisations and then averaging their predictions boosts accuracy considerably. This method is commonly used in all challenges (e.g. Kaggle competitions)
    \item \textbf{Imbalance}: Class-wise weighting, loss penalisation and data augmentation are all explored to deal with this issue
    \item \textbf{RNN}: the LSTM model is most commonly used for the temporal dimension, but GRU may be better suited given the relatively small dataset
    \item \textbf{Training}: A combination of other FER datasets are used for pre-training, but also some models were trained on other tasks (e.g. multi-task learning). Relatively high resolution used for the images, this may an issue in this paper due to the limited computational power available
    \item \textbf{Audio}: Two of the papers use openSMILE to extract features but for the whole video. One paper did split the audio into separate frames, however, they utilised spectrograms rather than openSMILE because the ``features cannot accurately characterise the spatio-temporal information...since they are a combination of diverse speech features". Outputting LLDs rather than the full suite of functionals will hopefully solve this issue (see section \ref{audio_design}). Limited results are published for solely the audio, therefore it is difficult to compare approaches. Also, none of the approaches directly link the visual and audio models (i.e. early fusion, see section \ref{fusion_design}) as stated in the introduction to this paper
    \item \textbf{Implementation}: The submissions seem to have greater computational capacity and more time to experiment / train than made available for this project, making it difficult for this paper to compete results-wise
\end{itemize}

%%%%%%%%%%%%%%%%%%%%%%%%%%%%%%%%%%%%
\chapter{Legal and Ethical Considerations} \label{ethics}

After reviewing the `BCS Code of Conduct' \cite{BCS}, `IET Rules of Conduct' \cite{IET} and `Engineering Council Statement of Ethical Principles' \cite{statement_of_ethics}, I believe I have met their high standards in terms of working diligently, professionally and with integrity throughout this project. 

\section{Direct Implications}

Although FER does rely on data relating to human activity, the AFEW dataset is a collection of clips from well-known ``movies / television sources" \cite{deep_fer_survey}, hence the data is publicly available and with the competition being so widely known, all relevant and required permissions to use the clips have been obtained. Also, the model output is just a label as envisaged by the rules of the challenge, rather than for example manipulating the video and re-publishing it, so I am certain that the project in it's current form has broken no laws.

A possible future legal concern for my project as a piece of FER software would be requiring a license for the usage of openSMILE if the program created was commercialised. However, in its current form, the final model presented in this paper cannot be considered to be in production. 

This undertaking can solely be thought of as a research piece into applying machine learning techniques to the problem of FER. It may be that some parts of this paper are judged to be novel (e.g. deploying cutting-edge models in a new field), but all the constituent parts have been published in their own right and implemented elsewhere previously, therefore I do not believe I have trodden any new ground not previously considered ethically.

\section{Future Implications}

As machine learning grows as a field, its ability to accurately identify human emotions will improve substantially. This improvement will increase the legal and ethical ramifications of the technology.

At the time of writing this paper, the related area of automated facial recognition is in the news regularly, for example, its deployment in the King's Cross part of London without public knowledge \cite{kings_x_ethics} and Amazon's sale of ``facial recognition technology to US police forces'' \cite{amazon_fr_ethics}. Although some may argue that by tracking and monitoring individuals on a scale previously not possible will keep the general public safe, individual privacy is a very sensitive topic for obvious reasons. Of course there will be some cases where people happily and freely give consent for the technology to be used, such as, facial recognition on phones to unlock the device.

In the case of FER, although the reason for applying the technology may be different, I believe a lot of the ethical concerns raised are similar to those made when objecting to the above cases. To help illustrate this, I have included two futuristic hypothetical examples below, with one likely to be considered ethically acceptable and the other questionable:

\begin{itemize}
    
    \item \textbf{Acceptable Usage}: An individual has an issue with their account, so they video call their bank. It is explained that FER software is being used to improve the service and consent is given to proceed. Through correctly identifying the emotion being displayed by the person, the automated customer service program is able to tailor its responses (i.e. individual is visibly distressed then program would affect a gentle tone / approach) to increase client satisfaction
    
    \item \textbf{Questionable Usage}: FER is software is applied to live `cctv' footage in all public places. The Government's main motivation being that if a group of people are evidencing clearly `Angry' facial indicators then there is potentially a high-risk situation developing. It alerts local police, who are able scrambled to the scene to investigate the threat. However, the constant surveillance of the public can be considered a huge invasion of privacy especially if it is impossible for the population to opt out of the system
    
\end{itemize}

Privacy is a vital consideration when implementing technologies such as FER. A common reason given by those arguing for a `nanny-state' is that if you are not doing anything wrong then there should be no issue with being watched. However, it is celar that when individuals know they are being watched then their behaviour changes and hence the concept of free-will is compromised. 

Also, at least for the foreseeable future the technology will not be perfect (or even close to it \cite{fr_ethics}), with the consequences being more extreme in the negative cases such as the second example above. Key concerns that may lead to unfair / damaging outcomes are:

\begin{itemize}
    
    \item Accuracy of the models
    
    \item Ability for the software to be tricked / fooled
    
    \item Bias in the data 
    
    \item Data protection breaches for the storing of footage plus accompanying emotional labels
    
\end{itemize}

Although the use of FER technology is not widespread, it is certainly being utilised more often, for example, Facebook applying it to pictures on their social network platform. 

AI ethics is quickly becoming a focus area, with policy starting to catch-up with the practical advancements, including numerous committees / panels being established. In the UK the two main relevants bodies are the `Centre for Data and Ethics' (responsible for recommending data-driven policy and building a robust governance system for ethical AI innovation) and the `Information Commissioner's Office' (responsible for enforcing Data Protection laws). 

The application of automated FER certainly has the ability to be a great force for good, but as has been outlined in this section it could easily be abused for ill purposes. Hopefully policymakers and practitioners alike are able to avoid the latter becoming a reality.

%%%%%%%%%%%%%%%%%%%%%%%%%%%%%%%%%%%%
\chapter{Design Approach} \label{design}

This chapter focuses on the high-level decisions made at the start of the project and the general set-up of the models. These decisions were made based on perceived deep learning best practice, successful approaches in last year's EmotiW FER AV challenge and new ideas / methods that we thought may perform well. Low-level decisions (e.g. hyperparameters) and small amendments based on initial results are discussed in Chapter \ref{experimentation_opt}. 

The project pipeline can be split into 5 training sections, which are formed from the two modalities audio and visual. These sections were carried out sequentially, building the wider model out to more accurately capture the complex mapping from the input data to the 7 basic emotion categories. These sections are:

\begin{enumerate}
    \item \textbf{Pre-Training Visual}: The 6 chosen CNN models were all initialised with pre-trained weights from other general datasets (e.g. ImageNet for ResNet-50, DenseNet, SE-ResNet, SE-ResNeXt and NASNet). These networks will accurately capture high-level features, but to improve the models performance on faces (particularly recognising emotion) they are further trained on the datasets discussed in section \ref{Static_Images}. Also, this step is helpful due the lack of data available for the task and improves generalisability of the models
    
    \item \textbf{Fine-Tuning Visual}: These pre-trained CNN models were then trained again on only the AFEW dataset, which forms the basis of the EmotiW FER AV challenge. This helps the models to become more specific to this task and reduces covariance shift (i.e. image quality and environment variation)
    
    \item \textbf{Temporal Visual}: To better capture the sequential nature of the videos, the CNN models are used to extract feature maps for each frame of a video, which are then fed into the RNN to produce a classification for the whole video
    
    \item \textbf{Audio}: As discussed in section \ref{audio_models}, there are three common approaches to signal processing, (i) `Raw Form' such as CNN+RNN or SoundNet, (ii) `Spectrograms' to transform the data or (iii) `openSMILE' the state-of-the art tool for feature extraction. The application of these three methods will vary depending on the type and length of audio clips
    
    \item \textbf{Fusion}: There are two fusion options available across the modalities and models:
    \begin{enumerate}
        \item \textbf{Early Fusion}: Combine features or model outputs at the frame level, which can then be inputted into a final classifier model
        \item \textbf{Late Fusion}: Combine the results of the final stage models using an ensemble method (e.g. weighted average, majority voting, etc.) 
    \end{enumerate}
    
\end{enumerate}

Based on last year's submissions to the EmotiW FER AV challenge and general performance on audio vs. image tasks, I expect the visual component to outperform the audio models. Therefore more time in the project was spent on improving the visual networks' accuracy levels, with the audio models there to provide an additional boost when the features / results were included.

\section{Visual Models} \label{visual_design}

The CNN models to be explored in this paper are:

\begin{itemize}
    \item VGG-Face
    \item ResNet-50
    \item DenseNet-121
    \item SE-ResNet-50
    \item SE-ResNeXt-50
    \item NASNet
\end{itemize}

An explanation of each of the above model's architectures and key advantages can be found in section \ref{CNNs}. These models have been chosen based on their strong performance in other image challenges, with the hope that that could be applied to the FER task. The top-3 models have been used in other submissions to the EmotiW FER AV 2018 challenge, but the bottom-3 models were published very recently and thus have yet to be considered fully for FER. There are important reasons, mentioned in section \ref{CNNs}, as to why these models may work well.

The underlying theme behind each of the models is to allow more complex mappings to be learnt, whilst limiting the number of parameters to aid learning. This approach is key for successfully classifying the AFEW dataset because of its limited size. It should be noted that deeper versions of some of the above networks are in common use (e.g. ResNet-101 has 51 more layers than the ResNet-50 network we implement in this paper), but were ignored because of the increased number of parameters required to be trained.

All models have been trained in a similar fashion and evaluated to provide a helpful comparison between them. Also, late fusion of model results is enhanced by independence, this can either be done by (i) creating varied subsets of the data through sampling (i.e. the `bootstrap algorithm'), which is difficult for a small dataset like AFEW or (ii) applying different types of models that will capture a diverse set of features, when aggregated the prediction power of the models increases. The latter is the approach taken in this paper.

\section{Audio Models} \label{audio_design}

Of the 3 options presented in section \ref{audio_models} and mentioned above, I believe openSMILE is best suited to the FER task and will produce the best results for the following reasons:

\begin{itemize}
    \item There are only 773 videos in the AFEW training dataset, which makes training models from scratch on the raw-data and spectrograms difficult. Also, making parameter / model architecture decisions is hard for limited data. openSMILE has been trained and demonstrated success on a range of audio datasets, meaning it is likely to more accurate and robust
    \item The openSMILE tool captures Mel-Frequency-Cepstral Cofficients (MFCC) in the LLDs, as well as another number of other useful components (see Appendix A for list), meaning it performs similar analysis to spectrograms, whilst capturing other helpful information
    \item The software itself has a focus on emotion recognition, with a number of the challenges the team have competed in being in this field, therefore it is well aligned with this paper's FER goal
    \item A number of the entrants in the EmotiW challenge last year chose to apply openSMILE for the audio part of the task
    
\end{itemize}

Depending on the type of audio input, the output of openSMILE may be chosen as follows:

\begin{enumerate}
    \item For long clips (i.e. whole videos) then statistics (i.e. functionals) would be a more helpful representation. They efficiently transform the large amount of audio data into a manageable feature vector
    \item For short clips (i.e. edited to match the image frames) then LLDs would be of greater use. The LLDs capture the key components of the audio input, with statistical analysis less telling due to the limited clip length
\end{enumerate}

Based on the feature outputs above, the classifier models chosen for the audio segment are:

\begin{itemize}
    \item \textbf{Whole Video}: The feature map for each video will be a single vector (i.e. tabular form), hence a `Forward-Neural Network' would be a good option (SVM and linear regression are also possibilities). This approach is utilised by 2 of the top entrants to the 2018 EmotiW FER AV challenge, with \cite{occam} finding that the `Forward-Neural Network' performs best
    \item \textbf{Video Clips}: The LLD feature maps will be per frame, hence an RNN layer model will best capture the temporal relationship between the frames, with a fully connected layer for the final classifications
\end{itemize}

\section{Sequential Models} \label{seq_model_design}

RNNs by themselves are not good at handling complex raw-data, because it is unable to efficiently learn the underlying structure. This is because there is only a single weight applied to the input ($W_{xh}$, see section \ref{sequence_models} for further details) rather than a deep network capable of complex mapping functions. Therefore we can use the trained CNNs or openSMILE as feature extractors for the raw-data, providing an input of greater meaning with a more manageable dimensionality for the RNN.

As discussed in section \ref{sequence_models}, the architectures best suited to capturing long range dependencies (required given the length of the videos) are LSTM and GRU models. The GRU network has fewer parameters, so is easier to train on a small dataset like AFEW. Hence, this will be choice approach for all temporal modelling.

In addition, the benefit of including bi-directional information (i.e. BGRU) and the attention mechanism will be implemented to gauge the relative positive or negative impact.

\section{Fusion Approach} \label{fusion_design}

\subsubsection{Early Fusion}

For `early fusion', the method of combining feature maps / model outputs commonly used is simple concatenation. A weighted approach could be employed, but this would add an extra set of parameters for the model to learn and the GRU will naturally learn weights to apply to the input (i.e. new combined feature map).

By fusing the visual and audio feature maps, the aim is to increase the model certainty for a certain classification where the two streams agree and suppress the uncertain cases (i.e. visual data suggests subject is possibly `Angry' but the audio data thinks the subject is `Happy', although the frame classification might still be `Angry' by reducing the activation the frame will contribute less to the final sequence prediction). This will hopefully boost the predictive power overall by introducing the additional information.

The main decision for early fusion between the audio and visual models on the frame-level clips is when to apply it, the four options explored were (see section \ref{fusion_exp} for further details and results):

\begin{enumerate}
    \item Concatenate the CNN feature map with the LLD features from openSMILE and feed these into a GRU
    \item Concatenate the CNN feature map with the output of GRU applied to the LLD features, this new combined feature map per frame is inputted to a new GRU
    \item Concatenate the CNN feature map with the output of GRU applied to the LLD features, this new combined feature map per frame is then passed through a fully-connected layer 
    \item Concatenate the visual CNN + GRU output states with the output of GRU applied to the LLD features, this new combined feature map is inputted to a new GRU. The advantage here being that the feature maps of both modalities are more evenly matched in dimension and therefore there contribution is weighted better
\end{enumerate}

In each of the above cases, the weights from the training stages discussed above were used to initialise the models. 

\subsubsection{Late Fusion}

In the case of `late fusion', the aim is to find model and class weights that produce the highest accuracy across all model output permutations. Since there are limitless permutations, I will use the following criteria to select networks for inclusion based on ensemble methods best practice:

\begin{itemize}
    \item Those with the highest accuracy. Ensemble methods works by combining the predictive power of weak learners, but the more accurate the weak learners, the better the outcome. Including too many poor performing models may make it difficult to weight the correctly
    
    \item As discussed in the introduction to this section, including different types of models that will capture a diverse set of features and hence produce more independent network outputs. When aggregated the predictive power of the models increases because hopefully the strengths of each model drives final classifications (e.g. model A predicts `Angry' and model B predicts `Disgust' correctly with near certainty, if combined intelligently then the resulting model should now classify both emotions accurately)
    
\end{itemize}

There are numerous ways to find the weights, those explored in this paper are:

\begin{itemize}
    \item Majority voting
    \item Class predictions weighted by model accuracy
    \item Model logits weighted by model accuracy
    \item Linear regression for model weights
    \item Include class weights to reduce imbalance in data
    \item Learn class weights to reflect class specific model performance 
\end{itemize}

%%%%%%%%%%%%%%%%%%%%%%%%%%%%%%%%%%%%
\chapter{Implementation} \label{implementation}

In Chapter \ref{design}, I laid out the initial vision for how the FER challenge was going to be approached, in this chapter I will go into a bit more detail on how the models chosen were actually applied. 

\section{Data Pre-Processing} \label{data_process}

The format of the data is important to ensure as much information as possible is extracted from the raw form and to help the models run efficiently. Also, standardising and normalising the data is key to improving model behaviour and learning, because this allows the model to focus on only the important features.

\subsection{Images}

\begin{figure}[htbp]

\centering
\subfigure[Undetected Image]{\label{fig:face_uncropped}\includegraphics[height=50mm,width=50mm]{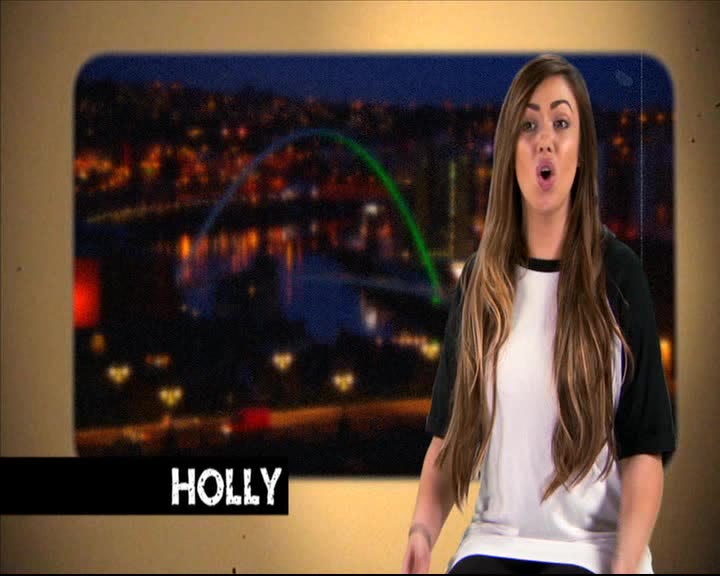}}
\subfigure[Detected Image]{\label{fig:face_cropped}\includegraphics[height=50mm,width=50mm]{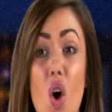}}

\caption{Example of undetected to detected image}
\label{fig:face_detection_process}

\end{figure}

\begin{figure}[htbp]

\centering
\subfigure[Angry Image]{\label{fig:face_det_1}\includegraphics[width=40mm]{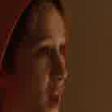}}
\subfigure[Happy Image]{\label{fig:face_det_2}\includegraphics[width=40mm]{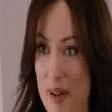}}
\subfigure[Surprise Image]{\label{fig:face_det_3}\includegraphics[width=40mm]{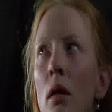}}

\caption{Selection of face detected images}
\label{fig:face_detect}

\end{figure}

The datasets mentioned in section \ref{Static_Images} all have alignment and normalization software applied to standardise the image inputs. In the case of the AFEW dataset, the software is employed on a frame-by-frame basis, with the output being in `.jpg' format (see figures \ref{fig:face_detection_process} and \ref{fig:face_detect} for examples of this process, note that a perfect transformation is not always possible). However, in certain instances (e.g. obscured, poor lighting, odd angle, etc.) the software has been unable to detect a face and therefore these frames have been skipped (including 25 training videos, where the software was unable to detect a face at all). A manual process could have been applied, but the impact of the missing frames was felt to be minimal, with missing videos all in the largest category.

All images (both static and dynamic) have been scaled to the following dimensions, 112*112*3, using the TensorFlow tool `$image.resize\_images$' to ensure the same input size to the models throughout otherwise errors would occur in the fully connected layers. The decision on the dimensions had to be made at the start of the project, with a trade-off between image resolution and memory usage the main consideration. Also, the pixel values in the images are transformed to the range $[-1,1]$ again for standardisation purposes.

The `.jpg' images can be processed in TensorFlow using the `$image.decode\_jpeg$' tool.

\subsection{Audio} \label{audio_imp}

As explained in section \ref{Audio_data}, the audio data is obtained by converting each `.avi' video file into a `.wav' file (taking just the audio)using the python open-source library ffmpeg. The following steps were applied then applied to the `.wav' files:

\begin{itemize}
    \item \textbf{Whole Video}: Use the entire `.wav' file as the input to openSMILE, with the `$emo\_large$' configuration deployed to `extract a larger feature set with more functionals and more LLD enabled (total 6,552 features)' 
    
    \item \textbf{Video Clips}: The multi-stage process applied was:
    \begin{itemize}
        \item Use the `AudioSegment' tool within the python library `pydub' to turn the entire raw-audio into clips of 0.04 seconds to match the video FPS
        \item As explained above, the face detection software often skipped frames, so the audio clips had to be aligned exactly to ensure the audio and visual models could be fused later on. This meant some of the audio clips had to be discarded
        \item Apply a modified openSMILE configuration file (see attached code files) to only output the LLDs for the specific clip size. Output was 112 key features per clip. The decision to amend an existing configuration file was made for timing reasons
        
    \end{itemize}

\end{itemize}

The functionals / LLDs computed by openSMILE are saved in an `.arff' file, this can then be converted into a `.csv' file, which in turn can be processed in TensorFlow using the `$io.decode\_csv$' tool.

\subsection{Labels} 

The labels for the static image datasets in section \ref{Static_Images} were not identical to the AFEW dataset, so had to be amended and aligned to ensure the models were learning correctly. The labels were converted to one-hot encoding, which is the format required for the TensorFlow loss function `$losses.softmax\_cross\_entropy$' used in the training process.

\subsection{Text Location and Label Files} 

Although the data for the visual and audio are stored in `.jpg' and `.csv' formats respectively, these are not inputted directly to the model scripts. Instead the file locations and labels were saved into a `.txt' file, which TensorFlow could read-in and process the data directly. 

$$/vol/gpudata/cn308/Data/Audio\_data/csv\_clips/Val\_AFEW/Angry/004514600/00001.csv$$ 
$$/vol/phoebe/dk15/databases\_facesoft/aligned/emotiw19/valid/Angry/004514600/00001.jpg$$
$$ 0 \: 1 \: 0 \: 0 \: 0 \: 0 \: 0 $$

\subsection{Data Manipulation} \label{data_manip}

Before the data could be processed, two further manipulations were needed for training on the AFEW dataset (see the `$video\_process$' and `$video\_clip\_process$' code files for full details):

\begin{itemize}
    \item For training and evaluation, the image / audio clips have to be arranged into blocks to ensure sequences for the same video have information regarding labels, length of the sequence and video identification correctly aligned (required for RNN and `early fusion' models)
    \item TensorFlow requires that all entries in a tensor have the same dimensions, this means that sequences must have the same length, which requires padding. I chose the padded value to be the last frame of the sequence
    
\end{itemize}

\section{Sequential Models Implementation} \label{seq_model_imp}

The feature maps extracted from the raw-data are inputted to the RNN model as follows:

\begin{itemize}
    
    \item \textbf{Visual}: The CNN model is set-up, but only the weights for the layers up to the first dense / fully connected layer (e.g. for VGG-Face this would be the 4096 dimensional layer following the final convolutional block) are restored. The raw-data per frame can be inputted to the CNN, with the activations for this first dense / fully connected layer forming the feature map for the image and the input to the RNN. Note that the final classification layer(s) in the CNN are disregarded 
    
    \item \textbf{Audio}: RNN only applied to the `Video Clip' scenario outlined in section \ref{audio_design}. The openSMILE output of LLDs is the feature map for the audio raw-data and is of a reasonable dimension, so this can be fed directly into the GRU for each time-step
    
\end{itemize}

The RNN outputs a state for each time-step (i.e. frame), with dimensions matching the number of hidden units for the RNN. This cell state output is then passed through a fully connected layer to get a prediction for the frame (i.e. one of the 7 basic emotion categories). The predictions per frame can be aggregated into a single sequence prediction the same way as discussed in section \ref{training_process}. 

To improve model performance, the exact sequence length is also inputted to the RNN model, so the model outputs of the padded frames are ignored during the training procedure without having to manually remove these frames.

\section{Model Scripts} 

A key implementation decision made, was which machine learning library to use to create the project pipeline. The two options considered were (i) PyTorch (used in the Imperial MSc Machine Learning courseworks) and (ii) TensorFlow. The latter was the decided upon because it has been around slightly longer and therefore has a lot of supporting online literature, strong community useful for bug fixing and the models mentioned in this paper could be easily constructed in TensorFlow or found on `github' in the TensorFlow format then amended.

The code for all models can be found attached to this report. The network architectures are constructed in separate `.py' scripts. As discussed in the section below, to standardise the pipeline for efficiency purposes, each model had the same python `Class' structure and output format.

\section{Project Pipeline} 

\begin{figure}[htbp]
    \centering
    \includegraphics[width=0.7\textwidth]{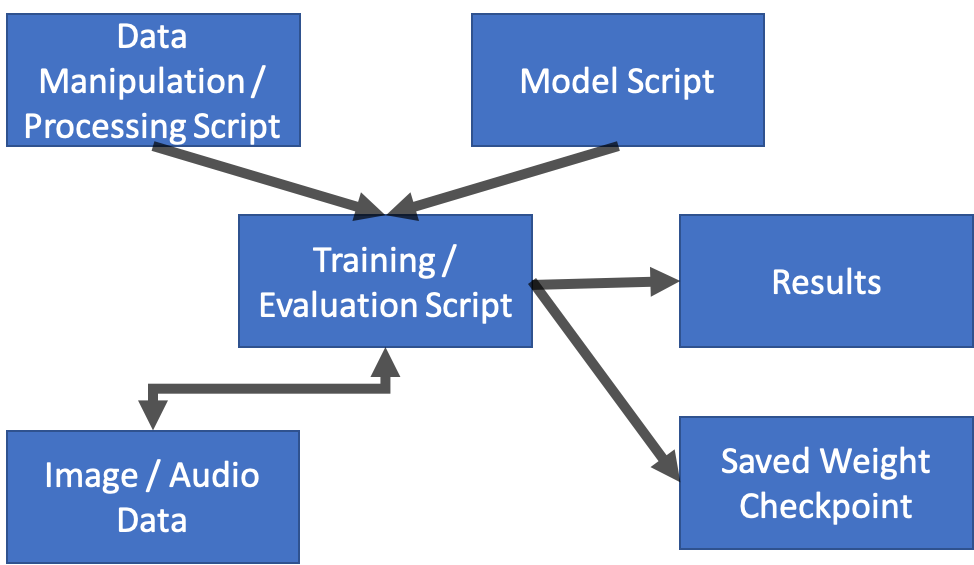}
    \caption{Project pipeline for training and evaluation}
    \label{fig:model_setup}
\end{figure}

The workflow outlined in figure \ref{fig:model_setup} represents the interaction between scripts (e.g. reads-in the relevant model), data inputs (e.g. fetches the correct data) and outputs (e.g. saves the results / weights into run dependent folder). The aim was to standardise the pipeline as much as possible, so that only minor changes had to be made to the train / evaluation script at each stage (e.g. structurally different, such as, CNN, CNN + GRU, GRU + GRU, etc.) and to improve consistency of results.

The train / evaluation scripts (in the code attached to this paper) have `flags' at the top that when altered (i) change the source of the data, weights to restore, location of saved output and model type being used or (ii) select different settings for data manipulation and model architecture. 

\section{Training and Evaluation Scripts} \label{train_eval_script}

The train and evaluation scripts followed the same basic steps (see the attached code files for greater detail), which have been summarised below:

\begin{enumerate}
    \item Select `flags' specific to desired model run
    \item Load `.txt' file with data locations and labels
    \item Transform data into correct tensor shape (e.g. number of sequences, sequence length, raw-data)
    \item Create data loader, then shuffle (if training), decode and batch. Initially used the TensorFlow `$tf.data.Dataset$' API, but had to switch to `$tf.train.slice\_input\_producer$' tool for the later larger audio-visual models, which is better at handling multiple different data types 
    \item Reform the tensors into correct shape for the models (e.g. RNNs and CNNs have different dimension inputs in TensorFlow)
    \item Setup network and feed-in data
    \item Choose variables to train / restore / save 
    \item Network output is the sequence prediction
    \item Calculate the loss and define optimiser (if training)
    \item Initialise TensorFlow session and variables to run script
    \item Save checkpoints (if training) and results
\end{enumerate}

Certain checks were put in place throughout the script (i.e. assert X == Y, print shapes of tensors, etc.) to ensure models were operating as expected.

\section{Execution of the Models} \label{execution}

Given the size of the models and amount of data, the models could not be run on normal Imperial lab computers or my personal computer. Instead the majority of the scripts were run on the Imperial GPU clusters. A summary of the steps for execution are included below:

\begin{itemize}
    \item To begin with I had to create a python virtual environment specific to my project to install the python libraries required, ensuring the correct versions were available
    \item Create `.sh' scripts to send to the GPU clusters, which launched the correct python training / evaluation scripts
    \item ssh into a lab computer to then ssh into the GPU clusters, activating the `pbs job' manager
    \item In certain cases specify which GPU the job should be run on (i.e. due to size, speed, memory usage)
    
\end{itemize}

In addition to using the GPU clusters, I created some of the models in `Google colab' due to its ease-of-use (GPU cluster output takes a while to return and cannot be viewed during run-time, which makes bug-fixing a slow process) and ran some of the larger models directly on the server `sicklebill' through a tmux session (due to its increased RAM).

\section{Storage} \label{storage}

The data, scripts, model weight checkpoints and results were stored on Imperial drives (i) vol/phoebe and (ii) vol/gpudata) due their large memory availability.

The data folders were split according to train / validation, then by emotional category and finally the separate videos. 

Due to the large number of runs carried out, the model weights and results were stored by stage, then model type (e.g. ResNet-50, DenseNet, etc.) and finally by the run specifications.

\section{Training} \label{training_process}

Given the training process outlined in section \ref{train_eval_script}, the hyperparameter choices will be further detailed in Chapter \ref{experimentation_opt}, but the other main decision to be made was how the make the final prediction. The considerations are summarised below.

To help explain each of the points below, I will refer back to the following example case. A video is of length 100 frames, but the chosen sequence length is 80 frames. This means the video is split into 2 entries to feed into the model (1st of length 80 frames and a 2nd of length 20 frames, but the latter is padded to 80 frames). The label for the video is `Angry', but this is based on a short window where the subject momentarily reacts badly to something. Therefore the majority of the video frames are `Neutral', with only a few frames being `Angry'. 

\begin{itemize}
    \item As mentioned in section \ref{data_manip}, the sequences had to be padded to ensure the input tensors were of the same dimensions. This meant that there were three classifying approaches:
    \begin{enumerate}
        \item \textbf{Classify per frame}: The frame prediction and label for the whole video were passed into the loss function. A benefit being there a more data points for the model to train on, but the predictions will be more noisy (e.g. mainly `Neutral' predictions for the `Angry' label) meaning the model may be learning incorrectly 
        \item \textbf{Classify for the exact sequence}: The prediction is made by aggregating (see next bullet point) across all actual sequence frames, based on our example above the 1st entry is reduced from 80 classifications to 1 final prediction and the 2nd entry is reduced from 20 classifications to 1 final prediction. This helps to remove some of the noise, but now only have one input to the loss function per sequence 
        \item \textbf{Classify for model output}: Similar to the above approach, but the 2nd entry is reduced from 80 classifications (includes padded frames) to 1 final prediction
    \end{enumerate}
    
    \item The method of aggregating the sequence frames is also important, the `$argmax$' can then be taken to give the prediction label, with two options explored:
    \begin{enumerate}
        \item \textbf{Reduce Mean}: Take the mean of the logits (i.e. along the sequence length)
        \item \textbf{Reduce Median}: Take the median of the logits
    \end{enumerate}
\end{itemize}

The above example highlights a major difficulty of classifying video sequences, not all the data perfectly relates to the label. Hence the model has to extract strong results from the relevant frames and suppress the outputs from the rest.

Logically I would expect the `Classify for the exact sequence' and `Reduce Mean' options to perform best, as they use only the exact video data and the mean should give greater weight to strong activations (i.e. model is certain that the subject is `Angry' for certain frames). All the above options are explored in the next chapter.

\section{Evaluation} \label{eval_implementation}

The performance metric used by the competition is accuracy, but for an imbalanced dataset, as discussed in section \ref{discussion}, this measure can often be misleading. Therefore, I have also computed the `F1-Score' and `Confusion Matrix' for each evaluation run to provide further feedback on model behaviour. 

In the early training stages we are looking for the best feature extractors, because their weights will be used by other models, which means judging solely on accuracy may once again be deceptive.

\section{Fusion Implementation} \label{fusion_implementation} 

The process for implementing the two forms of fusion mentioned in section \ref{fusion_design} are outlined below:

\subsubsection{Early Fusion}

\begin{itemize}

    \item Based on the four options described in section \ref{fusion_design}, the TensorFlow function ‘pd.concat()’ is used to merge the audio and visual feature maps before being fed into the classifier
    
    \item The main difficulty being to ensure that for both types of models being called, the appropriate weights are restored and correct parts of the network trained
    
\end{itemize}

\subsubsection{Late Fusion}

\begin{itemize}
    
    \item When evaluating a model, as well as the 3 main performance metrics (accuracy, F1-score and confusion matrix), a  `.txt’ file is also stored containing the video id, logits, predictions and labels for all data points
    
    \item In a separate script, the name of the models to be fused are stored in a list, which then loads each of the files and orgnaises the data into 3 dictionaries: (i) accuracy per model to be used to calculate weights, (ii) predictions for each model by video and (iii) logits for each model by video
    
    \item Each fusion method is stored as a function, which follows the general method of running through each video id and then looking up the results for each model selected. The results have the weights learned (if applicable) and applied, with the argmax of the summed result being the final late fused prediction for the video
    
    \item The performance metrics are then calculated over the whole dataset and can be compared to identify the best method
    
\end{itemize}

\section{Testing} 

To produce final submissions a number of the steps mentioned in this chapter were repeated for both visual and audio workstreams (e.g. face detection and alignment software, extracting the audio / converting into clips, applying the openSMILE tool, etc.). However, amendments were required at each stage because labels are no longer included and the data is stored in a different way (i.e. not split according to emotion). 

The input to the models is still a `.txt' file with image frame and audio feature map locations included, but evaluation scripts were altered for testing purposes to solely produce a prediction per video sample. No performance metrics can be calculated because the labels are non-existent.

The final stage of `late fusion' was adapted to store predictions in the format designated by the EmotiW 2019 competitions rules. A `.txt' file is created for each sample, containing solely one of the 7 labels.

%%%%%%%%%%%%%%%%%%%%%%%%%%%%%%%%%%%%
\chapter{Experimentation and Optimisation} \label{experimentation_opt}

Chapters \ref{design} and \ref{implementation} discussed the general approach envisaged to tackle this FER challenge and the plan on how to carry that out. This chapter goes into greater depth at each stage of the pipeline, recording the base results and detailing what changes were made to optimise performance. These changes produced new results, which could then be analysed, with new hypotheses for improvement formed. This follows the classic `machine learning model feedback loop' set-out in figure \ref{fig:model_fb}. I have tried to include rationale for my decisions made where possible.

\subsubsection{Overview of Chapter}

The following helps to act as a guide to this chapter:

\begin{itemize}
    
    \item \textbf{Order}: This chapter follows the training stages outlined at the start of Chapter \ref{design}, which roughly follows the chronological order the project was carried out in
    
    \item \textbf{Model Run Breakdown}: For each visual stage there are 6 models to run, to begin with I ran a base case for each CNN, with all experimentation done with the VGG-Face network to save time. Although not a strict rule, the idea was that amendments that produce noticeably stronger results for VGG-Face would likely transfer to the other models
    
    \item \textbf{Model Chronology}: Due to limited time, increasing the efficiency of the project pipeline was key. Although in general each of the training sections were completed in order, in some cases it was necessary to start on the next section to understand model behaviour, whilst still optimising the previous training section. I have highlighted throughout this chapter where non-optimal weights have been restored in a new model for the sake of streamlining implementation and to allow greater exploration
    
    \item \textbf{Standardising for Comparison}: A main aim of this project is compare the application of recently published CNN networks to the FER task. To aid with this comparison, I have tried to keep certain parameters fixed for all / the majority of training runs in a stage, for example, all models are trained for 30 epochs in the `Pre-Training Visual' stage. Another approach would be to keep training until maximum performance is found for a particular run (i.e. NASNet is a much larger model, so will likely require more training epochs) and then compare, but given the limited time and the fact that this optimal point may not be located, I have chosen to standardise 
    
    \item \textbf{Metrics}: All results below are for the AFEW validation dataset. The main metrics used to gauge the relative success / failings of a model is the `accuracy rate', followed by the `F1-score'. The reasons for this hierarchy are explained in section \ref{eval_implementation}. I have included graphs of the training loss / accuracy across the batch iterations and the confusion matrix for certain model runs of particular interest
    
    \item \textbf{Key Results}: In the tables below, I have highlighted certain results in yellow to indicate where a key decision was made or that the run is of particular importance
    
    % \item \textbf{Additional Results}: Some further runs were carried out and additional result tables can be found in Appendix 2
    
\end{itemize}

\section{Pre-Training Visual} \label{pre_train_exp}

There are two objectives in this training stage:

\begin{itemize}

    \item \textbf{Transfer Learning}: The 6 CNN models have all been proven to be successful on other visual tasks (e.g. ImageNet challenge), the aim is to initialise our models with these weights and then train these models specifically for the FER task \cite{detection}. The advantage of this approach is that the network doesn't need to learn everything from scratch (i.e. can utilise the high-level features already learned), which may be time consuming and ineffective given the smaller size of the available FER datasets
    
    \item \textbf{Model Generalisability}: If training was done directly on the AFEW training dataset, the models would likely suffer from overfitting very quickly given the limited number of videos, which would hinder performance during testing. By using the environmentally and subject diverse datasets outlined in section \ref{Static_Images}, we are able to keep the models from becoming too narrow in their focus  
    
\end{itemize}

Given the above two aims, in this stage we are not trying to maximise performance to the extreme, but more interested in what general settings seem to work well for the FER task. In later stages where we are fine-tuning on the AFEW dataset, more time will be spent trying different hypotheses. All runs are carried out for 30 epochs and have a batch size of 64.

\begin{center}
\begin{tabular}{ |p{1cm}||p{4cm}|p{2cm}|p{2cm}|p{2cm}|p{2cm}|}
 \hline
 \rowcolor{lightgray} \multicolumn{6}{|c|}{Table 1: VGG-Face} \\
 \hline \hline
 Run & Description & Metric & RAFDB & AffectNet & FER2013 \\
 \hline \hline

 One & LR=0.0001 & Accuracy & 71.5\% & 52.6\% & 67.9\% \\
 & stationary & F1-Score & 57.1\% & 50.6\% & 67.2\% \\
 \hline \hline
 
 Two & LR=0.00001 & Accuracy & 72.7\% & 51.9\% & 68.9\% \\
 & stationary & F1-Score & 59.6\% & 50.6\% & 67.3\% \\
 \hline \hline
 
 \rowcolor{yellow} 
 Three & LR=0.0001 & Accuracy & 73.8\% & 52.1\% & 69.6\% \\
 \rowcolor{yellow} 
 & reducing & F1-Score & 59.2\% & 50.4\% & 67.8\% \\
 
 \hline

\end{tabular}
\end{center}

The VGG-Face model in table 1 is mainly investigating the impact of altering the learning rate (LR). As explained in section \ref{deep_learning_basics}, the learning rate controls the magnitude of the update and based on `Run Three' reducing the learning rate over time produces the best results.

The method for reducing the learning rate is:

\begin{itemize}
    \item Choose an initial learning rate, in the case of `Run Three' this is 0.0001
    \item Decide when the reduction will occur (i.e. every 5000 steps), the size of the reduction (i.e. reduce by 0.95) and the profile of the step (i.e. staircase profile rather than something smoother)
    \item Instruct the optimiser (e.g. Adam optimisation) to adjust the learning rate according to the above when training
\end{itemize}

For the models in tables 2 to 5, an additional option was to collapse the final convolutional feature map to a smaller dimension (e.g. (batch size, 4, 4, 2048) to (batch size, 1, 1, 2048)), this is recommended for straight classification tasks by the original papers (i.e. `global pool = True'), but in doing this we may lose some helpful information and for the temporal stages where we are feeding the feature maps into an RNN rather than classifying directly.  The latter does however increase the number of parameters in the model. Therefore the below 4 tables explore the impact of choosing global pool to be True or False, with the best results highlighted in yellow:

\begin{center}
\begin{tabular}{ |p{1cm}||p{4cm}|p{2cm}|p{2cm}|p{2cm}|p{2cm}|}
 \hline
 \rowcolor{lightgray} \multicolumn{6}{|c|}{Table 2: ResNet-50} \\
 \hline \hline
 Run & Description & Metric & RAFDB & AffectNet & FER2013 \\
 \hline \hline
 
 One & LR=0.0001, static & Accuracy & 67.5\% & 50.6\% & 68.2\% \\
 & global pool = True & F1-Score & 53.2\% & 48.6\% & 67.2\% \\
 \hline \hline
 
 \rowcolor{yellow} 
 Two & LR=0.0001, static & Accuracy & 71.9\% & 51.8\% & 67.5\% \\
 \rowcolor{yellow} 
 & global pool = False & F1-Score & 57.8\% & 50.0\% & 65.9\% \\
 \hline \hline
 
 Three & LR=0.0001, reduce & Accuracy & 75.1\% & 52.4\% & 68.4\% \\
 & global pool = False & F1-Score & 61.6\% & 51.1\% & 66.2\% \\
 
 \hline

\end{tabular}
\end{center}

\begin{center}
\begin{tabular}{ |p{1cm}||p{4cm}|p{2cm}|p{2cm}|p{2cm}|p{2cm}|}
 \hline
 \rowcolor{lightgray} \multicolumn{6}{|c|}{Table 3: DenseNet-121} \\
 \hline \hline
 Run & Description & Metric & RAFDB & AffectNet & FER2013 \\
 \hline \hline
 
 \rowcolor{yellow} 
 One & LR=0.0001, static & Accuracy & 63.5\% & 51.1\% & 67.1\% \\
 \rowcolor{yellow} 
 & global pool = True & F1-Score & 48.1\% & 48.8\% & 65.2\% \\
 \hline \hline

 Two & LR=0.0001, static & Accuracy & 61.0\% & 32.6\% & 44.6\% \\
 & global pool = False & F1-Score & 29.3\% & 22.4\% & 25.6\% \\
 \hline \hline
 
 Three & LR=0.0001, reduce & Accuracy & 71.0\% & 45.1\% & 60.9\% \\
 & global pool = True & F1-Score & 50.0\% & 38.4\% & 47.3\% \\
 
 \hline

\end{tabular}
\end{center}

\begin{center}
\begin{tabular}{ |p{1cm}||p{4cm}|p{2cm}|p{2cm}|p{2cm}|p{2cm}|}
 \hline
 \rowcolor{lightgray} \multicolumn{6}{|c|}{Table 4: SE-ResNet-50} \\
 \hline \hline
 Run & Description & Metric & RAFDB & AffectNet & FER2013 \\
 \hline \hline
 
 One & LR=0.0001, static & Accuracy & 74.3\% & 52.6\% & 68.0\% \\
 & global pool = True & F1-Score & 58.6\% & 50.9\% & 66.5\% \\
 \hline \hline
 
 \rowcolor{yellow} 
 Two & LR=0.0001, static & Accuracy & 75.8\% & 54.2\% & 68.0\% \\
 \rowcolor{yellow} 
 & global pool = False & F1-Score & 61.0\% & 52.5\% & 66.6\% \\
 \hline \hline
 
 Three & LR=0.0001, reduce & Accuracy & 78.5\% & 53.5\% & 68.3\% \\
 & global pool = False & F1-Score & 63.9\% & 52.1\% & 67.1\% \\
 
 \hline

\end{tabular}
\end{center}

\begin{center}
\begin{tabular}{ |p{1cm}||p{4cm}|p{2cm}|p{2cm}|p{2cm}|p{2cm}|}
 \hline
 \rowcolor{lightgray} \multicolumn{6}{|c|}{Table 5: SE-ResNeXt-50} \\
 \hline \hline
 Run & Description & Metric & RAFDB & AffectNet & FER2013 \\
 \hline \hline
 
 \rowcolor{yellow} 
 One & LR=0.0001, static & Accuracy & 73.8\% & 53.3\% & 69.1\% \\
 \rowcolor{yellow} 
 & global pool = True & F1-Score & 58.5\% & 51.5\% & 68.1\% \\
 \hline \hline

 Two & LR=0.0001, static & Accuracy & 73.2\% & 52.1\% & 67.1\% \\
 & global pool = False & F1-Score & 58.2\% & 50.7\% & 65.9\% \\
 \hline \hline
 
 Three & LR=0.0001, reduce & Accuracy & 76.7\% & 54.4\% & 69.7\% \\
 & global pool = True & F1-Score & 60.2\% & 53.0\% & 68.2\% \\
 
 \hline

\end{tabular}
\end{center}

Global pool is applied to the NASNet model, but given the cell structures are pre-trained in a specific way, I chose not to amend this network:

\begin{center}
\begin{tabular}{ |p{1cm}||p{4cm}|p{2cm}|p{2cm}|p{2cm}|p{2cm}|}
 \hline
 \rowcolor{lightgray} \multicolumn{6}{|c|}{Table 6: NASNet} \\
 \hline \hline
 Run & Description & Metric & RAFDB & AffectNet & FER2013 \\
 \hline \hline

 One & LR=0.0001 & Accuracy & 70.6\% & 44.1\% & 61.5\% \\
 & stationary & F1-Score & 48.9\% & 37.3\% & 47.7\% \\
 \hline \hline

 Two & LR=0.0001 & Accuracy & 74.2\% & 43.5\% & 60.2\% \\
 & reducing & F1-Score & 51.8\% & 36.9\% & 46.9\% \\

 \hline

\end{tabular}
\end{center}

\begin{figure}[htbp]
    \centering
    \includegraphics[width=0.8\textwidth]{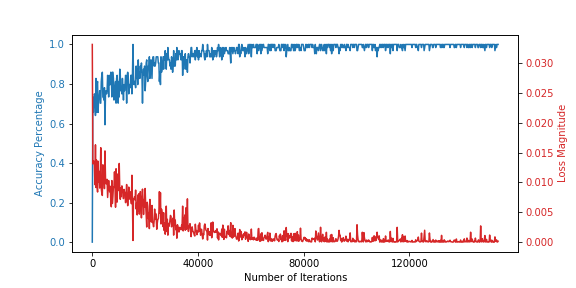}
    \caption{VGG Training Accuracy and Loss results}
    \label{fig:vgg_training}
    
\end{figure}

\begin{figure}[htbp]

\centering
\subfigure[FER2013]{\label{fig:vgg_cm_1}\includegraphics[width=80mm]{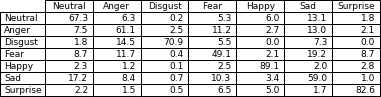}}
\subfigure[RAF-DB]{\label{fig:vgg_cm_2}\includegraphics[width=80mm]{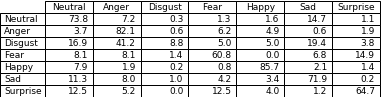}}
\subfigure[AffectNet]{\label{fig:vgg_cm_3}\includegraphics[width=80mm]{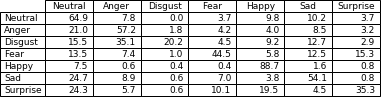}}

\caption{VGG Confusion Matrices by dataset}
\label{fig:vgg_cm}

\end{figure}

\begin{figure}[htbp]
    \centering
    \includegraphics[width=0.8\textwidth]{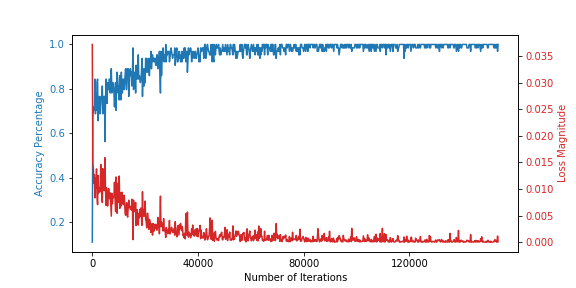}
    \caption{SE-ResNeXt Training Accuracy and Loss results}
    \label{fig:resnext_training}
    
\end{figure}

\begin{figure}[htbp]

\centering
\subfigure[FER2013]{\label{fig:resnext_cm_1}\includegraphics[width=80mm]{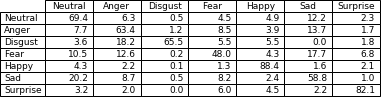}}
\subfigure[RAF-DB]{\label{fig:resnext_cm_2}\includegraphics[width=80mm]{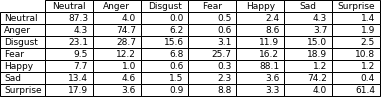}}
\subfigure[AffectNet]{\label{fig:resnext_cm_3}\includegraphics[width=80mm]{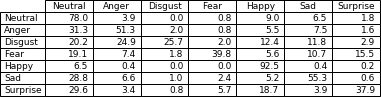}}

\caption{SE-ResNeXt Confusion Matrices by dataset}
\label{fig:resnext_cm}

\end{figure}

I have included training graphs and the confusion matrices for the VGG-Face and SE-ResNeXt models for their respective best runs (other networks follow a similar pattern), given the former is a good benchmark for the task throughout and the latter performs the best in this stage (see below for further discussion).

In figures \ref{fig:vgg_training} and \ref{fig:resnext_training} the path of the accuracy and loss are plotted against number of iterations. The increasing accuracy and decreasing loss over time is the shape we would expect to see if the networks are learning. There is some noise in the levels, but the general direction is more important. Both paths start to plateau around the 80,000 iteration mark for the 2 models, which could suggest significantly more training (i.e. 100 epochs) would result in overfitting, but the reduced noise in the accuracy at the end indicates the model is becoming more consistent and thus 30 epochs is a reasonable training length.

From the confusion matrices in figures \ref{fig:vgg_cm} and \ref{fig:resnext_cm}, the following patterns emerge (although a possible explanation is provided, further analysis of the actual convolutional layer activations would be required to confirm each hypothesis):

\begin{itemize}
    
    \item The emotion the 2 networks consistently classify correctly is `Happy'. This could be because a smile is a clear identifier for the models or it is the only clearly positive reaction and therefore the indicators are more defined / separate from the other emotional tells. The paper \cite{fan} states that the `Happy' ``facial expressions mihg consist of more distinguishable Action units"
    
    \item The lowest performance is for the `Disgust' category, which the networks often mistakes for `Fear', the mistake is somewhat understandable given the closeness of the two reactions facially 
    
    \item The emotions `Fear' and `Surprise' are the next worst, but there is no obvious explanation here as the models predict a range of other classes for the 3 datasets
    
    \item Aggregating across all emotions, the two networks often predict `Neutral' when incorrect, especially for the AffectNet dataset. Perhaps this is because if for a particular emotion there are facial expressions common to other emotions, the model becomes uncertain and so thinks the face is `Neutral'
    
\end{itemize}

Given the results above, the following conclusions can be made which will be carried forward to the next stage:

\begin{itemize}

    \item \textbf{Learning Rate}: A decreasing learning rate can be a benefit (seen most clearly in tables 1 and 6). The reason for this is that a larger step-size in the early training phase helps move the weights towards a suitable local minima and later on small steps are beneficial to get closer to this desired point
    
    \item \textbf{Global Pool}: The impact of reducing the feature map varies across the models, for example, True performs better in tables 3 and 5, but False has higher accuracy in tables 2 and 4. One explanation for this would be the information treatment, reducing the dimensionality by taking the mean (i.e. True case) may help average out inconsistencies in the features or lose important information by aggregating (i.e. False case). The better performing option will be used in later stages
    
    \item \textbf{Best Model(s)}: The two best performing models were SE-ResNet-50 (best on RAFDB) and SE-ResNeXt-50 (best on AffectNet and FER2013), which I would expect given they both have the `squeeze-and-excitation' mechanic (so should beat the ResNet-50 network) and the latter has more efficient cell-blocks according to \cite{resnext_paper} whilst still having a similar number of parameters
    
    \item \textbf{Other Models}: VGG-Face and ResNet-50 performed reasonably well, the former having the advantage of being pre-trained on facial images, whilst the latter is a slightly deeper network with good information flow. The worst performers were DenseNet-121 and NASNet, which given their increased size may be due to the limited number of training epochs (particularly important for the latter given the size of the search space it must navigate)
    
\end{itemize}

\section{Fine-Tuning Visual} \label{fine_tune_exp}

The purpose of this stage is build on the general FER training carried out in section \ref{pre_train_exp} and make the models more specific to the EmotiW FER AV challenge. To do this we will now only use the AFEW dataset.

We will still employ some of the more successful practices learned above, but explore a greater range of ideas / techniques in order to achieve high accuracy. These are:

\begin{itemize}
    
    \item \textbf{Sequence Length}: Given we are no longer dealing with static images, but a sequence of images, we can decide how many frames to input to the network. Figures \ref{fig:train_lens} and \ref{fig:valid_lens} show the distribution of sequence lengths for the AFEW dataset. A clear issue is that the distributions do not match, the effect of this is further examined in section \ref{data_problems}. Therefore a few different input sequence lengths were tried
    
    \item \textbf{Aggregation}: Given an input sequence length of greater than 1, there are two main aggregation decisions, the (i) reduction method and (ii) type of calculation applied. These options were laid out and discussed in section \ref{training_process} of Chapter \ref{implementation}. The results of putting these options into effect can be seen below
    
    \item \textbf{Batch Size}: The benefits of `Stochastic Gradient Descent' and the impact of the batch size were explained in section \ref{deep_learning_basics}. Considering the small number of samples in the AFEW dataset, the batch size was adjusted accordingly for this section
    
    \item \textbf{Training Stages}: All weights are restored from networks trained in section \ref{pre_train_exp} (possible to do this because image input dimensions are the same). Given we have new data, but hopefully have already developed robust hierarchical feature extractors in the early layers, it may be beneficial to train solely the classification part of the network first (e.g. dense layers). If we allowed the model to update all layers from the start, when the loss should be it's greatest, the convolutional weights may update incorrectly (i.e. source of error mainly from fully connected layers that are yet to learn patterns in new AFEW dataset). Instead we can stagger the learning process, updating only the fully connected layers for a certain number of epochs, before switching to training the entire network
    
\end{itemize}

The above ideas / techniques can be combined in different ways with varying hyperparameters. Given the number of permutations possible, I have tried to be efficient by only running a selection that had a good chance of succeeding based on classic deep learning theory.

Note the * in the tables below indicates model runs based the highest non learning rate reducing run from Pre-Training Visual stage (e.g. ResNet-50 would be `Run Two' with global pool = False). 

\begin{center}
\begin{longtable}{ |p{1.75cm}||p{6cm}|p{3cm}|p{2cm}|}
 \hline
 \rowcolor{lightgray} \multicolumn{4}{|c|}{Table 1: VGG-Face} \\
 \hline \hline
 Run & Description & Metric & AFEW \\
 \hline \hline
 
 One* & LR=0.0001, reduce & Accuracy & 15.4\% \\
 & Seq Length = Per Frame & F1-Score & 3.8\% \\
 \hline \hline

 Two* & LR=0.0001, reduce & Accuracy & 32.9\% \\
 & Seq Length = 40 & F1-Score & 29.0\% \\
 \hline \hline
 
 \rowcolor{yellow} 
 Three* & LR=0.0001, reduce & Accuracy & 40.5\% \\
 \rowcolor{yellow} 
 & Seq Length = 80 & F1-Score & 37.0\% \\
 \hline \hline
 
 Four* & LR=0.0001, reduce & Accuracy & 38.6\% \\
 & Seq Length = 144 & F1-Score & 34.7\% \\
 \hline \hline

 Five & LR=0.0001, static & Accuracy & 38.6\% \\
 & Seq Length = 80 & F1-Score & 34.3\% \\
 \hline \hline
 
 \rowcolor{yellow} 
 Six & LR=0.00001, reduce & Accuracy & 45.7\% \\
 \rowcolor{yellow} 
 & Seq Length = 80 & F1-Score & 42.7\% \\
 \hline \hline

 \rowcolor{yellow} 
 Seven & LR=0.00001, reduce & Accuracy & 47.8\% \\
 \rowcolor{yellow} 
 & Seq Length = 80 & F1-Score & 45.3\% \\
 \rowcolor{yellow} 
 & Batch size = 4 & & \\
 \hline \hline

 Eight & LR=0.00001, reduce & Accuracy & 47.5\% \\
 & Seq Length = 80 & F1-Score & 45.0\% \\
 & Batch size = 8 & & \\
 \hline \hline
 
 Nine & LR=0.00001, reduce & Accuracy & 42.8\% \\
 & Seq Length = 80 & F1-Score & 38.1\% \\
 & Batch size = 4 & & \\
 & Reduce Mean & & \\
 \hline \hline
 
 Ten & LR=0.00001, reduce & Accuracy & 46.2\% \\
 & Seq Length = 80 & F1-Score & 43.4\% \\
 & Train epochs = 5, 25 & & \\
 \hline \hline
 
 Eleven & LR=0.00001, reduce & Accuracy & 45.7\% \\
 & Seq Length = 80 & F1-Score & 43.8\% \\
 & Train epochs = 10, 20 & & \\
 \hline \hline
 
 Twelve & LR=0.00001, reduce & Accuracy & 46.5\% \\
 & Seq Length = 80 & F1-Score & 43.5\% \\
 & Median (not reduced) & & \\
 \hline \hline
 
 \rowcolor{yellow} 
 Thirteen & LR=0.00001, reduce & Accuracy & 46.0\% \\
 \rowcolor{yellow} 
 & Seq Length = 80 & F1-Score & 42.0\% \\
 \rowcolor{yellow} 
 & Batch size = 4 & & \\
 \rowcolor{yellow} 
 & Reduce Mean & & \\
 \rowcolor{yellow} 
 & Train epochs = 5, 25 & & \\
 \hline \hline
 
 Fourteen & LR=0.00001, reduce & Accuracy & 45.4\% \\
 & Seq Length = 80 & F1-Score & 43.2\% \\
 & Batch size = 4 & & \\
 & Reduce Mean & & \\
 & Train epochs = 10, 20 & & \\
 \hline \hline
 
 Fifteen & LR=0.00001, reduce & Accuracy & 43.1\% \\
 & Seq Length = 80 & F1-Score & 40.8\% \\
 & Batch size = 4 & & \\
 & Reduce Median & & \\
 & Train epochs = 10, 20 & & \\
 \hline \hline
 
 Sixteen & LR=0.00001, reduce & Accuracy & 42.8\% \\
 & Seq Length = 80 & F1-Score & 38.1\% \\
 & Batch size = 4 & & \\
 & Reduce Mean & & \\
 & Train epochs = 5, 15 & & \\
 \hline \hline
 
 Seventeen & LR=0.00001, reduce & Accuracy & 44.1\% \\
 & Seq Length = 80 & F1-Score & 31.8\% \\
 & Batch size = 4 & & \\
 & Reduce Mean & & \\
 & Train epochs = 5, 25 & & \\
 & FC layer initialised & & \\

 \hline
 
\end{longtable}
\end{center}

 From the table above, I inferred the following about the general fine-tuning of CNN models on the AFEW dataset:
 
 \begin{itemize}
     
     \item Runs 1-4 indicate that an input sequence length of 80 performs best (i.e. `Run Three'). Based on figures \ref{fig:train_lens} and \ref{fig:valid_lens} summarising the sequence lengths, this value makes sense. There is a drop-off in the number of videos with sequence longer than 80 in the validation dataset, which is similar to the training dataset behaviour allowing the model to sufficiently learn
     
     \item `Run Six' shows the benefit of applying learning rate decay with a initial learning rate of 0.00001, so this becomes our new benchmark (applied to `Run Three' in the previous section). After this run, the different ideas / techniques explained above are trialled
     
     \item By comparison to `Run Six' increasing the batch size, training in different stages and using the median (non-reduced) seem to all have positive impacts. The biggest improvement comes from increasing the batch size to 4, with training for 5 then 25 epochs beating `Run Eleven'
     
     \item The only decrease comes in `Run 9' from reducing the sequence to the exact length, rather than using the full padded sequence. This result is surprising because one would expect repeating the last frame to distort the prediction. A possible explanation is that for a reasonably large proportion of the videos the last frame strongly exhibits the labelled emotion
     
     \item Combining the different elements of runs 7 to 12 can be seen in `Run Fifteen', where a decrease in accuracy has occurred. I hypothesised at the time that I would see a 1-2\% rise, with noise in the data and model being a possible explanation. Hence I reverted back to mean aggregation over the median in `Run Fourteen', which to me should produce better results as frames with large activation signals will have slightly more weight 
     
     \item The best of the runs combining the different ideas / techniques discussed at the start of this section was `Run Thirteen'. The accuracy is slightly higher than my benchmark of `Run Six' 
     
     \item Given the `F1-Score' is lower in `Run Thirteen' than `Run Fourteen' and combining ideas / techniques didn't show the expected benefit, I thought the network might be overfitting. Therefore I tried to reduce the number of epochs (see `Short Stopping' in section \ref{deep_learning_basics} as a form of regularisation) in `Run Sixteen', however, this led to another decrease in the accuracy
     
     \item Another avenue explored was to randomly re-initialise the fully connected layers, rather than fully restoring them (as well the convolutional layers) from the weights in section \ref{fine_tune_exp}. The idea being that the models hopefully have developed robust hierarchical feature extractors in the early layers, but given we have new data, we want the classification part to learn the unknown patterns in the data. However, as can be seen in `Run Seventeen' there was a slight dip in performance. This approach may work for longer training periods (e.g. 50+ epochs)
     
 \end{itemize}

In the absence of more time to continue fine-tuning or repeating runs to smooth out results, I decided to proceed with the set-up of `Run Thirteen' for the other models rather than the run with the highest accuracy (i.e. `Run Seven'). The reason being that rationally I would expect training in stages and reducing the sequence to have positive impacts, with model and data noise possibly explaining some of the gap in performance. Ultimately, these CNNs will be used as feature extractors for the RNN, hence accuracy may not be the best gauge at this juncture because the strength and robustness of the convolutional layers is more important, rather than the classification (largely influenced by the final layers). 

Also, because an aim of this project is to provide a comparison between the 6 different CNNs, I slightly took into account what I thought would work well for those networks. For example, training in stages for 5 and then 25 epochs should logically suit the other models because they each only have 1 fully connected layer, hence less training time is needed. 

The results tables below for the other models includes runs based on `Run Three' (equivalent to `Run One' below) and `Run Thirteen' (equivalent to `Run Two' below).

\begin{center}
\begin{tabular}{ |p{1.5cm}||p{6cm}|p{3cm}|p{2cm}|}
 \hline
 \rowcolor{lightgray} \multicolumn{4}{|c|}{Table 2: ResNet-50} \\
 \hline \hline
 Run & Description & Metric & AFEW \\
 \hline \hline
 
 One* & LR=0.00001, reduce & Accuracy & 42.8\% \\
 & Seq Length = 80 & F1-Score & 37.2\% \\
 \hline \hline

 Two & LR=0.00001, reduce & Accuracy & 41.5\% \\
 & Seq Length = 40 & F1-Score & 39.0\% \\
 & global pool = False & & \\
 \hline \hline
 
 Three & LR=0.00001, reduce & Accuracy & 41.3\% \\
 & Seq Length = 40 & F1-Score & 34.5\% \\
 & global pool = True & & \\

 \hline
 
\end{tabular}
\end{center}

\begin{center}
\begin{tabular}{ |p{1.5cm}||p{6cm}|p{3cm}|p{2cm}|}
 \hline
 \rowcolor{lightgray} \multicolumn{4}{|c|}{Table 3: DenseNet-121} \\
 \hline \hline
 Run & Description & Metric & AFEW \\
 \hline \hline
 
 One* & LR=0.00001, reduce & Accuracy & 31.3\% \\
 & Seq Length = 80 & F1-Score & 24.9\% \\
 \hline \hline

 Two & LR=0.00001, reduce & Accuracy & 32.6\% \\
 & Seq Length = 40 & F1-Score & 24.6\% \\

 \hline
 
\end{tabular}
\end{center}

\begin{center}
\begin{tabular}{ |p{1.5cm}||p{6cm}|p{3cm}|p{2cm}|}
 \hline
 \rowcolor{lightgray} \multicolumn{4}{|c|}{Table 4: SE-ResNet-50} \\
 \hline \hline
 Run & Description & Metric & AFEW \\
 \hline \hline
 
 One* & LR=0.00001, reduce & Accuracy & 46.0\% \\
 & Seq Length = 80 & F1-Score & 42.5\% \\
 \hline \hline

 Two & LR=0.00001, reduce & Accuracy & 42.8\% \\
 & Seq Length = 80 & F1-Score & 37.9\% \\
 & global pool = False & & \\
 \hline \hline
 
 Three & LR=0.00001, reduce & Accuracy & 39.9\% \\
 & Seq Length = 40 & F1-Score & 35.0\% \\
 & global pool = True & & \\
 
 \hline
 
\end{tabular}
\end{center}

Note ** indicates that due to exhausted memory, smaller batches had to be applied, batch size of 2 for table 5 and batch size of 1 for table 6.

\begin{center}
\begin{tabular}{ |p{1.5cm}||p{6cm}|p{3cm}|p{2cm}|}
 \hline
 \rowcolor{lightgray} \multicolumn{4}{|c|}{Table 5: SE-ResNeXt-50**} \\
 \hline \hline
 Run & Description & Metric & AFEW \\
 \hline \hline
 
 One* & LR=0.00001, reduce & Accuracy & 35.2\% \\
 & Seq Length = 80 & F1-Score & 33.6\% \\
 \hline \hline

 Two & LR=0.00001, reduce & Accuracy & 38.9\% \\
 & Seq Length = 80 & F1-Score & 34.1\% \\

 \hline
 
\end{tabular}
\end{center}

\begin{center}
\begin{tabular}{ |p{1.5cm}||p{6cm}|p{3cm}|p{2cm}|}
 \hline
 \rowcolor{lightgray} \multicolumn{4}{|c|}{Table 6: NASNet**} \\
 \hline \hline
 Run & Description & Metric & AFEW \\
 \hline \hline
 
 One* & LR=0.00001, reduce & Accuracy & 37.9\% \\
 & Seq Length = 80 & F1-Score & 30.3\% \\
 \hline \hline

 Two & LR=0.00001, reduce & Accuracy & 37.1\% \\
 & Seq Length = 80 & F1-Score & 29.5\% \\

 \hline

\end{tabular}

\end{center}

The impact of applying the ideas / techniques discussed at the start of this section varies across the different models (i.e. `Run One' the benchmark vs. `Run Two' the combined). Tables 3 and 5 showed a small increase in accuracy and f1-score, but tables 2 and 4 showed a drop compared to the benchmark. Note that the latter two models have global pool = False, so one possible explanation is the feature map with this setting becomes too large to train on the AFEW dataset. Hence an extra run was included in tables 2 and 4 with the setting `global pool = True', with a fall in accuracy for SE-ResNet-50 and marginal decrease for ResNet-50.

\begin{figure}[htbp]
    \centering
    \includegraphics[width=0.8\textwidth]{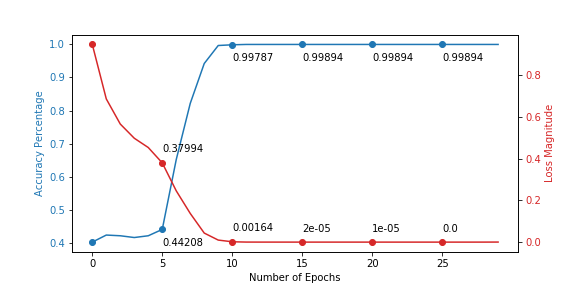}
    \caption{VGG-Face Training Accuracy and Loss results for `Run Thirteen'}
    \label{fig:stage_2_training}
    
\end{figure}

\begin{figure}[htbp]
    \centering
    \includegraphics[width=0.8\textwidth]{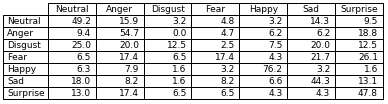}
    \caption{VGG-Face confusion matrix for `Run Thirteen'}
    \label{fig:stage_2_cm}
    
\end{figure}

The training graph in figure \ref{fig:stage_2_training} and confusion matrix in figure \ref{fig:stage_2_cm} for `Run Thirteen' of table 1 demonstrates that:

\begin{itemize}
    
    \item The model learns very quickly after 5 epochs (i.e. when all layers are updating, rather than just the fully connected layers), with accuracy and loss improving dramatically and then flattening off. After 10 epochs the accuracy and loss change marginally, suggesting overfitting may start to become an issue, but the results of `Run Fifteen' of table 1 suggest this is not the case for 30 epochs
    
    \item The confusion matrix tells a similar story to section \ref{pre_train_exp}, with `Happy' performing best, with `Disgust' the worst. This could be the same errors being carried forward in the model since weights are re-used, but the results from other FER papers are consistent with this finding
    
\end{itemize}

\begin{center}
\begin{tabular}{ |p{2cm}||p{3cm}|p{2cm}|}
 \hline
 \rowcolor{lightgray} \multicolumn{3}{|c|}{Table 7: Other 2018 Entrants} \\
 \hline \hline
 Paper & Model & AFEW \\
 \hline \hline
 
 \cite{fan} & VGG-Face &  45.2\% \\
 & ResNet-50 &  40.1\% \\
 & DenseNet-121 & 44.1\% \\
 \hline \hline

 \cite{lu} & VGG-Face & 50.3\% \\
 & ResNet-50 & 48.1\% \\

 \hline
 
\end{tabular}
\end{center}

Given the results above, the following conclusions can be made which will be carried forward to the next stage:

\begin{itemize}

    \item \textbf{Dynamic}: Comparing the accuracy results of section \ref{pre_train_exp} to those above, it is clear that the dynamic FER task is significantly harder than just classifying static images. The main difference being, static images will certainly display the labelled emotion, where as only a few frames may in video sequences
    
    \item \textbf{Overfitting}: As expected overfitting is something to bear-in-mind when training models on the AFEW dataset given it's limited size, this will be monitored in the next stage of training. The large difference in accuracy between the training and evaluation runs also typically suggests overfitting, but this could just be a consequence of the limited training data available / size of the validation dataset relative to the training dataset (i.e. roughly 50\% is high compared to datasets in section \ref{pre_train_exp})
    
    \item \textbf{Entry Comparison}: Table 7 includes published results for the AFEW validation dataset for solely CNN networks from the 2018 EmotiW FER AV competition. \cite{fan} accuracy levels are slightly lower (with the exception of the DenseNet-121 model) and \cite{lu} results are slightly better. In both cases, extensive data augmentation is applied and the networks are adapted to produce final predictions, but it does indicate the models training this paper are performing relatively well 
    
    \item \textbf{Best Model(s)}: As discussed earlier in this section, although some runs performed better (i.e. `Run Seven' was the highest in table 1), our focus at this stage is to find the best feature extractors for the RNNs, which logically I believe the combined run will do (i.e. reduced mean, train in stages and batch size of 4) 
    
    \item \textbf{Model Trend}: At this point it seems that two groups are forming based on performance, a higher accuracy group (VGG-Face, ResNet-50, SE-ResNet-50 and SE-ResNeXt-50) and a lower accuracy group (DenseNet-121 and NASNet)  
    
\end{itemize}

\section{Temporal Visual} \label{temporal_exp}

This training stage is designed to better capture the temporal relationship between the frames in a sequence. As explained in section \ref{seq_model_imp}, the CNNs are first used to extract a feature map from the raw image data, with this being inputted to the RNN for the final sequence classification. The tables correspond directly to the combined visual model from the last section (e.g. `Run Thirteen' for VGG-Face), with the type of GRU detailed in the description of the run.

The following settings were used for all runs below, unless otherwise stated:

\begin{itemize}

    \item Reduce mean applied for the reasons outlined in section \ref{fine_tune_exp}
    
    \item Batch size of 4 achieved the best results in the previous training stage, so has been used again here
    
    \item Training solely the RNN layer is carried out for 10 epochs, before the whole network is trained for 30 epochs. The rationale being there are more weights in the RNN layer (see section \ref{sequence_models}) than in a single fully connected layer, therefore more training time is required. Also, the convoluational layer weight have been fine-tuned already for the AFEW dataset, so limiting further updates may be positive
    
    \item Initial learning rate of 0.00001 was seen to work best in table 1 of section \ref{fine_tune_exp}, with early tests (not included below) showing a similar benefit for the RNN training stage
    
    \item The GRU implemented has 2 layers, with each cell having 128 hidden units. This architecture follows the optimal set-up deployed in the paper \cite{multicomp} \cite{multicomp1}
    
    \item The VGG-Face model has 3 fully connected layers, the first of these (4096-dimensional feature map) is the input to the RNN based on the findings in the paper \cite{affwild1}. The other networks only have one dense layer, so this is used as the input to the RNN. The output of the CNN models is controlled by the `$rnn\_seq$' in the training script

\end{itemize}

\begin{center}
\begin{longtable}{ |p{1.5cm}||p{6cm}|p{3cm}|p{2cm}|}
 \hline
 \rowcolor{lightgray} \multicolumn{4}{|c|}{Table 1: VGG-Face} \\
 \hline \hline
 Run & Description & Metric & AFEW \\
 \hline \hline
 
 One & LR=0.00001, reduce & Accuracy & 41.8\% \\
 & Seq Length = 16 & F1-Score & 37.9\% \\
 & Batch size = 4 & & \\
 & Train epochs = 10, 20 & & \\
 & Reduce Mean & & \\
 \hline \hline

 Two & LR=0.00001, reduce & Accuracy & 43.6\% \\
 & Seq Length = 32 & F1-Score & 40.0\% \\
 \hline \hline
 
 \rowcolor{yellow}
 Three & LR=0.00001, reduce & Accuracy & 44.4\% \\
 \rowcolor{yellow}
 & Seq Length = 40 & F1-Score & 40.6\% \\
 \hline \hline
 
 Four & LR=0.00001, reduce & Accuracy & 43.1\% \\
 & Seq Length = 60 & F1-Score & 40.2\% \\
 \hline \hline

 Five & LR=0.00001, reduce & Accuracy & 43.1\% \\
 & Seq Length = 80 & F1-Score & 39.8\% \\
 \hline \hline
 
 Six & LR=0.00001, reduce & Accuracy & 44.4\% \\
 & Seq Length = 144 & F1-Score & 41.3\% \\
 \hline \hline
 
 \rowcolor{yellow}
 Seven & LR=0.00001, reduce & Accuracy & 45.4\% \\
 \rowcolor{yellow}
 & Seq Length = 40 & F1-Score & 42.6\% \\
 \rowcolor{yellow}
 & Bi-Directional & & \\
 \hline \hline

 Eight & LR=0.00001, reduce & Accuracy & 42.8\% \\
 & Seq Length = 40 & F1-Score & 39.3\% \\
 & Attention Mechanism & & \\
 \hline \hline
 
 Nine & LR=0.00001, reduce & Accuracy & 41.8\% \\
 & Seq Length = 40 & F1-Score & 40.1\% \\
 & Bi-Directional & & \\
 & Attention Mechanism & & \\
 
 \hline
 
\end{longtable}
\end{center}

There are two main things being investigated in table 1:

\begin{itemize}
    
    \item \textbf{Sequence Length}: The CNN is just a feature extractor, so the sequence length used for the CNN (outputs are aggregated after the convolutional layers in section \ref{fine_tune_exp}) does not need to match that of the RNN. From the first five runs, the best performing is `Run Three', with the accuracy falling away on either side (helping to support the conclusion that sequence length of 40 is optimal). An explanation of this behaviour is the trade-off between the `long-term complex dependencies' (discussed in section \ref{sequence_models}) and amount of information being inputted to the RNN network. If a sequence is too long the GRU's memory is insufficient to understand patterns in the data and too short then the memory attribute is not being fully utilised. The accuracy jumps back up in `Run Six', which takes the whole sequence as the input, which has the advantage of not having to combine the different sequence results to provide a prediction for the whole video. Given the memory issues experienced in this project, particularly for the larger models, sequence length of 40 will be used in the rest of this section. Note competition entrants from last year used sequence lengths of 8 and 16, this might be because the image sizes and / or number of hidden units used was greater, hence the RNN struggled to efficiently learn for a sequence length of 40
    
    \item \textbf{RNN Model}: The last 3 runs show the performance of the different RNN models, with the Bi-directional GRU in `Run Seven' having the highest accuracy. The advantage being the network is able to use future and historic context to improve the predictive power. Including the attention mechanism seems to have a negative effect, possibly due to the additional parameters involved in the training process
    
\end{itemize}

Although the accuracy for `Run Seven' is the same as that of the underlying CNN model in the previous section, the `F1-Score' is slightly higher suggesting a slight improvement in performance. 

\begin{center}
\begin{tabular}{ |p{1.5cm}||p{6cm}|p{3cm}|p{2cm}|}
 \hline
 \rowcolor{lightgray} \multicolumn{4}{|c|}{Table 2: ResNet-50} \\
 \hline \hline
 Run & Description & Metric & AFEW \\
 \hline \hline
 
 One & LR=0.00001, reduce & Accuracy & 37.9\% \\
 & Seq Length = 40 & F1-Score & 33.1\% \\
 & global pool = False & & \\
 \hline \hline

 Two & LR=0.00001, reduce & Accuracy & 41.3\% \\
 & Seq Length = 40 & F1-Score & 34.9\% \\
 & global pool = True & & \\
 \hline \hline

 \rowcolor{yellow}
 Three & LR=0.0001, reduce & Accuracy & 41.8\% \\
 \rowcolor{yellow}
 & Seq Length = 40 & F1-Score & 34.5\% \\
 \rowcolor{yellow}
 & global pool = True & & \\
 \rowcolor{yellow}
 & Bi-Directional & & \\

 \hline
 
\end{tabular}
\end{center}

\begin{center}
\begin{tabular}{ |p{1.5cm}||p{6cm}|p{3cm}|p{2cm}|}
 \hline
 \rowcolor{lightgray} \multicolumn{4}{|c|}{Table 3: DenseNet-121} \\
 \hline \hline
 Run & Description & Metric & AFEW \\
 \hline \hline
 
 One & LR=0.0001, reduce & Accuracy & 32.4\% \\
 & Seq Length = 40 & F1-Score & 23.5\% \\

 \hline
 
\end{tabular}
\end{center}

\begin{center}
\begin{tabular}{ |p{1.5cm}||p{6cm}|p{3cm}|p{2cm}|}
 \hline
 \rowcolor{lightgray} \multicolumn{4}{|c|}{Table 4: SE-ResNet-50} \\
 \hline \hline
 Run & Description & Metric & AFEW \\
 \hline \hline
 
 One & LR=0.00001, reduce & Accuracy & 44.1\% \\
 & Seq Length = 40 & F1-Score & 38.7\% \\
 & global pool = False & & \\
 
 \hline \hline

 Two & LR=0.00001, reduce & Accuracy & 41.8\% \\
 & Seq Length = 40 & F1-Score & 33.8\% \\
 & global pool = True & & \\

 \hline \hline
 
 \rowcolor{yellow}
 Three & LR=0.00001, reduce & Accuracy & 46.0\% \\
 \rowcolor{yellow}
 & Seq Length = 40 & F1-Score & 40.6\% \\
 \rowcolor{yellow}
 & global pool = False & & \\
 \rowcolor{yellow}
 & Bi-Directional & & \\
 
 \hline
 
\end{tabular}
\end{center}

Note ** indicates that due to exhausted memory, smaller batches had to be applied, batch size of 2 for table 5 and batch size of 1 for table 6.

\begin{center}
\begin{tabular}{ |p{1.5cm}||p{6cm}|p{3cm}|p{2cm}|}
 \hline
 \rowcolor{lightgray} \multicolumn{4}{|c|}{Table 5: SE-ResNeXt-50**} \\
 \hline \hline
 Run & Description & Metric & AFEW \\
 \hline \hline
 
 \rowcolor{yellow}
 One & LR=0.00001, reduce & Accuracy & 43.6\% \\
 \rowcolor{yellow}
 & Seq Length = 40 & F1-Score & 39.6\% \\
 \rowcolor{yellow}
 & global pool = True & & \\
 \hline \hline
 
 Two & LR=0.00001, reduce & Accuracy & 43.1\% \\
 & Seq Length = 80 & F1-Score & 40.6\% \\
 & Bi-Directional & & \\

 \hline
 
\end{tabular}
\end{center}

\begin{center}
\begin{tabular}{ |p{1.5cm}||p{6cm}|p{3cm}|p{2cm}|}
 \hline
 \rowcolor{lightgray} \multicolumn{4}{|c|}{Table 6: NASNet**} \\
 \hline \hline
 Run & Description & Metric & AFEW \\
 \hline \hline
 
 One & LR=0.0001, reduce & Accuracy & 37.9\% \\
 & Seq Length = 40 & F1-Score & 35.0\% \\

 \hline

\end{tabular}

\end{center}

The results for the other models are once again mixed, with the following findings:

\begin{itemize}
    
    \item DenseNet-121 and NASNet models continue to under perform, suggesting they are not well-suited to this FER task. Possible explanations being, for the former the full connectivity introduces additional parameters that perhaps cannot be efficiently learnt for the AFEW dataset given its limited size and for the latter 30 epochs is not long enough to exhaustively explore the search space to find the optimal permutation of convolutional cells (size of dataset might also be factor)
    
    \item The other 3 models show a slight accuracy improvement for their respective best runs in this section compared to the underlying CNN performance recorded in section \ref{fine_tune_exp}. In particular the SE-ResNet-50 and SE-ResNeXt-50 exhibit considerable jumps in accuracy, suggesting that introducing the `squeeze-and-excitation' technique produces well-defined feature maps that the RNN can efficiently exploit for classification
    
    \item The Bi-directional GRU resulted in higher accuracy levels for ResNet-50 and SE-ResNet-50 models, but lower for SE-ResNeXt-50. The could possibly to do with the number of parameters in each model, with the `Bi-directional' approach doubling the number of parameters for the RNN network and becoming too many weights to efficiently learn for the larger models
    
    \item For the ResNet-50 and SE-ResNet-50 models, we explore the impact of the `global pool' parameter following the findings in section \ref{fine_tune_exp}. It seems that `global pool = True' (i.e. reduce the feature map) substantially improves performance for the former model, but worsens accuracy for the latter. This could be due to the quality of the feature maps produced, as discussed in section \ref{fine_tune_exp}, it may help average out inconsistencies in lower quality feature maps (i.e. ResNet-50) or lose important information by aggregating for higher quality feature maps (i.e. SE-ResNet-50)
    
\end{itemize}

\begin{figure}[htbp]
    \centering
    \includegraphics[width=0.8\textwidth]{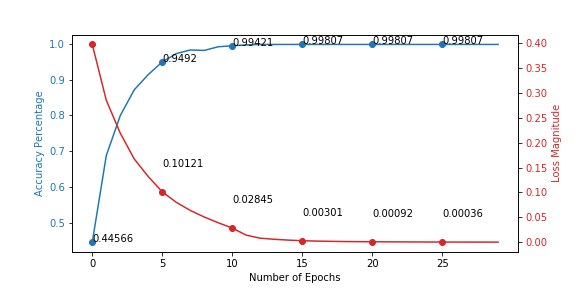}
    \caption{SE-ResNet-50 Training Accuracy and Loss results for `Run One'}
    \label{fig:stage_3_training}
    
\end{figure}

\begin{figure}[htbp]
    \centering
    \includegraphics[width=0.8\textwidth]{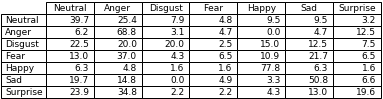}
    \caption{SE-ResNet-50 confusion matrix for `Run One'}
    \label{fig:stage_3_cm}
    
\end{figure}

The training graph in figure \ref{fig:stage_3_training} and confusion matrix in figure \ref{fig:stage_3_cm} for ‘Run One’of table 4 indicates similar behaviour to the last section. The only main difference being the model learns slightly quicker in the first 10 epoch training stage, which is a result of already good feature extraction.

Given the results above, the following conclusions can be made which will be carried forward to the next stage:

\begin{itemize}
    
    \item \textbf{Model Choices}: The DenseNet-121 and NASNet models will be dropped given their noticeably lower performance, meaning only 4 models will now be utilised
    
    \item \textbf{Comparison to CNNs}: The best RNN runs in the tables above are similar in performance to the underlying CNNs for VGG-Face and ResNet-50, but are better for SE-ResNet-50 and SE-ResNeXt-50. The per frame CNN outputs in the previous section are aggregated, which is equivalent to to giving equal weights to each classification in time. The RNN is trying to model the temporal relationship between the frames better, i.e. boosting activations for frames where the input and surrounding frames suggest a certain emotion is being displayed, whilst suppressing activations where neighbouring frames disagree with a specific frame in the sequence. Given the RNN has more weights than a single fully connected layer (VGG-Face has 3 FC layers, so is the exception), this temporal relationship may be difficult to learn for small datasets. However, for high quality feature maps, it is easier for the RNN to understand the temporal behaviour and hence there is an increase accuracy. The paper \cite{occam} also noted limited benefit of adding an RNN for the ResNet model
    
    \item \textbf{Overfitting}: The training and evaluation results for the temporal models are similar to those in section \ref{fine_tune_exp}, hence the same fears of `Overfitting' remain. Data augmentation and including the audio information would help reduce this risk 
    
    \item \textbf{Best Model(s)}: The `Bi-directional GRU' for the SE-ResNet-50 model produces the highest accuracy of 46.0\%. The same run for the VGG-Face CNN is only just behind, but has a slightly higher `F1-score'
    
    \item \textbf{Other Model(s)}: The GRU run of the SE-ResNext-50 is the next best (which may have performed better had the batch size not been limited to 2), with ResNet-50 fourth. It seems that newer versions of the ResNet architecture family perform better on this FER challenge, but in the case of SE-ResNeXt it is difficult to know the impact of the cardinality (see section \ref{CNNs}) vs. the `squeeze-and-excitation' technique without further tests
    
\end{itemize}

\section{Audio} \label{audio_exp}

There are two approaches discussed in section \ref{audio_design} to classify the video sequences solely based on the audio modality. The results and analysis in this section are split accordingly.

\subsubsection{Whole Video}

The openSMILE tool is applied to the entire raw-audio data, with a feature map of dimension 6,552 (i.e. `functionals') inputted to a `Forward-Neural Network'. The paper \cite{occam} found that a layer with ``64 units, batch-normalization, dropout and ReLu" performed the best. Given we have used a different configuration file (e.g. `$emo\_large$') to produce the feature map, I will explore different architectures to find what works best in our case. 

The accuracy on the AFEW validation datasets published in the papers \cite{occam} and \cite{liu} (different approaches but openSMILE used in both cases) were 33.5\% (additional training on the IEMOCAP dataset) and 31.1\% (PCA applied before classifier). 

To produce the results below I have used dropout (with a rate of 0.5) throughout, because I felt regularisation was important to prevent overfitting. Other parameter considerations were:

\begin{itemize}
    
    \item \textbf{Batch Size} The batch size used (unless otherwise stated) was 16. This decision is based on (i) the data input being less memory intensive than the images, (ii) the audio data being potentially noiser and (iii) more training epochs being used for certain runs
    
    \item \textbf{Hidden Units}: The feature map has dimension 6,552, which is much larger than that used in \cite{occam}, so wider layers will be explored to see if they better learn the underlying structure 
    
    \item \textbf{Training Epochs}: The feature map input is less memory intensive and the models used are smaller in size, hence the number of epochs can easily be increased without significantly impacting the project pipeline efficiency (e.g. run takes minutes rather than the hours required for the visual models to finish)
    
\end{itemize}

\begin{center}
\begin{longtable}{ |p{1.5cm}||p{6cm}|p{3cm}|p{2cm}|}
 \hline
 \rowcolor{lightgray} \multicolumn{4}{|c|}{Table 1: Whole Video} \\
 \hline \hline
 Run & Description & Metric & AFEW \\
 \hline \hline
 
 One & LR=0.0001, reduce & Accuracy & 15.9\% \\
 & Hidden Units = 128 & F1-Score & 9.5\% \\
 & Batch size = 4 & & \\
 & Train epochs = 30 & & \\
 & Dropout & & \\
 \hline \hline

 Two & LR=0.0001, reduce & Accuracy & 13.3\% \\
 & Hidden Units = 256 & F1-Score & 10.0\% \\
 \hline \hline
 
 Three & LR=0.00001, reduce & Accuracy & 22.7\% \\
 & Hidden Units = 256 & F1-Score & 19.0\% \\
 & Batch size = 16 & & \\
 & Train epochs = 100 & & \\
 \hline \hline
 
 Four & LR=0.00001, reduce & Accuracy & 24.8\% \\
 & Hidden Units = 512 & F1-Score & 21.2\% \\
 \hline \hline

 \rowcolor{yellow}
 Five & LR=0.00001, reduce & Accuracy & 27.4\% \\
 \rowcolor{yellow}
 & Hidden Units = 1024 & F1-Score & 23.5\% \\
 \hline \hline
 
 Six & LR=0.00001, reduce & Accuracy & 26.6\% \\
 & Seq Length = 2048 & F1-Score & 22.7\% \\
 \hline \hline

 Seven & LR=0.00001, reduce & Accuracy & 24.8\% \\
 & Hidden Units = 1024 & F1-Score & 21.9\% \\
 & Batch size = 32 & & \\
 & Train epochs = 200 & & \\
 \hline \hline

 Eight & LR=0.00001, reduce & Accuracy & 21.7\% \\
 & Hidden Units = 1024 & F1-Score & 12.0\% \\
 & Batch size = 16 & & \\
 & Train epochs = 100 & & \\
 & Batch Normalisation & & \\
 \hline \hline
 
 Nine & LR=0.00001, reduce & Accuracy & 11.2\% \\
 & Hidden Units = 256, 64 & F1-Score & 7.8\% \\
 & Batch size = 16 & & \\
 & Train epochs = 100 & & \\
 
 \hline
 
\end{longtable}
\end{center}

It seems that for the `$emo\_large$' openSMILE feature map, the optimal number of hidden units is 1,024 (i.e. `Run Five'). Although, the accuracy does decrease from `Run One' to `Run Three', the F1-score increases suggesting that models with fewer hidden units underfit (e.g. `Run One' predicts `Neutral' almost always, which gives a higher accuracy than `Run Three' because of class imabalance). Hence increasing model complexity was explored (e.g. number of hidden units), which decreasing the learning rate and number of training epochs supports. 

Given the size of the AFEW dataset, model complexity cannot be increase too much (i.e. poor performance for 2,048 hidden units and multiple layers). Also, including batch-normalization further regularises (see section \ref{regularization}) the model (decreases model complexity) and therefore we see a slight drop in performance when applied in `Run Nine'.

\begin{figure}[htbp]
    \centering
    \includegraphics[width=0.8\textwidth]{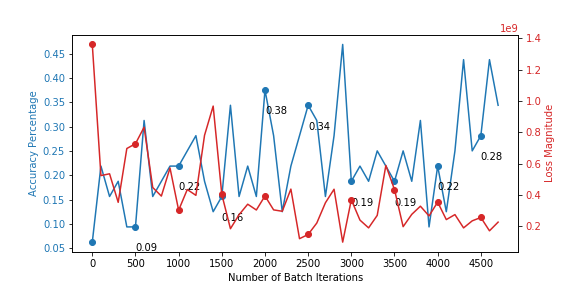}
    \caption{FFN Training Accuracy and Loss results for `Run Five'}
    \label{fig:stage_4_1_training}
    
\end{figure}

\begin{figure}[htbp]
    \centering
    \includegraphics[width=0.8\textwidth]{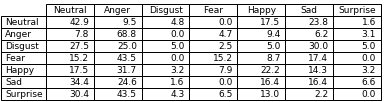}
    \caption{FFN confusion matrix for `Run Five'}
    \label{fig:stage_4_1_cm}
    
\end{figure}

The accuracy and loss results seen in figure \ref{fig:stage_4_1_training} are far more volatile during training that for the visual models. This reflects the greater noise in the audio data, but also the effect of applying a shallow network with dropout.

The confusion matrix for `Run Five' (figure \ref{fig:stage_4_1_cm}) produces some interesting results when compared to those for the visual models. The audio model performs as well on the `Angry' emotion (easily detected by shouting for example) and `Neutral' (this could be due to class imbalance or the model guessing `Neutral' if uncertain). It seems to slightly outperform the SE-ResNet-50 network for the `Fear' emotion, but otherwise performs poorly on the other classes.

Given  the  results  above,  I  have made  the  following  conclusions:

\begin{itemize}
    
    \item \textbf{EmotiW Comparison}: It seems my best results are slightly behind the two papers utilising openSMILE in last year's challenge. However, they used other techniques to boost performance and I believe given more time to fine-tune the model higher accuracy could be achieved. Also, using a larger feature map (i.e. 6,552 vs. 1,582 dimensions) may hinder the classifier if those additional features included are not as strong
    
    \item \textbf{Model Independence}: This audio model follows a very different approach to the visual models in previous sections. Although the performance may not be as high, when combining the predictions in the `late fusion' stage there may still be a benefit because of this independence 
    
\end{itemize}

\subsubsection{Video Clips}

The LLD feature maps outputted from openSMILE for the audio clips (see section \ref{audio_imp}) are of dimension 112 per frame. Given the feature map is not too large, it can be inputted directly to the GRU network to capture the temporal relationship.

There are no performance comparisons to last year's EmotiW entrants, but clearly the results can be measured against the accuracy levels in the section above. 

A main consideration for choosing the below parameters was, `what will create outputs that complement the visual model outputs?'. Therefore when combined we have a distinct and diverse new feature map for the final classifier.

The main parameters available to vary were:

\begin{itemize}
    
    \item \textbf{Sequence Length}: Similar to the choice presented for the visual models. There exists a trade-off between the amount of data to include and the memory capabilities of the GRU network. In addition, we must consider what sequence length would work best for `early fusion' given that the visual model is likely to be the stronger predictor and the memory available to run the combined models
    
    \item \textbf{Hidden Units}: The ratio of hidden units to size of the input feature map is key, too small and the model may lead to high bias, but too large (i.e. greater than 112) and the GRU may learn inefficiently. Hence a parameters of 32, 64, and 128 were tried
    
    \item \textbf{Additional Complexity}: A concern of the last section was `Underfitting' given the openSMILE feature map output. To combat this effect, we can add layers and also include an attention mechanism 
    
\end{itemize}

\begin{center}
\begin{longtable}{ |p{1.5cm}||p{6cm}|p{3cm}|p{2cm}|}
 \hline
 \rowcolor{lightgray} \multicolumn{4}{|c|}{Table 2: Video Clips} \\
 \hline \hline
 Run & Description & Metric & AFEW \\
 \hline \hline
 
 One & LR=0.001, static & Accuracy & 13.6\% \\
 & Hidden Units = 32 & F1-Score & 8.5\% \\
 & Sequence Length = 80 & & \\
 & Batch size = 16 & & \\
 & Train epochs = 30 & & \\
 & Layers = 2 & & \\
 \hline \hline

 Two & LR=0.001, static & Accuracy & 14.1\% \\
 & Hidden Units = 64 & F1-Score & 8.9\% \\
 \hline \hline
 
 Three & LR=0.001, static & Accuracy & 14.1\% \\
 & Hidden Units = 128 & F1-Score & 4.9\% \\
 \hline \hline
 
 Four & LR=0.001, reduce & Accuracy & 15.1\% \\
 & Hidden Units = 128 & F1-Score & 8.1\% \\
 \hline \hline

 \rowcolor{yellow}
 Five & LR=0.001, reduce & Accuracy & 16.7\% \\
 \rowcolor{yellow}
 & Hidden Units = 64 & F1-Score & 10.2\% \\
 \rowcolor{yellow}
 & Attention & & \\
 \rowcolor{yellow}
 & Layers = 2 & & \\
 \hline \hline
 
 \rowcolor{yellow}
 Six & LR=0.00001, reduce & Accuracy & 20.1\% \\
 \rowcolor{yellow}
 & Hidden Units = 64 & F1-Score & 8.6\% \\
 \rowcolor{yellow}
 & Attention & & \\
 \rowcolor{yellow}
 & Layers = 4 & & \\
 \hline \hline
 
 Seven & LR=0.00001, reduce & Accuracy & 17.0\% \\
 & Hidden Units = 64 & F1-Score & 7.7\% \\
 & Sequence Length = 80 & & \\
 & Train epochs = 50 & & \\
 & Attention & & \\
 & Layers = 4 & & \\
 \hline \hline

 Eight & LR=0.00001, reduce & Accuracy & 15.7\% \\
 & Hidden Units = 64 & F1-Score & 6.1\% \\
 & Sequence Length = 40 & & \\
 & Batch size = 16 & & \\
 & Train epochs = 50 & & \\
 & Layers = 4 & & \\
 \hline \hline
 
 Nine & LR=0.00001, reduce & Accuracy & 17.8\% \\
 & Hidden Units = 64 & F1-Score & 11.0\% \\
 & Sequence Length = 144 & & \\
 & Attention & & \\
 & Layers = 4 & & \\
 
 \hline
 
\end{longtable}
\end{center}

The performance metrics above suggest that the optimal number of hidden units is 64, with `Run Two' having the same accuracy but higher F1-score than `Run Three'. Also, similar to the `Whole Video' model findings, increasing complexity seems to improve results (best achieved in `Run Six' with both increased layers and the attention mechanism) until a saturation point is reached and then accuracy declines (i.e. drop see for `Run Seven'). 

Given the low accuracy scores, one has to be careful about hideen `Underfitting', an example of this is `Run Eight', which predicts `Neutral' pretty much exclusively. Since this is one of the larger categories in both the training and validation AFEW dataset, the accuracy looks relatively acceptable compared to other runs. However, if this feature map were combined with the visual model output, it may hinder performance rather than enhance it. A number of different parameter settings were explored for sequence length of 40, partly because this worked best for the visual models. Unfortunately, the network was not able to achieve strong results.

The F1-score for the `Run Nine', which captures the whole sequence is the highest seen, therefore with further fine-tuning I believe that sequence length of 144 could produce the highest accuracy (i.e. if a similar behaviour to that seen between `Run Five' and `Run Six' is exhibited).

\begin{figure}[htbp]
    \centering
    \includegraphics[width=0.8\textwidth]{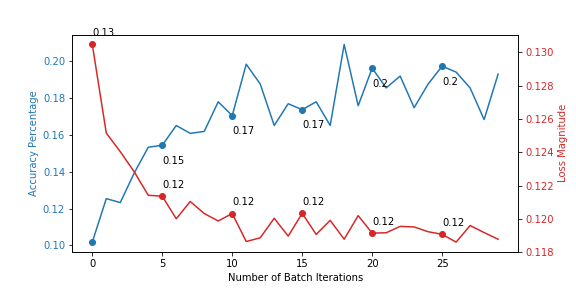}
    \caption{Audio GRU Training Accuracy and Loss results for `Run Six'}
    \label{fig:stage_4_2_training}
    
\end{figure}

Based on the training and loss path seen in figure \ref{fig:stage_4_2_training}, it seems that the model is still learning after 30 epochs. Also, the closeness between the training accuracy and evaluation accuracy (c. 20\%) suggests that `Overfitting' is certainly not yet a factor. Therefore, it is possible that some further training would lead to an increase in accuracy, but that the optimal point is below 50 epochs (i.e. drop in performance for `Run Eight').  

\begin{figure}[htbp]
    \centering
    \includegraphics[width=0.8\textwidth]{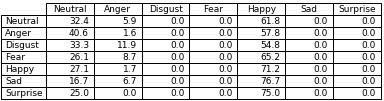}
    \caption{Audio GRU confusion matrix for `Run Six'}
    \label{fig:stage_4_2_cm}
    
\end{figure}

Given the results above, the following conclusions can be made which will be carried forward to the next stage:

\begin{itemize}
    
    \item \textbf{Class Performance}: The confusion matrix shows that this audio model struggles with the majority of the emotions. The two that it performs relatively well on, `Happy' and `Neutral', the visual model already has high accuracy for. Although `early fusion' will be investigated further in the next section, this characteristic suggests little benefit will be gained by combining the two modalities in this way. It may be more helpful to restore the weights from `Run Five', which has a higher F1-score (i.e. does slightly better across the different emotion categories), so this avenue will also be investigated
    
    \item \textbf{Sequence Length}: As stated above, adequate performance could not be reached for sequence length of 40, with 80 achieving higher accuracy and F1-score. A possible explanation being that because there is a large amount of noise in the raw-audio data, more data is required as an input to the network (i.e. longer sequences) before the underlying pattern can be learned, hence sequence length of 40 is just not long enough unless model complexity could be increased substantially to see through the noise better. Increasing the sequence length beyond 80 leads to memory issues during the `early fusion' stage when combined with the visual models
    
\end{itemize}

\subsubsection{Overall Conclusion}

`Underfitting' is evident during both training processes, however, it is difficult to increase model complexity significantly before `Overfitting' occurs due to the limited size of the AFEW dataset. I believe using other audio datasets for pre-training or data augmentation would help greatly improve performance in this section and should be used in future. For the video clips part, finding relevant datasets and / or augmetning the data is difficult because the need for alignment with the image frames. 

When comparing accuracy levels across the two audio models, the `Whole Video' approach outperforms the `Video Clip' approach (difference of c. 7.1\% in favour of the former). This was partly anticipated, hence why the output of the latter method is being combined with the visual model output. The reason for this gap in performance is possibly due to the purely statistical approach of openSMILE which suppresses the impact of the noise in the data, where as the GRU when learning finds it harder to filter out the noise without more data and thus is more heavily impacted.

\section{Fusion} \label{fusion_exp}

There are two forms of fusion discussed in section \ref{fusion_design}, the results for both approaches are included below:

\subsubsection{Early Fusion} 

\begin{figure}[htbp]

\centering
\subfigure[Option 1 Setup]{\label{fig:option_1}\includegraphics[width=60mm]{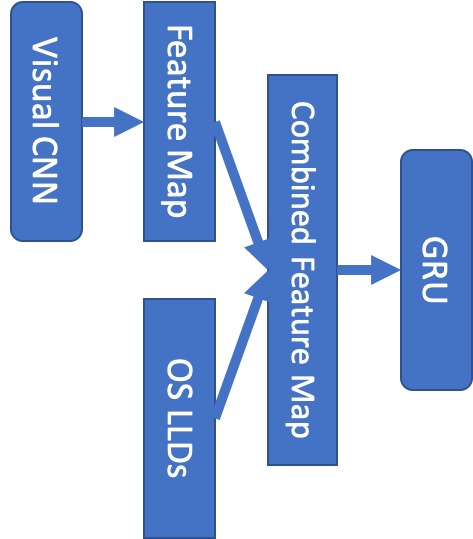}}\hspace{15mm}
\subfigure[Option 2 Setup]{\label{fig:option_2}\includegraphics[width=60mm]{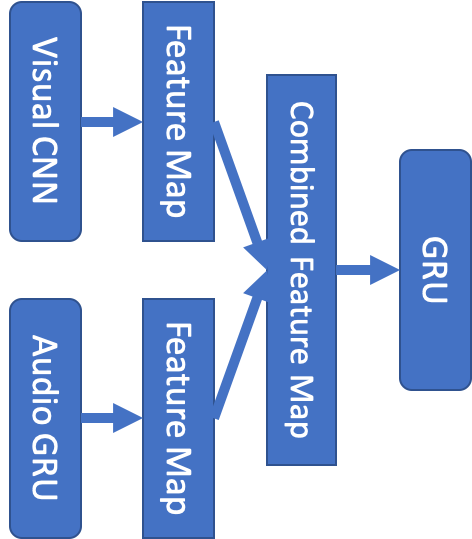}}
\subfigure[Option 3 Setup]{\label{fig:option_3}\includegraphics[width=60mm]{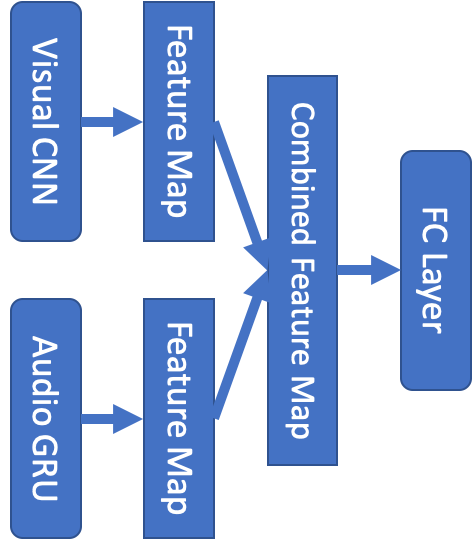}}\hspace{15mm}
\subfigure[Option 4 Setup]{\label{fig:option_4}\includegraphics[height=70mm,width=60mm]{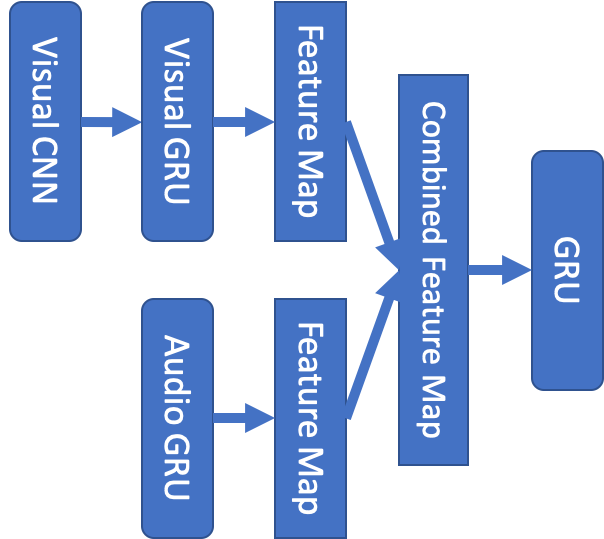}}

\caption{The 4 options presented for early fusion}
\label{fig:early_fusion_options}

\end{figure}

There are four approaches outlined in section \ref{fusion_design} for `early fusion' of the audio and visual modalities. To help visualise the different networks, these have been graphically represented in figure \ref{fig:early_fusion_options}. 

Additional parameter variations were limited to help with the comparison, but crucial to performance were:

\begin{itemize}
    
    \item \textbf{Feature Extractor}: The feature extractor model used for both audio and visual data workstreams to feed into the classifier. In the case of the audio data, the exact run is listed in the tables below, whereas for the visual models the CNN runs restored were those used in the temporal section \ref{temporal_exp} to aid contrast 
    
    \item \textbf{Sequence Length}: For options 2 to 4 in figure \ref{fig:early_fusion_options}, the sequence length for both audio GRU and the final classifying GRU has to be consistent. As mentioned in the The `Video Clip' part of the `Audio' section above, the performance for length 40 was poor, hence a sequence length of 80 has been used throughout for all results below
    
\end{itemize}

\begin{center}
\begin{longtable}{ |p{1.5cm}||p{6cm}|p{3cm}|p{2cm}|}
 \hline
 \rowcolor{lightgray} \multicolumn{4}{|c|}{Table 1: VGG-Face} \\
 \hline \hline
 Run & Description & Metric & AFEW \\
 \hline \hline
 
 One & Option 1 & Accuracy & 15.1\% \\
 & openSMILE features & F1-Score & 6.2\% \\
 & LR=0.0001, reduce & & \\
 & Batch size = 2 & & \\
 & Train epochs = 20 & & \\
 \hline \hline
 
 Two & Option 2 & Accuracy & 38.1\% \\
 & Audio `Run Two' & F1-Score & 36.3\% \\
 & LR=0.0001, reduce & & \\
 & Batch size = 2 & & \\
 & Train epochs = 20 & & \\
 \hline \hline
 
 Three & Option 3 & Accuracy & 35.8\% \\
 & Audio `Run Two' & F1-Score & 30.7\% \\
 & LR=0.0001, reduce & & \\
 & Batch size = 2 & & \\
 & Train epochs = 20 & & \\
 \hline \hline
 
 Four & Option 4 & Accuracy & 28.5\% \\
 & Audio `Run Two' & F1-Score & 24.3\% \\
 & LR=0.0001, reduce & & \\
 & Batch size = 2 & & \\
 & Train epochs = 10, 20 & & \\
 \hline \hline
 
 \rowcolor{yellow}
 Five & Option 2 & Accuracy & 41.8\% \\
 \rowcolor{yellow}
 & Audio `Run Five' & F1-Score & 39.4\% \\
 \rowcolor{yellow}
 & LR=0.00001, reduce & & \\
 \rowcolor{yellow}
 & Batch size = 4 & & \\
 \rowcolor{yellow}
 & Train epochs = 20 & & \\
 \hline \hline

 Six & Option 2 & Accuracy & 35.8\% \\
 & Audio `Run Five' & F1-Score & 32.7\% \\
 & LR=0.00001, reduce & & \\
 & Batch size = 4 & & \\
 & Train epochs = 10, 20 & & \\
 \hline \hline
 
 \rowcolor{yellow}
 Seven & Option 2 & Accuracy & 44.6\% \\
 \rowcolor{yellow}
 & Audio `Run Six' & F1-Score & 38.9\% \\
 \rowcolor{yellow}
 & LR=0.00001, reduce & & \\
 \rowcolor{yellow}
 & Batch size = 4 & & \\
 \rowcolor{yellow}
 & Train epochs = 20 & & \\
 \hline
 
\end{longtable}
\end{center}

The main inferences to be made from the above results are:

\begin{itemize}
    
    \item Option 2 performs the best of the initial model runs. Possible explanations for why the setup in `Run Two' achieves a higher accuracy are:
    \begin{itemize}
        
        \item Compared to option 1 the audio GRU in option 2 is able to improve the quality of the audio feature maps by including information about how each clips relates to those surrounding it
        
        \item Compared to option 3 the ability of the GRU classifier in option 2 to capture the temporal relationship of the combined per frame feature maps is an advantage over the fully connected layer
        
        \item Compared to option 4 the direct use of the visual CNN feature map in option 2 performs better. The reduction in the visual feature map, e.g. from 4096 dimension for VGG-Face network to 128 dimensions from the visual GRU, seems to lose too much information and outweighs the benefit of more evenly matching the dimensions of the audio and visual feature maps (partly because the audio model is a weaker predictor than the visual network)
        
    \end{itemize}
    
    \item Training in stages (i.e. classifier and then the whole network) applied in `Run Four' has a negative impact, one potential reason being that when propagating back the loss in the second stage, the network doesn't know how to correctly update the separate parts (i.e. should be weighted according to positive contribution) and therefore the convolutional layers degrade in quality  
    
    \item Restoring the weights from the audio model with a higher F1-score translates to a higher F1-score in the `early fusion' stage (i.e. `Run Five'), with the same trend appearing for accuracy (i.e. `Run Seven'). This seems to suggest, based on the conclusions drawn in the `Audio' section of this chapter, that the audio features carry forwards their `Overfitting' or `Underfitting' characteristics when re-used as an input to a new classifier

\end{itemize}

\begin{center}
\begin{tabular}{ |p{1.5cm}||p{6cm}|p{3cm}|p{2cm}|}
 \hline
 \rowcolor{lightgray} \multicolumn{4}{|c|}{Table 2: ResNet-50} \\
 \hline \hline
 Run & Description & Metric & AFEW \\
 \hline \hline
 
 One & Option 2 & Accuracy & 37.9\% \\
 & Audio `Run Five' & F1-Score & 32.2\% \\
 & global pool = True & & \\
 & LR=0.00001, reduce & & \\
 & Batch Size = 4 & & \\
 & Train = 20 epochs & & \\
 \hline \hline
 
 Two & Option 2 & Accuracy & 37.6\% \\
 & Audio `Run Six' & F1-Score & 31.8\% \\
 & global pool = True & & \\
 & LR=0.00001, reduce & & \\
 & Batch size = 4 & & \\
 & Train = 20 epochs & & \\
 
 \hline
 
\end{tabular}
\end{center}

\begin{center}
\begin{tabular}{ |p{1.5cm}||p{6cm}|p{3cm}|p{2cm}|}
 \hline
 \rowcolor{lightgray} \multicolumn{4}{|c|}{Table 3: SE-ResNet-50} \\
 \hline \hline
 Run & Description & Metric & AFEW \\
 \hline \hline
 
 One & Option 2 & Accuracy & 42.8\% \\
 & Audio `Run Five' & F1-Score & 37.9\% \\
 & global pool = False & & \\
 & LR=0.00001, reduce & & \\
 & Batch Size = 4 & & \\
 & Train = 20 epochs & & \\
 \hline \hline
 
 Two & Option 2 & Accuracy & 42.8\% \\
 & Audio `Run Six' & F1-Score & 38.9\% \\
 & global pool = False & & \\
 & LR=0.00001, reduce & & \\
 & Batch size = 4 & & \\
 & Train = 20 epochs & & \\

 \hline
 
\end{tabular}
\end{center}

Note ** indicates that due to exhausted memory, smaller batches had to be applied, therefore batch size of 2 for table 4

\begin{center}
\begin{tabular}{ |p{1.5cm}||p{6cm}|p{3cm}|p{2cm}|}
 \hline
 \rowcolor{lightgray} \multicolumn{4}{|c|}{Table 4: SE-ResNeXt-50**} \\
 \hline \hline
 Run & Description & Metric & AFEW \\
 \hline \hline
 
 One & Option 2 & Accuracy & 36.0\% \\
 & Audio `Run Five' & F1-Score & 32.3\% \\
 & global pool = True & & \\
 & LR=0.00001, reduce & & \\
 & Batch Size = 2 & & \\
 & Train = 20 epochs & & \\
 \hline \hline
 
 Two & Option 2 & Accuracy & 37.3\% \\
 & Audio `Run Six' & F1-Score & 34.0\% \\
 & global pool = True & & \\
 & LR=0.00001, reduce & & \\
 & Batch size = 2 & & \\
 & Train = 20 epochs & & \\

 \hline
 
\end{tabular}
\end{center}

Comparing performance of the above models to the best purely visual CNN + GRU networks in section \ref{temporal_exp}, all perform slightly worse, with slight drops in accuracy for VGG-Face and SE-ResNet-50 models, but significant decreases for ResNet-50 and SE-ResNeXt-50. The impact will be a function of the strength of the particular CNN feature extractor, which will largely drive prediction, and how well it complements the audio feature map (e.g. pick up different characteristics and represents them in a coherent manner). 

Unlike the impact seen in table 1, the difference between using Audio `Run Five' and `Run Six' is varied and smaller in magnitude.

\begin{figure}[htbp]
    \centering
    \includegraphics[width=0.8\textwidth]{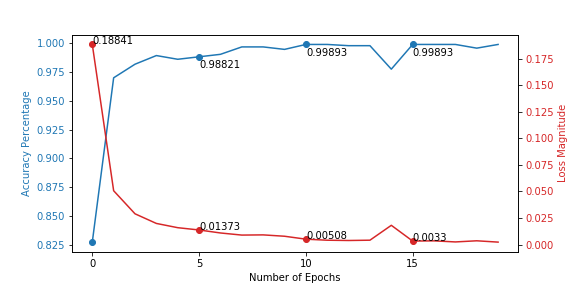}
    \caption{Early Fusion Option 1 Training Accuracy and Loss results for `Run Four'}
    \label{fig:stage_5_1_training}
    
\end{figure}

\begin{figure}[htbp]
    \centering
    \includegraphics[width=0.8\textwidth]{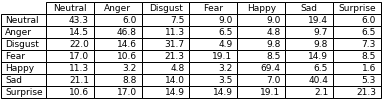}
    \caption{Early Fusion Option 1 confusion matrix for `Run Four'}
    \label{fig:stage_5_1_cm}
    
\end{figure}

The training accuracy and loss levels follow a similar path to those seen in the temporal section \ref{temporal_exp}, however, path has slightly higher variance which will be due to the noise in the audio data.

Analysing the confusion matrix in figure \ref{fig:stage_5_1_cm} compared to that in \ref{fig:stage_3_cm} for the SE-ResNet-50 + GRU network, we see that accuracy is more evenly matched across the emotion categories. The previously worst performing classes (e.g. `Fear', `Disgust' and `Surprise') have experienced a slight boost, but the stronger classes (e.g. `Happy' and `Anger') have seen drops. 

Given the results above, the following conclusions can be drawn:

\begin{itemize}
    
    \item \textbf{Performance Comparison}: As stated above, performance for the early fused VGG-Face and SE-ResNet-50 networks falls slightly, but drops more significantly for the other two models, when compared to the temporal results in section \ref{temporal_exp}. It may be that because different behaviours are exhibited (seen in the contrasting confusion matrices) that there is still a benefit to including these first two models in the `late fusion' stage, this will be further explored in the next section
    
    \item \textbf{Improvements}: In theory, combining the audio and visual workstreams should have a positive impact because it gives the final classifier more information to work with / has a slight regularisation effect on the visual model. However, if the audio feature map is not strong / robust enough, then it can bring down performance overall. Hence I believe more time should be invested in improving the audio feature extraction method in the `Video Clip' part of section \ref{audio_exp} to then have the desired effect during `early fusion'
    
\end{itemize}

\subsubsection{Late Fusion}

In section \ref{fusion_design}, possible techniques for fusing the predictions from different models were outlined. Also included was an explanation of the two key criteria for choosing models (i) high accuracy and (ii) model independence.

The 5 fusion methods applied at each stage below are (number will be used to reference the best approach in the following results tables):

\begin{enumerate}
    
    \item Class predictions weighted by model accuracy
    
    \item Class logits weighted by model accuracy
    
    \item Majority voting for logits (i.e. equal model weighting)
    
    \item Class logits weighted by model accuracy and the square root of the class weights \cite{liu}
    
    \item Linear regression to learn model weights and class weights based on 5 fold cross validation
    
\end{enumerate}

In the tables below, where visual models are included, results from only 5 of the networks are used unless otherwise stated. Given consistently lower results at each stage, the DenseNet-121 predictions have been discarded. 

\begin{center}
\begin{tabular}{ |p{1.5cm}||p{5cm}|p{2.5cm}|p{2cm}|p{2cm}|}
 \hline
 \rowcolor{lightgray} \multicolumn{5}{|c|}{Table 1: Fine-Tune Visual CNN Models} \\
 \hline \hline
 Run & Description & Fus. Method & Accuracy & Avg. Acc\\
 \hline \hline
 
 One & Optimal run per CNN & Four & 50.1\% & 42.7\% \\
 \hline \hline
 Two & 2 diff runs per CNN & Four & 50.4\% & 41.6\% \\
 \hline \hline
 Three & 2 additional VGG runs & One & 50.7\% & 42.6\% \\

 \hline
 
\end{tabular}
\end{center}

As would be generally expected with ensemble methods, the more CNN models included in the `late fusion' process, the higher the accuracy achieved. `Run Three' has two additional VGG-Face runs included (median and sequence length of 60) that performed well on the task and had slightly different setups to increase independence. The boost to accuracy is fairly significant, with the average of the models being 42.6\%, when combining the predictions an accuracy of 50.7\% is achievable.

\begin{center}
\begin{tabular}{ |p{1.5cm}||p{5cm}|p{2.5cm}|p{2cm}|p{2cm}|}
 \hline
 \rowcolor{lightgray} \multicolumn{5}{|c|}{Table 2: Fine-Tune Visual Temporal Models} \\
 \hline \hline
 Run & Description & Fus. Method & Accuracy & Avg. Acc\\
 \hline \hline
 
 One & Optimal GRU run per CNN  & One & 47.9\% & 42.8\% \\
 \hline \hline
 Two & BGRU runs & One & 48.3\% & 44.1\% \\
 \hline \hline
 Three & GRU + BGRU runs & One & 48.3\% & 43.3\% \\

 \hline
 
\end{tabular}
\end{center}

Slightly different behaviour is observed in table 2, in `Run Three' increasing the number of models used does not improve the accuracy over `Run Two' that just included the BGRU models. Two contributing factors might be the GRU models bring down the average model accuracy for `Run Three' and I expect the GRU and BGRU networks are not as independent as the different CNNs. The latter reason also possibly explains why the accuracy of `Run Three' in table 2 is lower than that of `Run Three' in table 1, despite a higher average across all models involved.

\begin{center}
\begin{tabular}{ |p{1.5cm}||p{5cm}|p{2.5cm}|p{2cm}|p{2cm}|}
 \hline
 \rowcolor{lightgray} \multicolumn{5}{|c|}{Table 3: Early Fusion Models} \\
 \hline \hline
 Run & Description & Fus. Method & Accuracy & Avg. Acc\\
 \hline \hline
 
 One & Audio `Run Five' & Two & 47.8\% & 39.6\% \\
 \hline \hline
 Two & Audio `Run Six' & Two & 47.5\% & 40.6\% \\
 \hline \hline
 Three & Combined above runs & Two & 48.3\% & 40.1\% \\
 
 \hline
 
\end{tabular}
\end{center}

Although the accuracy of all chosen `early fusion' models after ensembling is the same as that of the equivalent run in table 2, the average accuracy across the models in table 3 is lower. This is consistent with the analysis and hypotheses presented in the `early fusion' section above. It suggests that the inclusion of audio features introduces greater intra-model independence, with the relationship between visual and audio outputs providing another differentiating factor.

\begin{center}
\begin{tabular}{ |p{1.5cm}||p{5cm}|p{2.5cm}|p{2cm}|p{2cm}|}
 \hline
 \rowcolor{lightgray} \multicolumn{5}{|c|}{Table 4: All Models} \\
 \hline \hline
 Run & Description & Fus. Method & Accuracy & Avg. Acc \\
 \hline \hline
 
 One & All visual runs & Two & 50.6\% & 42.7\% \\
 & (inc. early fusion) & & & \\
 \hline \hline
 
 Two & Inc. `Whole Video' audio & One & 48.6\% & 40.8\% \\
 \hline \hline
 
 Three & Advance models only & One & 47.9\% & 41.9\% \\
 & (One, Two \& Two above) & & & \\
 \hline \hline
 
 Four & Simplistic Models & Three & 52.7\% & 42.4\% \\
 & (One \& One above) & & & \\
 \hline \hline
 
 Five & Highest Average Models & Three & 51.2\% & 43.9\% \\
 \hline \hline
 
 Six & Excluding `early fusion' & Two & 52.5\% & 43.6\% \\
 \hline \hline
 
 Seven & Trial-and-Error & Two & 53.0\% & 42.4\% \\
 & (inc. early fusion) & & & \\
 \hline \hline

 \hline
 
\end{tabular}
\end{center}

The `late fusion' of all the stages is a very volatile process and therefore hard to predict (especially with a large number of models and different fusion methods available). For example, swapping the SE-ResNet-50 + GRU for the equivalent BGRU network leads to a c. 1\% drop in accuracy, despite there being a difference of 2\% between these two models in the other direction. 

My approach was to try different combinations to understand what generally worked best, before then trialling different minor alterations (non-exhaustive) until a likely near optimal run could be found. 

To begin with I wanted to investigate the impact of including the `Whole Video' audio, the paper \cite{occam} references that ``audio can bring a +3\% accuracy gain when combined with the visual modality" in the EmotiW challenge. However, I only found a drop occurred (see `Run One' and `Run Two') which is similar to the findings in \cite{lu}. Therefore the `Whole Video' method was removed for all future runs. Note the audio is still being utilised through the `early fusion' models.

The next step was to explore permutations of the visual models, with `Run Three' and `Run Four' suggesting that including simpler models (e.g. GRU rather than BGRU) gave a higher accuracy. One possible explanation being that models that might be subject to `Overfitting' are not well suited to ensemble methods, which is an obvious effect if all networks included exhibit similar behaviours (e.g. all predict `Happy' correctly, but poor performance for other classes, then this pattern is just reinforced). Also, I found that including too many models from one particular training stage lowered the fusion accuracy, this will be due to the over weighting of a model type and decrease in independence. Hence, striking the right balance is an important factor.

The `early fusion' approach does have some benefit, which can be seen when comparing `Run Four' and `Run Seven' with `Run Six'. The reasons for this minor boost are likely to be similar to those given directly after table 3. 

The best performing combination is `Run Seven', which consists of early / varied CNN runs (i.e. `Run Seven' and `Run Twelve' in table 1 and `Run One' of table 2 - 6 in section \ref{fine_tune_exp}), predominantly GRU networks and `early fusion' models utilising `Run Five' audio features. 

Given the results above, the following conclusions can be drawn:

\begin{itemize}
    
    \item \textbf{Ensemble Impact}: The increase in performance at each training stage between a single model and the fused predictions is about 5-9\%. This shows the possible benefit of the `brute force approach' (i.e. training lots of models and then aggregating) over creating one powerful predictor. Although, the former approach will see marginal returns over time (i.e. each new model added has minimal effect) and hence smarter networks are still needed
    
    \item \textbf{Fusion Method}: The optimal method seems to vary according to the stage. For example, the best approach in table 2 is `One', where as in table 3 it is `Two'. A likely explanation for this is the noise in the models and the relative scales of their logits. A straight prediction will just pick a category based on the largest activation of the model output, but it might be that two logits are very similar in magnitude because the model is unsure. Therefore the logits may hold more information, but be far noisier when fusing (especially if the logtis are not normalised). The `early fusion' models (i.e. table 3) all have the same architecture (i.e. Option 2), therefore the logits which are possibly better proportioned across the classes may have more meaning when combined, unlike two different RNN networks (i.e. GRU and BGRU in table 2) where smoothing the outputs is a benefit
    
    \item \textbf{Linear Regression}: This fusion method does not feature in the above tables because it consistently performed the worst. Given the size of the AFEW dataset, it seems that including a final layer with additional weights is hard to train. This is similar to the findings of the other entrants to the 2018 EmotiW FER AV competition, where one of the above fusion methods was employed or a grid search approach used
    
    \item \textbf{Scaling Logits}: In addition to the 5 fusion methods listed at the start of this section, I also tried re-scaling the logits to the range 0 to 1 for methods 2, 3 and 4. However, the late fusion accuracy levels were lower than the optimal approach reported in each case. Implying that the logits and their true relative magnitudes does contain useful information as discussed above
    
    \item \textbf{Best Result}: `Run Seven' achieves the highest accuracy across all combinations tried and was largely found through a general `trial-and-improvement' process. The fusion method for this run was `Two', which reflects the findings of the paper \cite{liu} (winner of the 2018 EmotiW competition), which states ``since emotion is a complex subject, fused all the different predictions by merely giving weights for the possibilities of each method". The set-up for this run will be replicated and used for final submission to the EmotiW organisers
    
\end{itemize}

%%%%%%%%%%%%%%%%%%%%%%%%%%%%%%%%%%%%
\chapter{Evaluation} \label{evaluation}

The previous chapter details the training and validation results, making inferences about what worked and what could be improved, before carrying that information forward to the next stage. Finally, in the last section we fused the predictions of certain models to boost performance to give final accuracy levels for each stage and the whole FER task. 

To help give some context to the results, I have for comparison included the published findings of other entrants to the EmotiW FER AV competition from last year. It is difficult in some cases to ascertain exactly which models they have included in a reported result, but I have tried to include the most relevant run to this paper (i.e. similar networks deployed) and / or their best outcome. I have added some commentary where possible to explain differences, but further information on the different entrant's approaches can be found in section \ref{other_papers}.

TP indicates results for `This Paper', which are taken directly from the previous chapter and hence a better explanation of the exact setup of the run can be found there. The results of the other entrants are reported in descending order (i.e. winner at the top to lowest in the bottom row for the overall challenge).

\begin{center}
\begin{tabular}{ |p{1.5cm}||p{2cm}|p{6.5cm}|p{1.5cm}|p{1.5cm}|}
 \hline
 \rowcolor{lightgray} \multicolumn{5}{|c|}{Table 1: CNN Models only} \\
 \hline \hline
 Entrant & Run & Description & Val. Acc & Test Acc\\
 \hline \hline
 
 TP & Final Run & & 50.7\% & \\
 \hline \hline

 \cite{liu} & Base Run & 4 CNNs with Landmark Detection & 55.1\% & 55.4\% \\
 & Opt Run & Above run with Class Weights & 49.6\% & 59.1\%\\
 \hline \hline

 \cite{fan} & Base Run & 2xDSN-CNN + 2xCNNs & 55.1\% & 58.0\% \\
 & Opt Run & 2xDSN-CNN + 2xCNNs & 56.1\% & 59.6\%\\
 \hline \hline

 \cite{lu} & Base Run & ResNet-50 (augmentation) & 48.6\% &  \\
 & Opt Run & VGG-Face (augmentation) & 50.3\% & \\
 \hline \hline
 
 \cite{occam} & Base Run & CNN with feature aggregation & 49.7\% & \\
 & Opt Run & Above run with Class Weights & 62.7\% & 60.6\% \\

 \hline
 
\end{tabular}
\end{center}

\begin{center}
\begin{tabular}{ |p{1.5cm}||p{2cm}|p{6.5cm}|p{1.5cm}|p{1.5cm}|}
 \hline
 \rowcolor{lightgray} \multicolumn{5}{|c|}{Table 2: CNN + RNN Models only} \\
 \hline \hline
 Entrant & Run & Description & Val. Acc & Test Acc\\
 \hline \hline
 
 TP & Final Run & CNNs + BGRU & 48.3\% &  \\
 \hline \hline
 
 \cite{lu} & Opt Run & VGG / ResNet-50 + BLSTM & 48.2\% & 49.5\%\\

 \hline \hline
 \cite{occam} & Base Run & VGG-LSTM & 49.4\% & \\
 & Opt Run & Above run w/ weighting fusion & 58.2\% & \\

 \hline
 
\end{tabular}
\end{center}

\begin{center}
\begin{tabular}{ |p{1.5cm}||p{2cm}|p{6.5cm}|p{1.5cm}|p{1.5cm}|}
 \hline
 \rowcolor{lightgray} \multicolumn{5}{|c|}{Table 3: Audio Models only} \\
 \hline \hline
 Entrant & Run & Description & Val. Acc & Test Acc\\
 \hline \hline
 
 TP & Final Run & openSMILE `$emo\_large$' & 27.2\% &  \\
 \hline \hline
 
 \cite{liu} & Base Run & openSMILE w/ PCA & 31.0\% & \\
 \hline \hline
 
 \cite{occam} & Base Run & openSMILE w/ pre-training & 33.5\% & \\

 \hline
 
\end{tabular}
\end{center}

\begin{center}
\begin{tabular}{ |p{1.5cm}||p{2cm}|p{6.5cm}|p{1.5cm}|p{1.5cm}|}
 \hline
 \rowcolor{lightgray} \multicolumn{5}{|c|}{Table 4: Late Fusion} \\
 \hline \hline
 Entrant & Run & Description & Val. Acc & Test Acc\\
 \hline \hline
 
 EmotiW & Baseline &  & 38.8\% & 40.5\% \\
 \hline \hline
 
 TP & Final Run &  & 53.0\% & 48.5\% \\
 \hline \hline
 
 \cite{liu} & Opt Run & CNNs, Landmark, LSTM, OS audio  & 50.7\% & 61.9\% \\

 \hline \hline
 \cite{fan} & Opt Run & Fusion matrix of DSN + CNNs & 57.4\% & 61.1\% \\

 \hline \hline
 \cite{lu} & Opt Run & 2x CNN+RNN, C3D, Audio Spectr. & 56.1\% & 60.6\% \\
 
 \hline \hline
 \cite{occam} & Opt Run & Fusion of all visual + audio & 62.7\% & 60.6\% \\

 \hline
 
\end{tabular}
\end{center}

Given the results above, the following conclusions can be drawn:

\begin{itemize}
    
    \item \textbf{Performance Across Stages}: Comparing validation accuracy levels, it seems my approach is slightly below other entrants in tables 1 and 3, but fairly similar in table 2. However, with further fine-tuning (particularly applying data augmentation, which would help boost performance of the larger models in this paper) I believe the gaps could be reduced. After `late fusion' my validation results continue to be slightly lower in general, although they are higher than the winner from last year's competition. This is likely due to weaker performance in the earlier stages being carried forward rather than a reflection on the `late fusion' approach
    
    \item \textbf{Main Driver}: Based on the accuracy results reported in tables 1 - 3, it seems the solely visual models provide the majority of the predictive power. The temporal and audio stages may boost performance slightly, but CNNs are the core part of the overall model. This conclusion is consistent with the statements made in Chapter \ref{experimentation_opt}
    
    \item \textbf{Trend in Test Results}: Analysing the differences between validation and test accuracy in table 4, there is large jump upwards for 3 out of the 4 entrants, with the other experiencing a slight drop. In particular, \cite{liu} the winner last year experienced an increase of 11.2\%, which suggests that is difficult to understand / predict what will perform best on the test data given the validation results
    
    \item \textbf{Final Test Result}: The final results for this paper are included in the `TP' row of table 4. The model comfortably exceeds the baseline results set by the EmotiW competition organisers. However, my model suffers a significant drop in accuracy from the validation to testing phase (53.0\% to 48.5\%), going against the trend discussed in the bullet point above that would suggest the test accuracy would likely fall somewhere between 55\% and 60\%. Due to time constraints, only one test submission was made for this project, whereas other teams were able to submit predictions up to 7 times and report their best outcome. To illustrate the point of variability, \cite{liu} reports one run that suffers a decrease in accuracy from the validation to test dataset of c. 3\%, but a minor amendment (slightly different weights) to the exact same set of models leads to a lower validation (49.6\%), but much higher test accuracy (59.1\%). Therefore, given more time I would look to submit a greater range of models for testing, which would hopefully give a better indication of the relative success of the method proposed in this paper
    
\end{itemize}

%%%%%%%%%%%%%%%%%%%%%%%%%%%%%%%%%%%%
\chapter{Conclusion} \label{conclusion}

As stated in the introduction, there were two main contributions hoping to be made in this paper. The application of `state-of-the-art' visual models to the FER task and the `early fusion' of the audio and visual feature maps.

\section{Positive Findings}

An idea that has performed well at every stage of this project is including the `squeeze-and-excitation' technique in a network. Although it has outperformed all models on average, the best indicator of success is the difference between the ResNet-50 and SE-ResNet-50 networks given they have the same underlying architecture (`squeeze-and-excitation' block being the only change). For example, comparing the best runs in sections \ref{fine_tune_exp} and \ref{temporal_exp}, SE-ResNet-50 has a higher accuracy by 3.2\% and 4.2\% respectively.

A possible explanation for this superior performance is the ``spatial encodings throughout its feature hierarchy" \cite{squeeze_excitation}, which means the network can ``selectively emphasise informative features and suppress less useful ones". This is an obvious benefit given the subtle nature of the features in the FER task. Therefore the activations in the convolutional layers for raised eyebrows when a person is exhibiting `Surprise' can be boosted, making it easier to identify this emotion.

The main two advantages of the `squeeze-and-excitation' technique, as discussed in section \ref{CNNs}, is that it does not significantly increase the number of model parameters (important for the AFEW dataset in particular) and can be easily incorporated into convolutional blocks. Therefore, it could have a positive impact on the leading networks deployed in the 2018 EmotiW FER AV competition. One particular case of interest would be applying the technique to VGG-Face that generally outperforms the ResNet-50 model, but this could solely be due to the pre-training of the former model on a large face dataset.

\section{Further Exploration Required}

\subsubsection{Early Fusion}

Although spectrograms are used for the audio modality in \cite{lu}, with the whole audio clip split into sub-parts. There is no alignment and therefore no `early fusion' with the visual model outputs. The other entrants classify the entire audio sample in a similar way to the `Whole Video' approach applied in this paper.

The idea of early fusing has been used in other competitions, but this was an opportunity to investigate its application in the EmotiW challenge.

openSMILE is flexible enough to provide a feature map for small audio clips and is known for its strong performance in several other audio only competitions. However, based on the relatively poor results in the second part of section \ref{audio_exp}, the fusion benefit was always going to be limited. It seems that for such small clips the configuration being used in this paper was unable fully handle the noise in the audio data.

I believe the approach could still provide some benefit, with additional techniques worth considering in any future research included in Chapter \ref{future_work}.

\subsubsection{Temporal Models}

Although the results for the temporal model stage were not strictly better than solely the CNNs, there was a gain when including the RNN runs in the `late fusion' process. Therefore, it is worth considering the type of RNN applied.

Of the versions trialled in this paper, the BGRU performed best (see section \ref{temporal_exp}) on average. The advantage being the network is able to use future and historic context to improve the predictive power. 

It is difficult to compare the BGRU model presented in this paper to the BLSTM networks used by last year's EmotiW entrants. However, Table 2 in Chapter \ref{evaluation} shows almost identical validation accuracy levels between the CNN + BGRU and CNN + BLSTM runs despite the CNN performance being much higher in the latter case. Therefore, it might be that the GRU RNN on this limited dataset has a minor advantage over the LSTM and is worth additional investigation.

\subsubsection{Attention}

The impact of introducing the attention mecahnism to the VGG-Face model was slightly negative (see section \ref{temporal_exp}). However, given its success in other machine learning fields it is an idea that could still play a big part for automated FER. One explanation for the drop in accuracy is the limited size of the AFEW dataset, which hinders the ability of the model to learn the additional attention parameters (see section \ref{attention}). Once alternative larger and well-labelled dynamic FER datasets emerge, it would be worth revisiting this approach.

\section{Less Promising Avenues}

In the reverse situation to the SE-ResNet-50 model, the NASNet and DenseNet-121 networks were consistently the worst two performers. Therefore, based on my findings and training pipeline they should be discarded in favour of the other 4 models. 

Possible explanations for this outcome as discussed in section \ref{temporal_exp} are:

\begin{itemize}
    
    \item \textbf{DenseNet-121}: The full connectivity of all layers in the network introduces additional parameters that perhaps cannot be efficiently learnt for the AFEW dataset given limited size
    
    \item \textbf{NASNet}: In each training stage I limited the number epochs, for timing reasons and to allow a fairer comparison to the other visual models. However, given the `trial-and-improvement' approach of NASNet, this may not have been enough time to exhaustively explore the search space to find the optimal permutation of convolutional cells (size of dataset might also be factor)
    
\end{itemize}

I had expected the DenseNet-121 model to perform better given it is used as one of the CNNs by the winner of the 2018 EmotiW FER AV competition. An alternate training approach (i.e. different parameters and pre-training datasets) may be the reason or their method could benefit from deploying other models such as SE-ResNet-50.

\section{Accomplishments}

I believe the main contribution of this paper is to investigate the impact of the latest cutting-edge visual machine learning models to the FER task, including their ability to be used as feature extractors for RNN models. Although further testing is required to confirm the true impact, there are certainly some promising signs regarding the `squeeze-and-excitation' technique. 

Also, this project has shed some light on the early fusion of the audio and visual feature maps and other RNN models (i.e. RNN with attention and Bi-directional GRU with other teams using the slightly older LSTM model). Although the results are not a jump forwards, certain insights can be inferred and the lessons learned can be put into practice in the future to realise the possible benefit of these methods.

\section{Project Critique}

Although the aim of this project was to apply a few novel techniques to the FER task, a way of gauging their impact was required. The EmotiW FER AV 2019 competition is a long-running challenge and is hotly contested every year. The model results for this paper are slightly behind the leading entrants from 2018, but this outcome may be slightly misleading.

To properly compete the most important requirement would be more time. This project was completed in roughly 5 months (including time to write this report), whereas other entrants will work as groups, have a longer period to work on the challenge or are able to build on their entries from the previous year. This allows them more time to try a greater number of model permutations, additional execution runs (e.g. re-initialise multiple times), further optimise parameters and increase pre-processing of the data (e.g. extensive data augmentation). All of these factors may help to gain a few extra percent, thus making the comparison to my approach difficult to truly understand.

An example being, in the paper \cite{occam} from one of last year's entrants, they aggregated the results from 50 runs of their visual model. Given the Imperial College London course timings and my aim to explore other methods, there was only time to run each model permutation once.

Too fully evaluate performance across the different submissions, standardising certain aspects of the competition may help (e.g. limiting the number of model runs, computational power, data augmentation applied, use of other datasets, etc.). This would increase the focus on techniques implemented rather than the level of brute force applied.

\section{Final Thoughts}

Given the results of this paper and those from previous years of the EmotiW FER AV challenge, it seems performance for automated FER on the AFEW dataset is saturating. Only minor improvements have been achieved when comparing the 2017 and 2018 winning teams.

The trend discussed in section \ref{deep_learning_basics} of machine learning models going deeper has clearly reached its limits for this challenge. Hence the desire in this project was to apply new visual models that are trying not to increase the number of parameters, but instead they are trying to be smarter. For example, boosting / suppressing certain feature maps based on relevance or finding new combinations better suited to the problem. Hopefully advances in the wider machine learning community will continue along these lines and therefore can be applied to automated FER. 

In general, the networks built for this challenge could not be practically applied in modern portable devices such as phone due to their bulky nature. The research in paper \cite{occam} seems to be the only entry building a lightweight model that could be put into production. 

However, the accuracy levels are far behind other visual tasks and probably too low still to be safely and efficiently put into practice. The concerns raised in Chapter \ref{ethics} about FER software needing to be precise if it is to be helpful and not harmful (putting surveillance aside) have not been reached. Therefore I still believe further work has to be done in this area, without great advancements in machine learning visual and audio models, collecting larger and more diverse datasets would be a good place to start.

%%%%%%%%%%%%%%%%%%%%%%%%%%%%%%%%%%%%
\chapter{Issues Encountered} \label{issues_encountered}

Throughout this project there were a number of challenges faced, which may in part have been due to my relative inexperience of building models on this scale (previous courseworks were less technically demanding and in fairly sanitised environments), but also due to the well-known complexity and difficulty of the task. Below I have highlighted the main problems I came up against and how these were resolved.

\section{Data and Feature Extraction} \label{data_problems}

\begin{figure}[htbp]

\centering
\subfigure[Disgust Image]{\label{fig:face_bad_1}\includegraphics[width=40mm]{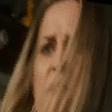}}
\subfigure[Fear Image]{\label{fig:face_bad_2}\includegraphics[width=40mm]{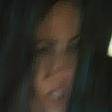}}
\subfigure[Sad Image]{\label{fig:face_bad_3}\includegraphics[width=40mm]{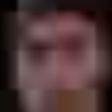}}

\caption{Selection of poorly captured facial images}
\label{fig:face_detect_bad}

\end{figure}

\begin{itemize}

    \item \textbf{Data Format}: There were initial problems with incorrect labels, wrong file formats and sizes. Each of these once identified have required time to be fixed / standardised
    
    \item \textbf{Missing Data}: Missing frames due to the face detection and normalisation software was mentioned in section \ref{data_process}. The same problem occurred for the datasets listed in section \ref{Static_Images}, although this would have been less impactful given the large number of images when these datasets are combined. The secondary impact of missing frames is in the training of the RNNs, for example, the data is fed in sequentially and so the GRU assumes each frame is the same time-step apart, however, it may be that the subsequent frame is actually X * 0.04 seconds later on in the video
    
    \item \textbf{Facial Alignment}: Figure \ref{fig:face_detect_bad} shows that despite the face detection and alignment software being cutting-edge, some frames are just very difficult to capture properly. In each of the three cases a human would struggle to correctly identify the emotion and shows how `in-the-wild' some of the videos are. An improvement in the software over time would subsequently better performance in FER

    \item \textbf{Image / Audio Variability}: Figure \ref{fig:face_detect_bad} is also a great example of `subject variability' and `environmenal variability' discussed at length in section \ref{discussion} along with possible solutions. The forms of variability and solutions apply to the audio data as well
    
    \item \textbf{AFEW Covariance Shift}: Although the `fine-tuning' stage helps to increase the specialisation of the models (see section \ref{design}), there is also covariance shift between the training, validation and test AFEW datasets. An example of this can been seen in the varying video sequence length distributions evident in figures \ref{fig:train_lens}, \ref{fig:valid_lens} and \ref{fig:test_lens}. The impact being that the model is trained on slightly longer sequences on average than it is evaluated and tested on

    \item \textbf{Frame Inconsistencies}: In section \ref{training_process}, I provided a hypothetical but very realistic frame-level breakdown of a video in the training, validation or test datasets. It demonstrated a key challenge for FER `in-the-wild' audio-visual sequences, the imperfect nature of the data:
    \begin{itemize}
    
        \item \textbf{Visual}: Issues include:
        \begin{itemize}
            \item The videos may only have a few frames actually displaying the correct emotion
            \item A range of different emotions are displayed
            \item There may be multiple subjects in the video, with only one of them displaying the labelled emotion
            \item Only subtle displays of the emotion (even if frequently shown)
        \end{itemize}
        
        \item \textbf{Audio}: Issues include:
        \begin{itemize}
            \item Background noise distorts feature extraction. Although this can be removed using python libraries, it
            can also filter out some of the desired audio
            \item The videos may only have a short section where the audio reflects the correct emotion
            \item Multiple sources of audio in the video
        \end{itemize}
        
    \end{itemize}
    
    \item \textbf{Model Variability}: In section \ref{experimentation_opt}, the process followed was to train a model, evaluate the performance on the validation dataset and then try to optimise it by making suitable changes. An issue with this approach is that even if the same model was run multiple times the results would vary significantly for this FER task. This could be due variations in the way the data was shuffled / batched, weights were intialised or the order the GPUs ran batches in. Some of these effects can be mitigated, but not entirely removed and were possibly magnified by the noise in the datasets. Given more time, I would carry out several runs for each model, before calculating the mean performance. Instead I was forced to make decisions based on single noisy runs, which may have led to some incorrect choices. This was further exacerbated by often having to extrapolate between different models runs (not considered best practice) 

\end{itemize}

The benefits of fusing the feature maps of audio and visual models were discussed in section \ref{fusion_design}, there can obviously be downsides. For example, the subject may be too shocked speak, hence displaying clear visual clues, but the audio model would suggest the frame is neutral. The final network that is fed the combined feature map may be able to learn some of this behaviour, but it won't always be clear given the small size of the AFEW dataset.

Some of the `Frame Inconsistencies' can be mitigated by our choices in the setup of the networks (see section \ref{training_process}). The hope is the model is sophisticated enough to pick-up on these subtle nuances in the data.

\section{Software and Tools} 

\begin{itemize}

    \item \textbf{openSMILE}:
    
    \begin{itemize}
        
        \item \textbf{Interaction}: The tool is written in `C++', can only be run from terminal and the output is an `.arff' file. Although these obstacles were all overcome, it meant extra time had to be spent to making it fit within the wider project pipeline
        
        \item \textbf{Configuration}: The standard openSMILE configuration files were not appropriate for the frame-wise audio clips, which then required careful amendments. The difficulty being choosing a suitable starting file, making the correct changes (e.g.  what to include and to output) and how to apply the LLDs. Also, some of the settings were hard-wired in the underlying software, so making an alteration didn't always have the intended impact
        
        \item \textbf{Instruction}: There is documentation provided for the software, but it doesn't cover my specific case. In addition there is limited guidance online and I struggled to find people within the department with experience of using openSMILE. Therefore the process of extracting meaningful features was a trial and improvement process, with limited time to test the outputs
        
    \end{itemize}
    
    \item \textbf{TensorFlow}:
    
    \begin{itemize}
    
        \item \textbf{Data Workstreams}: In building my original training script, I chose to follow the best practice advice provided by TensorFlow and use their `$tf.data.Dataset$' API, which performed efficiently in the `pre-training visual' stage. However, when trying to combine the audio (`.csv' format) and visual (`.jpg' format) data forms, the data was no longer being processed / errors were occurring. Part of the problem being the data had to be transformed into sequences first, before then being decoded. However, the TensorFlow APIs are applied directly to the `$tf.data.Dataset$' output (e.g. like the shuffle or batch methods). After a number of attempted fixes (e.g. saving the data in numpy format, reading the data in before initialising the dataloader and processing the location strings separately, which all failed for memory reasons / not compatible with other TensorFlow tools), I finally managed to process the data correctly by using the soon to be `deprecated' function `$tf.train.slice\_input\_producer$' (along with the accompanying queuing protocol in TensorFlow required for this method)
        
        \item \textbf{Versions}: A problem I consistently had with the TensorFlow software (including the above example) is handling the number of different versions available. The result being some functions are deprecated or have only been included in the latest version 1.14.0. Also, this makes it difficult to decide a best approach when the online documentation and tutorials disagree in method, with bug-fixing support similarly confused. This issue is magnified for particularly large scale / complex projects such as this, where some of the techniques used are not commonly deployed
        
    \end{itemize}
    
\end{itemize}

\section{Model Execution}

\begin{itemize}
    
    \item \textbf{Memory}: Given the size of the static image datasets, initially the models were crashing with every attempt. To solve this, I had to limit runs to the top-spec GPU clusters within the department and research extensively the most efficient ways of handling / feeding data in TensorFlow
    
    \item \textbf{Computational Power}: The standard GPUs within the department are not sufficient to run the models (i.e. model crashes), hence only `GeForce GTX TITAN X' GPUs can be used of which there are limited number. This means that typically only a couple of models can be run at the same time and in some cases changes to the model set-up had to be made to reduce the computational burden
    
    \item \textbf{GPU Clusters}: The process for running models on GPUs is done through the departmental online GPU clusters. Although powerful, they are fiddly to work with and difficult to monitor interim results / progress, making the process of training / analysis slow going. This was partly solved by running smaller programs in `Google colab'

    \item \textbf{Multiple Scripts}: If numerous runs were required for the same training stage (e.g. fine-tuning), the set-up of the GPU clusters meant that the same script could not be submitted repeatedly with minor changes (e.g. different CNN models). No error was outputted by the GPU clusters, so it took a while working with CSG to identify the problem. The solution was to create multiple scripts to run concurrently

\end{itemize}

\section{General}

\begin{itemize}

    \item \textbf{Imperial Servers and Labs}: There was a migration of servers from the Imperial DoC Huxley building to Slough. This included both `vol/phoebe' and `vol/gpudata' (see section \ref{storage}) being transferred without warning, which meant 2 weeks of run time were lost
    
    \item \textbf{FileZilla and VPN}: Access to the Imperial servers through FileZilla (for files) and the VPN (to ssh into lab computers) can be very temperamental, which slowed progress when working remotely
    
\end{itemize}

%%%%%%%%%%%%%%%%%%%%%%%%%%%%%%%%%%%%
\chapter{Future Work} \label{future_work}

There are some completely new ideas that could still be applied to the FER area and certain techniques deployed by past entrants to the EmotiW competition that may improve the results in this paper. Given more time these would be interesting to explore and a select few have been summarised below:

\subsubsection{Direct Changes to Existing Approach}

\begin{itemize}

    \item \textbf{Further Fine-Tuning}: Common machine learning techniques to improve model accuracy are:
    
    \begin{itemize}
        
        \item Re-initialise each of the models several times, then apply ensemble methods to improve the predictive power
        
        \item Fine-tune each visual model, rather than extrapolating from the findings of the VGG-Face runs
        
        \item Following the late fusion process and the optimal set-up / parameters being decided, re-train all models included on the combined training and validation datasets to help boost test results 
        
    \end{itemize}
    
    \item \textbf{Parameter Search}: Increase the search space for optimising parameters and hyperparameters. Other techniques, such as random grid search, Bayesian Optimisation and Evolutionary Algorithms could also be utilised 
    
    \item \textbf{Data}: Collecting new data like STED \cite{liu} or data augmentation \cite{augmentation} \cite{synthesis} for both audio and visual data (exploited by all 2018 EmotiW leading entries) would improve the robustness of the networks in this paper, it would also permit us to increase model complexity without suffering from `Overfitting' and thus capturing more intricate mappings 
    
    \item \textbf{Audio}: Given the relatively weak performance of the audio workstream in this paper, there are a number of other techniques to employ:
    
    \begin{itemize}
        
        \item Denoise the audio data first before applying feature extractors
        
        \item Trial other openSMILE configurations on this dataset
        
        \item Overlap the audio clips slightly to increase information being analysed by openSMILE to improve robustness 
        
        \item Create spectrograms for the audio data that align per frame with the current openSMILE and visual models, hence another feature map could be included in the `early fusion' process
        
    \end{itemize}
    
    \item \textbf{Loss}: Other loss functions, such as, `Island Loss' or `Locality-Preserving Loss' discussed in section \ref{loss_func} could be explored to improve training performance and address imbalance in the datasets

    \item \textbf{Attention Mechanism}: To better the connection between the audio and visual signals, the very powerful technique of attention could be applied first across the combined sequence. Rather than just concatenating the feature maps in the `early fusion' stage, the approach would allow the two modes to better support each other (i.e. smiling whilst saying something positive would further boost the `Happy' response at this point). 
    
    \item \textbf{Performance Analysis}: Deep dive results to find out what the main drivers are and try to fix / accentuate these features. For example, does the mouth shape whilst talking distort results and can this be mitigated
    
    \item \textbf{Squeeze-and-Excitation}: As stated in the conclusion chapter of this paper, applying this technique to other visual models, such as VGG-Face, could improve accuracy levels given it's evident impact on the ResNet-50 network

    \item \textbf{Other Models}: Try to incorporate other facial challenges for multi-task learning, such as Action-Units and Valence-Arousal models. Improves the generalisability of the model and provides access to more data
    
\end{itemize}

\subsubsection{New Additions}

\begin{itemize}
    
    \item \textbf{Multiple Facial Attributes \cite{deep_fer_survey}}: Increasing the number of facial attributes adds information to the model (important for small datasets). There are numerous types to be considered, for example, geometry, texture, curvature and normal components (x,y,z)

    \item \textbf{Scaled Models}: The DSN model showed good results, partly due to the multi-scale approach, with the side output technique being applicable to visual models in this paper. Also, the U-Net model uses down and up sampling to deal with scale, which helps to localise features. A similar model could be applied for FER
    
    \item \textbf{Spatial Transformations}: Although the faces are aligned and normalised, each face being analysed is still very different (e.g. distance between facial features varies greatly). Hence before the feature extraction occurs, the aim would be to standardise the faces through a transformation to a base to further to improve CNN performance 
    
    \item \textbf{Capsule Networks}: Recently published model has been proven to handle spatial problems well

\end{itemize}

%%%%%%%%%%%%%%%%%%%%%%%%%%%%%%%%%%%%
\appendix

\chapter{openSMILE}

The ``following (audio specific) low-level descriptors can be computed by openSMILE:

\begin{itemize}
    \item Frame Energy
    \item Frame Intensity / Loudness (approximation)
    \item Critical Band spectra (Mel/Bark/Octave, triangular masking filters)
    \item Mel-/Bark-Frequency-Cepstral Coefficients (MFCC)
    \item Auditory Spectra
    \item Loudness approximated from auditory spectra
    \item Perceptual Linear Predictive (PLP) Coefficients
    \item Perceptual Linear Predictive Cepstral Coefficients (PLP-CC)
    \item Linear Predictive Coefficients (LPC)
    \item Line Spectral Pairs (LSP, aka. LSF)
    \item Fundamental Frequency (via ACF/Cepstrum method and via Subharmonic-Summation (SHS))
    \item Probability of Voicing from ACF and SHS spectrum peak
    \item Voice-Quality: Jitter and Shimmer
    \item Formant frequencies and bandwidths
    \item Zero- and Mean-Crossing rate
    \item Spectral features (arbitrary band energies, roll-off points, centroid, entropy, maxpos, min- pos, variance (=spread), skewness, kurtosis, slope)
    \item Psychoacoustic sharpness, spectral harmonicity
    \item CHROMA (octave warped semitone spectra) and CENS features (energy normalised and smoothed CHROMA)
    \item CHROMA-derived Features for Chord and Key recognition
    \item F0 Harmonics ratios" \cite{openSMILE}
\end{itemize}

``In order to map contours of audio  low-level descriptors onto a vector of fixed dimensionality, the following functionals can be applied:

\begin{itemize}
    \item Extreme values and positions
    \item Means (arithmetic, quadratic, geometric)
    \item Moments (standard deviation, variance, kurtosis, skewness)
    \item Percentiles and percentile ranges
    \item Regression (linear and quadratic approximation, regression error)
    \item Centroid
    \item Peaks
    \item Segments
    \item Sample values
    \item Times/durations
    \item Onsets/Offsets
    \item Discrete Cosine Transformation (DCT)
    \item Zero-Crossings
    \item Linear Predictive Coding (LPC) coefficients and gain" \cite{openSMILE}
    
\end{itemize}

\pagebreak

% \chapter{Further Results}

%%%%%%%%%%%%%%%%%%%%%%%%%%%%%%%%%%%%
%%%%%%%%%%%%%%%%%%%%%%%%%%%%%%%%%%%%

%% bibliography

\printbibliography[title={Bibliography}]

\end{document}

%% file: includes.tex
\usepackage[a4paper,hmargin=2.8cm,vmargin=2.0cm,includeheadfoot]{geometry}
\usepackage{textpos}
\usepackage{tabularx,longtable,multirow,subfigure,caption}%hangcaption
\usepackage{fncylab} %formatting of labels
\usepackage{fancyhdr} % page layout
\usepackage{url} % URLs
\usepackage[english]{babel}
\usepackage{amsmath}
\usepackage{graphicx}
\usepackage{dsfont}
\usepackage{epstopdf} % automatically replace .eps with .pdf in graphics
\usepackage{array}
\usepackage{latexsym}
\usepackage[hypertexnames=false,colorlinks]{hyperref} % provide links in pdf

\hypersetup{pdftitle={},
  pdfsubject={}, 
  pdfauthor={},
  pdfkeywords={}, 
  pdfstartview=FitH,
  pdfpagemode={UseOutlines},% None, FullScreen, UseOutlines
  bookmarksnumbered=true, bookmarksopen=true, colorlinks,
    citecolor=black,%
    filecolor=black,%
    linkcolor=black,%
    urlcolor=black}

\usepackage[all]{hypcap}

\def\@makechapterhead#1{%
  \vspace*{10\p@}%
  {\parindent \z@ \raggedright \sffamily
    \interlinepenalty\@M
    \Huge\bfseries \thechapter \space\space #1\par\nobreak
    \vskip 30\p@
  }}

\def\@makeschapterhead#1{%
  \vspace*{10\p@}%
  {\parindent \z@ \raggedright
    \sffamily
    \interlinepenalty\@M
    \Huge \bfseries  #1\par\nobreak
    \vskip 30\p@
  }}

\allowdisplaybreaks

%% file: titlepage.tex
% Last modification: 2015-08-17 (Marc Deisenroth)
\begin{titlepage}

\newcommand{\HRule}{\rule{\linewidth}{0.5mm}} % Defines a new command for the horizontal lines, change thickness here

%----------------------------------------------------------------------------------------
%	LOGO SECTION
%----------------------------------------------------------------------------------------

\includegraphics[width = 4cm]{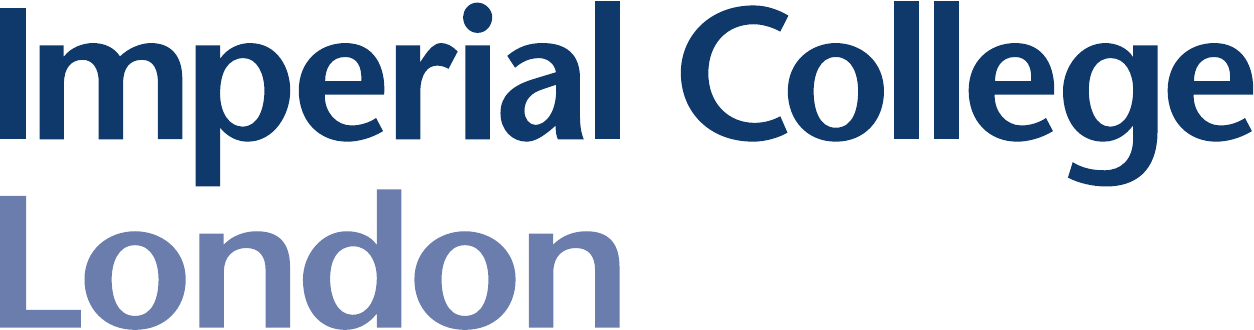}\\[0.5cm] 

\center % Center remainder of the page

%----------------------------------------------------------------------------------------
%	HEADING SECTIONS
%----------------------------------------------------------------------------------------

\textsc{\Large Imperial College London}\\[0.5cm] 
\textsc{\large Department of Computing}\\[0.5cm] 

%----------------------------------------------------------------------------------------
%	TITLE SECTION
%----------------------------------------------------------------------------------------

\HRule \\[0.4cm]
{ \huge \bfseries \reporttitle}\\ % Title of your document
\HRule \\[1.5cm]
 
%----------------------------------------------------------------------------------------
%	AUTHOR SECTION
%----------------------------------------------------------------------------------------

\begin{minipage}{0.4\textwidth}
\begin{flushleft} \large
\emph{Author:}\\
\reportauthor % Your name
\end{flushleft}
\end{minipage}
~
\begin{minipage}{0.4\textwidth}
\begin{flushright} \large
\emph{Supervisor:} \\
\supervisor % Supervisor's Name
\end{flushright}
\end{minipage}\\[4cm]

%----------------------------------------------------------------------------------------
%	FOOTER & DATE SECTION
%----------------------------------------------------------------------------------------
\vfill % Fill the rest of the page with whitespace
Submitted in partial fulfillment of the requirements for the MSc degree in
\degreetype~of Imperial College London\\[0.5cm]

\makeatletter
\@date 
\makeatother

\end{titlepage}